\documentclass[lettersize,journal]{IEEEtran}
\usepackage{amsmath,amsfonts}
\usepackage{algorithmic}
\usepackage{algorithm}
\usepackage{array}
\usepackage[caption=false,font=normalsize,labelfont=sf,textfont=sf]{subfig}
\usepackage{textcomp}
\usepackage{stfloats}
\usepackage{url}
\usepackage{verbatim}
\usepackage{graphicx}
\usepackage{cite}
\usepackage{booktabs}
\usepackage{makecell}
\usepackage{pifont}
\usepackage{xcolor}
\usepackage{soul}
\soulregister{\cite}{7}
\soulregister{\ref}{7}
\begin{document}

\title{Unsupervised Domain Adaption Harnessing Vision-Language Pre-training}

\author{Wenlve Zhou, Zhiheng Zhou *
\thanks{ 
Wenlve Zhou, Zhiheng Zhou are with School of Electronic and Information Engineering, South China University of Technology, Guangzhou 510641, Guangdong, China, and also with Key Laboratory of Big Data and Intelligent Robot, Ministry of Education, South China University of Technology, Guangzhou 510641, China (email: wenlvezhou@163.com; zhouzh@scut.edu.cn).(* Corresponding author: Zhiheng Zhou.)

Copyright © 2024 IEEE. Personal use of this material is permitted.
However, permission to use this material for any other purposes must be
obtained from the IEEE by sending an email to pubs-permissions@ieee.org.

The definitive version of this paper can be found at: 10.1109/TCSVT.2024.3391304

}}

\markboth{Journal of \LaTeX\ Class Files,~Vol.~14, No.~8, August~2021}%
{Shell \MakeLowercase{\textit{et al.}}: A Sample Article Using IEEEtran.cls for IEEE Journals}

\IEEEpubid{0000--0000/00\$00.00~\copyright~2021 IEEE}

\maketitle

\begin{abstract}
This paper addresses two vital challenges in Unsupervised Domain Adaptation (UDA) with a focus on harnessing the power of Vision-Language Pre-training (VLP) models. Firstly, UDA has primarily relied on ImageNet pre-trained models. However, the potential of VLP models in UDA remains largely unexplored. The rich representation of VLP models holds significant promise for enhancing UDA tasks. To address this, we propose a novel method called Cross-Modal Knowledge Distillation (CMKD), leveraging VLP models as teacher models to guide the learning process in the target domain, resulting in state-of-the-art performance. Secondly, current UDA paradigms involve training separate models for each task, leading to significant storage overhead and impractical model deployment as the number of transfer tasks grows. To overcome this challenge, we introduce Residual Sparse Training (RST) exploiting the benefits conferred by VLP's extensive pre-training, a technique that requires minimal adjustment (approximately 0.1\%$\sim$0.5\%) of VLP model parameters to achieve performance comparable to fine-tuning. Combining CMKD and RST, we present a comprehensive solution that effectively leverages VLP models for UDA tasks while reducing storage overhead for model deployment. Furthermore, CMKD can serve as a baseline in conjunction with other methods like FixMatch, enhancing the performance of UDA. Our proposed method outperforms existing techniques on standard benchmarks. Our code will be available at: https://github.com/Wenlve-Zhou/VLP-UDA.
\end{abstract}


\section{Introduction}
\IEEEPARstart{D}{eep} learning has been witnessed remarkable progress recently \cite{ref1, ref2, ref3}, yet it has not performed as well when applied to the actual target dataset as neural networks are sensitive to domain gaps. Fortunately, Unsupervised Domain Adaptation (UDA) has emerged as a critical research area that transferring knowledge from labeled source domains to unlabeled target domains, offering potential solutions to address the limitations of deep learning \cite{ref4,ref5,ref6,ref7}. While considerable advancements have been made in UDA, two crucial challenges have received insufficient attention. Firstly, the considerable potential of VLP's rich feature representation for UDA tasks remains largely unexplored, such as CLIP \cite{ref8}, FLIP \cite{ref9}, etc. Secondly, current UDA methods predominantly rely on training a distinct full-parameter model for each target domain, which not only poses inconveniences during model deployment but also demands substantial storage capacity for the parameters.
\begin{figure}[!t]
	\setlength{\abovecaptionskip}{0.cm}
	\setlength{\belowcaptionskip}{-0.cm}
	\centering
	\includegraphics[width=3.5in]{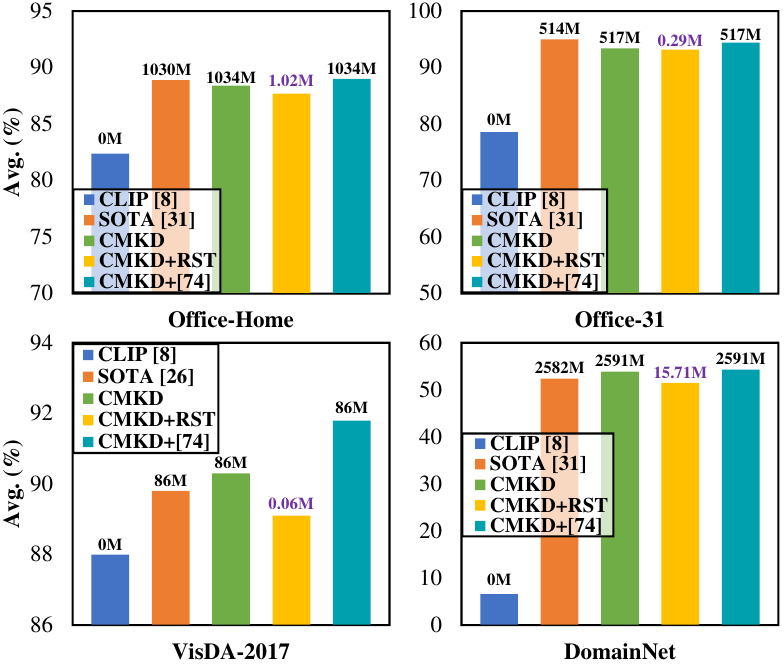}
	\caption{The bar chart displays the average accuracy of our method and previous State-Of-The-Art (SoTA) on popular benchmarks. The statics on each bar represent the Downstream Parameters (DSP) of the respective method (The definition of DSP is detailed in Section IV). Combining CMKD with RST leads to a substantial reduction in model storage overhead, while its combination with FixMatch [74] results in further enhanced model performance. CLIP [8] represents the zero-shot inference performance on target domain.}
	\label{fig_1}
\end{figure}

UDA approaches capitalize on prior knowledge from pre-trained models to adapt to target domains. Previous studies have primarily employed models pre-trained on ImageNet-1k \cite{ref10} or ImageNet-21k \cite{ref11}. In the era of large-scale vision-language pre-training, VLP models have gained significant traction. The extensive training provides rich representations that offer tremendous potential for boosting UDA tasks. 
\IEEEpubidadjcol
Surprisingly, despite extensive pre-training on vision-language datasets, VLP model still fall short of the state-of-the-art methods in terms of zero-shot performance on relevant benchmarks (see Figure 1). It suggests that the VLP model hasn't achieved comprehensive coverage across various data distributions. Moreover, applying previous methods to VLP has the potential to distort semantic feature structures and compromise class discriminability \cite{ref27}. Hence, effectively leveraging the pre-training features becomes a critical aspect. The two common tunning approaches for VLP models are ``Visual Encoder Only'' and ``Prompt Tuning'', as shown in Figure 2(a). The former involves excluding the text encoder and solely fine-tuning the visual encoder \cite{ref12}, which inadvertently sacrifices the rich general knowledge encapsulated in the text encoder—a pivotal aspect for acquiring domain-specific knowledge. Similar to how humans rely on common-sense to expedite learning in specialized subjects, models can benefit from this general knowledge. The latter approach, ``Prompt Tuning,'' freezes the vision-language encoder and exclusively fine-tunes newly added learnable tokens via Image-Text Contrastive learning (ITC) \cite{ref13}. Although this approach preserves the model's general knowledge and its efficiency has been validated in other fields \cite{ref80}, considering the limited model capacity in the UDA task, the representation of the frozen visual encoder may not necessarily cover the target domain in UDA. Additionally, this approach incurs additional computational costs during model inference due to the inclusion of the text encoder.

Another significant challenge manifests during the model deployment phase. As business requirements evolve, numerous diverse target domains may emerge. Traditional methods typically necessitate training a full-parameters model for each new target domain. In real-world deployment scenarios, preserving models for both source and target domains demands substantial parameter storage, resulting in inefficiencies. Therefore, it becomes crucial to explore strategies that facilitate model adaptation with minimal additional storage parameters for target domains. One promising avenue is the exploration of methods such as Parameter-Efficient Fine-Tuning (PEFT) \cite{ref14} in Large Language Models (LLMs) \cite{ref15}, where a few extra parameters are trained in the training phase, and inference without additional burden via re-parameterization tricks \cite{ref16}. The visual encoder in VLP models can act like LLMs, and the heavy storage and deployment problems can be solved by training a small number of additional parameters for each domain along the lines of PEFT, as shown in Figure 2(b). However, the effectiveness of such methods on relatively lightweight models remains uncertain, such as ResNet50 \cite{ref17} and ViT-B \cite{ref18}.

In this paper, we address these challenges by proposing novel methods to integrate visual-language models into UDA tasks and explore efficient model deployment strategies. To address Challenge 1, we propose a simple but effective method called Cross-Modal Knowledge Distillation (CMKD). By leveraging general knowledge from the text encoder as prior, this method assists the model in self-training on the target domain data. The CMKD can be easily implemented with a simple loss function. For Challenge 2, we introduce a novel parameter-efficient training method called Residual Sparse Training (RST), which does not require modifying the network architecture. With RST, model deployment can be achieved using a highly sparse weights with almost no performance degradation. Both of these techniques can be applied to the CNNs or Transformers architectures within VLP models. We conduct extensive experiments and evaluations on various benchmarks to demonstrate the effectiveness and efficiency of our proposed approaches.
\begin{figure*}[!t]
	\setlength{\abovecaptionskip}{0cm}  
	\setlength{\belowcaptionskip}{-0.2cm} 
	\centering
	\includegraphics[width=6in]{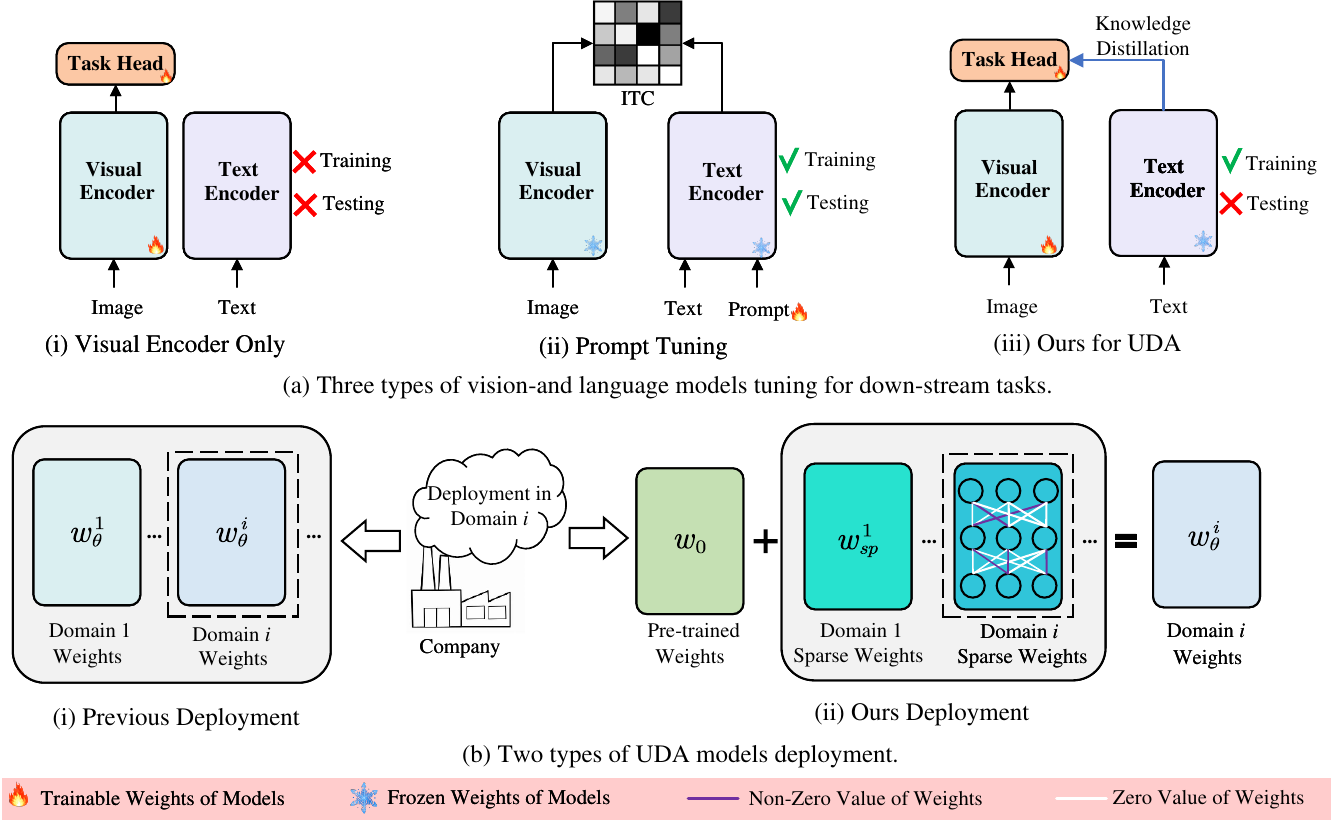}
	\caption{Overview of VLP models tuning and UDA deployment. (a) VLP models tuning. The previous methods struggled to balance the general knowledge utilization and domain-specific representation. We introduce cross-modal knowledge distillation enabling reconciliation and remove text encoder when inference. (b) UDA deployment. Conventional UDA deployment selects domain weights with full parameters based on the application scenario. Our method learns a highly sparse weight (approximately 0.1\%$\sim$0.5\% of the downstream models' parameters) that can be added to the pre-trained model for deployment.}
	\label{Fig2}
\end{figure*}
Our contributions are summarized as follows: 

(1)	We incorporate VLP into UDA tasks and propose Cross-Modal Knowledge Distillation (CMKD) to efficiently utilize the prior knowledge from the text encoder.

(2)	With the Residual Sparse Training (RST), we achieve the goal of domain adaptation, while significantly reducing the parameter storage and improving the deployment efficiency.

(3)	In extensive experimental evaluations, our method has demonstrated exceptional performance on multiple benchmarks, while also exhibiting efficient computational and memory requirements. This validates the feasibility and practicality of our approach.

The subsequent sections of this paper are organized as follows: Section II presents a comprehensive literature review. In Section III, we introduce our proposed methodologies. Section IV provides detailed information regarding the experimental setup and presents the evaluation results. Finally, Section V concludes the paper and outlines potential directions for future research.

\section{Related Work}
In this section, we present a brief survey on the most related fields of our works in four aspects: visual-language pre-training, unsupervised domain adaption, knowledge distillation, and parameter efficient fine-tuning.
\subsection{Visual-Language Pre-training}
Visual-language pre-training have been witnessed significant advancements in recent research, with two prominent paradigms: single-tower models and dual-tower models. Single-tower models aim to jointly encode images and text using a shared visual-linguistic encoder, such as the Transformer architecture, mapping them into a common embedding space \cite{ref34,ref35, ref86}. Through training on alignment tasks for image-text pairs, single-tower models learn semantic correlations between images and text. On the other hand, dual-tower models employ separate visual and language encoders to process image and text inputs independently \cite{ref8,ref9, ref88}. The visual encoder transforms images into visual feature representations, while the language encoder generates semantic vectors from text. By training on alignment tasks, dual-tower models learn to establish correspondences between visual and linguistic modalities.

To address the feature entanglement issue, the dual-tower model is utilized in this paper, as it emphasizes the characterization of individual modes instead of solely focusing on feature interactions between them. This enables the text encoder's knowledge to serve as common knowledge, facilitating the training of downstream tasks.

\subsection{Unsupervised Domain Adaption}
Unsupervised domain adaptation has emerged as a prominent research area in computer vision, aiming to address the domain shift problem. UDA methods seek to leverage labeled data from a source domain to improve the performance of a model on a target domain where labeled data is unavailable. Various approaches have been proposed to tackle UDA, including domain adversarial learning \cite{ref19,ref20}, self-training \cite{ref21}, and consistency regularization \cite{ref22}. These methods seek to align the feature distributions between the source and target domains, thereby facilitating knowledge transfer. In recent years, due to Transformers’ \cite{ref23} powerful contextual modeling capabilities, it has swept the major fields of computer vision, including unsupervised domain adaptation. These methods are based on Transformer's inductive bias, designing cross-domain attention \cite{ref24,ref7}, patch-mix mechanism \cite{ref26}. Thanks to the effectiveness of the Transformer framework, UDA tasks have seen significant performance improvements in recent years. 

However, prior studies have seldom taken into account the utilization of pre-training on visual-language data, and although Ge et al. \cite{ref27} introduced DAPL into UDA, its performance improvement is limited compared to the State-Of-The-Art (SoTA), and its application scenario may be constrained by the requirement for precise domain descriptions. Moreover, each of these methods necessitates “one-domain-one-model” paradigm, creating significant obstacles for the deployment of UDA.

\subsection{Knowledge Distillation}
Knowledge distillation has gained significant attention in machine learning as a powerful technique for transferring knowledge from a complex teacher model to a simpler student model. The primary objective is to enhance the performance of the student model by leveraging the knowledge encapsulated in the teacher model. By mimicking the output behaviour of the teacher model, the student model can benefit from the rich information provided by the teacher. Various approaches have been proposed to effectively carry out knowledge distillation. One common method is to minimize the discrepancy between the predictions of the student model and the soft targets generated by the teacher model. This can be achieved through techniques such as KL divergence, which encourages the student model to produce similar output probabilities as the teacher model \cite{ref28}. Additionally, intermediate representations \cite{ref29}, structural knowledge \cite{ref30} from the teacher model can be utilized as additional supervision signals to guide the learning process of the student model.

In this paper, drawing inspiration from knowledge distillation, the proposed CMKD effectively transfers common knowledge from the VLP models of text encoders to assist the model in unlabelled target data adaption.

\subsection{Parameter Efficient Fine-tuning}
Parameter efficient fine-tuning techniques aim to adapt a frozen pre-trained model to a specific target task or domain using minimal additional parameters. These methods \cite{ref76, ref77, ref31, ref32, ref33, ref78, ref79, ref14} are particularly valuable in resource-constrained environments or situations where storing multiple models is impractical. Approaches like prefix tuning \cite{ref31} and prompt tuning \cite{ref32, ref78, ref79, ref87} explore strategies to excavate base model with a few learnable tokens. Adapter \cite{ref33} adds new network layers or modules between network layers inside the pre-trained model to adapt to downstream tasks. More recently, LoRA \cite{ref14} has gained popularity for tuning large models due to its effectiveness and negligible additional inference overhead, which has achieved performance comparable to fine-tuning with the full parameters. Besides, for any downstream task, it is only necessary to save the relevant additional parameters instead of the entire model.

LoRA \cite{ref14} offers a solution to the ``one-domain-one-model'' problem. However, its effectiveness significantly diminishes when transitioning from LLMs to models with less parameters in VLP models. To address this limitation, we propose RST, a technique that achieves superior performance while requiring less storage for additional parameters.

\section{Methods}
In the unsupervised domain adaption, data from two different distributions will be sampled to form the source labeled domain ${{D}_{s}}=\text{ }\!\!\{\!\!\text{ (}x_{i}^{s}\text{,}y_{i}^{s}\text{) }\!\!\}\!\!\text{ }_{i=1}^{{{n}_{s}}}$ and unlabeled target domain ${{D}_{t}}=\{x_{j}^{t}\}_{j=1}^{{{n}_{t}}}$ datasets, where ${{n}_{s}}$ and ${{n}_{t}}$ denote the quantity of data in each domain. ${{y}^{s}}\in {{\mathbb{R}}^{{{n}_{s}}\times c}}$ and \emph{c} represent the number of classes in the UDA datasets. The primary objective of UDA is to learn a domain-invariant network that can adapt to the target domain using source data and labels, even in the absence of labeled target domain data during training.

In Section III.A, We will first give an introduction to the dual-tower model CLIP \cite{ref8} as preliminary. Based on the analysis of Section I and the inspiration from the knowledge distillation, we propose cross-modal knowledge distillation ${{\mathcal{L}}_{\text{cmkd}}}$ to assists the model in self-training on the target domain data, as described in Section III.B. Furthermore, in Section III.C, we introduce residual sparse training to improve the flexibility of deployment in response to the current heavy overhead of UDA deployment.

For UDA, the objective is formulated as
\begin{eqnarray}
	{{\mathcal{L}}_{\text{total}}}={{\mathcal{L}}_{\text{cls}}}+{{\mathcal{L}}_{\text{cmkd}}}
\end{eqnarray}
where ${{\mathcal{L}}_{\text{cls}}}$ is the standard classification loss of source domain implemented with Cross-Entropy.

\subsection{Preliminary}
Our backbone architecture is based on CLIP \cite{ref8}, consisting of an image encoder, denoted as $f(\cdot )$, and a text encoder, denoted as $\text{g}(\cdot )$. The image encoder can be either ResNet \cite{ref17} or Vision Transformer \cite{ref18}, while the text encoder adopts a Transformer \cite{ref23}. These encoders transform the high-dimensional image and text inputs into a lower-dimensional feature space.

CLIP \cite{ref8} is trained in a contrastive manner using image-text pairs. Each input text describes a category using the format ``a photo of a [CLASS]'' (where [CLASS] represents the class token). A positive pair consists of an image ${{x}_{i}}$ and its corresponding text ${{t}_{i}}$, which describes the category of ${{x}_{i}}$. On the other hand, a negative pair pairs an image ${{x}_{i}}$ with an irrelevant description ${{t}_{j}}$, where $i\ne j$. The training objective aims to maximize the cosine similarity of positive pairs and minimize the cosine similarity of negative pairs. This contrastive learning objective aligns the image and text representations in the same feature space.

By leveraging these aligned features, CLIP \cite{ref8} enables zero-shot inference. Given $K$ category descriptions, when forwarding an image $x$, the model predicts that it belongs to the category ${{y}_{i}}$ can be described as:
\begin{eqnarray}
	{{p}_{\text{g}}}(y=i|x)=\frac{\exp (\left\langle \text{g}({{t}_{i}}),f(x) \right\rangle )}{\sum\nolimits_{k=1}^{K}{\exp (\left\langle \text{g}({{t}_{k}}),f(x) \right\rangle )}}
\end{eqnarray}
where $\left\langle \cdot ,\cdot  \right\rangle $ denotes the cosine similarity.

Both a text encoder and a visual encoder are utilized to facilitate unsupervised domain adaptation. Throughout the training process, we introduce an auxiliary task head $h(\cdot )$ in conjunction with the visual encoder to establish a downstream model, while maintaining the text encoder in a frozen state. Due to the extensive pre-training of CLIP on a dataset consisting of 400 million image-text pairs \cite{ref8}, we posit that the ${{p}_{\text{g}}}$ arises from the common-sense reasoning of model. This general knowledge bolsters the downstream model in adaption, equipping the model to execute UDA. Additionally, during the inference phase, the text encoder is omitted to alleviate the burden on the inference process.
\begin{figure}[t]
	\setlength{\abovecaptionskip}{0cm}  
	\setlength{\belowcaptionskip}{-0.2cm} 
	\centering
	\includegraphics[width=2in]{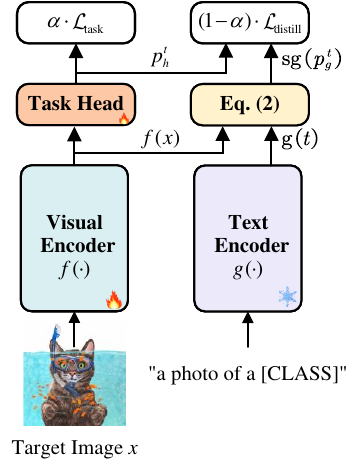}
	\caption{The pipeline of cross-modal knowledge distillation. CMKD is exclusively utilized for target data.}
	\label{fig_3}
\end{figure}
\subsection{Cross-Modal Knowledge Distillation}
Based on the previous analysis, we anticipate that the knowledge distillation framework, incorporating the common knowledge of the CLIP \cite{ref8} model, could facilitate adaptation for unlabeled target data. The classical knowledge distillation framework \cite{ref28} employs KL divergence to minimize the disparity between the distributions of the student and teacher models, facilitating effective knowledge transfer, which can be formulated as:
\begin{eqnarray}
	{{\mathcal{L}}_{\text{kd}}}=\alpha\! \cdot\! \text{KL(}y\text{ }\!\!||\!\!\text{ }{{p}_{\text{student}}}\text{)}\!+\!(1-a)\!\cdot \!\text{KL(\emph{sg}(}{{p}_{\text{teacher}}}\text{) }\!\!||\!\!\text{ }{{p}_{\text{student}}})
\end{eqnarray}
where $y$ represents the ground truth, ${{p}_{\text{student}}}$ denotes the output of the student model,  ${{p}_{\text{teacher}}}$ signifies the output of the teacher model, and $\text{\emph{sg}(}\cdot \text{)}$ is the operation of stop-gradient. For convenience, the first term quantifies the discrepancy between the predictions of the student model and the ground-truth, referred to as the task term ${{\mathcal{L}}_{\text{task}}}$. The second term assesses the divergence between the predictions of the student model and those of the teacher model, known as the distillation term ${{\mathcal{L}}_{\text{distill}}}$. The parameter $a$ represents the trade-off for balancing these two tasks.

However, due to the absence of corresponding labels in the target domains, executing the task term becomes challenging. To achieve a model capable of approximating learning in task-term, we adopt self-training \cite{ref37}. A common practice is to minimize the uncertainty of predictions on unlabeled data with Gibbs entropy. Yet the Gibbs entropy is verified that make the predictions over-confident \cite{ref21}, we use weaker penalization Gini impurity (GI) for self-training, where the objective and gradient function can be formulated as:
\begin{eqnarray}
{{\mathcal{L}}_{\text{task}}}=\text{GI}(p_{h}^{t})=1-\sum_{i=1}^{c}{[{{p_{h}^{t}(y=i|{{x}^{t}})}]^{2}}}
\end{eqnarray}
\begin{eqnarray}
	\frac{\partial {{\mathcal{L}}_{\text{task}}}}{\partial \theta }=-\sum_{i=1}^{c}{[2\cdot p_{h}^{t}(y=i|{{x}^{t}})\cdot \frac{\partial p_{h}^{t}(y=i|{{x}^{t}})}{\partial \theta }]}
\end{eqnarray}
where $p_{h}^{t}=h(f({{x}^{t}}))$ denotes the output from the task-head, and $\theta$ refer to the trainable components, namely the task-head and visual encoder parameters.

For the distillation term, a straightforward approach is to directly utilize ${{\mathcal{L}}_{\text{distill}}}$ in the vanilla KD \cite{ref28}, written as:
\begin{eqnarray}
	{{\mathcal{L}}_{\text{distill}}}=\text{KL(\emph{sg}(}p_{\text{g}}^{t})||p_{h}^{t}\text{)}
\end{eqnarray}
Yet Figure 1 reveals that CLIP's zero-shot performance lags behind that of SoTA. Optimizing Eq. (6) would punish the student perspective $p_{h}^{t}$ closer to the teacher perspective $p_{\text{g}}^{t}$, potentially adversely affecting model performance instead. Hence, the primary role of $p_{\text{g}}^{t}$ should be as a common-sense auxiliary for the expertise learning of the model instead of ground truth. Although the probability distribution of the task-head $p_{h}^{t}$ may differ from $p_{\text{g}}^{t}$, $p_{h}^{t}$ should remain dominant, while $p_{\text{g}}^{t}$ serves only as a constraining factor. The distillation term can be rewritten as:
\begin{eqnarray}
\mathcal{L}_{\text {distill }}=\mathrm{GI}\left(p_{m}^{t}\right)=1-\sum_{i=1}^{c}\left[p_{m}^{t}\left(y=i \mid x^{t}\right)\right]^{2}
\end{eqnarray}
\begin{eqnarray}
	p_{m}^{t}=0.5 \cdot\left(p_{h}^{t}+\text{\emph{sg}}\left(p_{\text{g}}^{t}\right)\right)
\end{eqnarray}

In the above objective, we employ self-training to train the mixed distribution of $p_{h}^{t}$ and $p_{\text{g}}^{t}$, with $p_{\text{g}}^{t}$ being held constant throughout the training process. To better understand the role, the gradient of the Eq. (7) is first given as:
\begin{eqnarray}
\begin{aligned}
	\frac{\partial \mathcal{L}_{\mathrm{distill}}}{\partial \theta}= & -\sum_{i=1}^{c} p_{h}^{t}\left(y=i \mid x^{t}\right) \cdot \frac{\partial p_{h}^{t}\left(y=i \mid x^{t}\right)}{\partial \theta} \\
	& -\sum_{i=1}^{c} p_{\text{g}}^{t}\left(y=i \mid x^{t}\right) \cdot \frac{\partial p_{h}^{t}\left(y=i \mid x^{t}\right)}{\partial \theta}
\end{aligned}
\end{eqnarray}
We observe that $p_{\text{g}}^{t}$ serves as a constraint on the gradient compared to Eq. (5). In cases where there is a significant disparity between $p_{h}^{t}$ and $p_{\text{g}}^{t}$, the model tends to gravitate towards the mixed distribution rather than attempting to force the constraint to match $p_{\text{g}}^{t}$.
The pipeline is shown in Figure 3 and the overall objectives of CMKD are:
\begin{eqnarray}
	{{\mathcal{L}}_{\text{cmkd}}}=\alpha \cdot {{\mathcal{L}}_{\text{task}}}+(1-a)\cdot {{\mathcal{L}}_{\text{distill}}}
\end{eqnarray}

After exploring the fundamental framework of CMKD, we observed that the hyper-parameter $\alpha$ in KD \cite{ref28} is typically determined through experience to strike a balance between distillation loss and task loss. However, in unsupervised domain adaptation, where labels for knowledge transfer are unavailable, we suggest employing a dynamic mechanism to produce distinct trade-offs for each sample. We opted to use the discrepancy between $p_{h}^{t}$ and $p_{\text{g}}^{t}$ as the trade-off $\alpha$, dynamically determining the weighting of the two tasks based on evaluating the consistency between the students' and the teachers' perspective, written as:
\begin{eqnarray}
	\alpha =\emph{sg}(\exp (-\text{KL}(p_{h}^{t}\parallel p_{\text{g}}^{t})))
\end{eqnarray}
where $\alpha \in {{\mathbb{R}}^{{{n}_{t}}}}$, range from 0 to 1. 

Furthermore, since the visual encoder is trainable, the distribution of $p_{\text{g}}^{t}$ will gradually drift. To ensure the stability of training, we incorporate a regularization term ${{\mathcal{L}}_{\text{reg}}}$ that constrains the parameter space based on the CLIP \cite{ref8} fine-tuning paradigm. As a result, Eq. (10) is reformulated as follows:
\begin{eqnarray}
	{{\mathcal{L}}_{\text{cmkd}}}={{\lambda }_{1}}\cdot (\alpha \cdot {{\mathcal{L}}_{\text{task}}}+(1-a)\cdot {{\mathcal{L}}_{\text{distill}}})+{{\mathcal{L}}_{\text{reg}}}
\end{eqnarray}
\begin{eqnarray}
\mathcal{L}_{\text {reg }}=\lambda_{2} \cdot \operatorname{KL}\left(y|| p_{g}^{s}\right)+\lambda_{3} \cdot \operatorname{GI}\left(p_{\text{g}}^{t}\right)
\end{eqnarray}
The trade-off parameters $\lambda_{1}$, $\lambda_{2}$ and $\lambda_{3}$ are introduced to balance the multi-loss

\begin{algorithm}[tp]
	\caption{RST Training.}\label{alg:alg1}
	\begin{algorithmic}
		\STATE 
		\STATE $ \textbf{Input: } \text{task } \emph{i} \text{, task weight } w_{\theta }^{i} \text{, pre-trained weight } {{w}_{0}} \text{, pre-}$
		\STATE $ \text{defined threshold } \tau $
		\STATE $ \textbf{for} \text{ t = 0 to MaxIter } \textbf{do}$
		\STATE \hspace{0.5cm}$\text{Obtain } w_{\theta }^{i} \text{ with UDA training.}$
		\STATE \hspace{0.5cm}$\text{With } {{w}_{0}} \text{ and }\tau \text{, update } w_{\theta }^{i} \text{ by Eq. (17)}$
		\STATE $ \textbf{end for}$
		\STATE$\text{With } {{w}_{0}} \text{ gain } w_{\text{sp}}^{i} \text{ by Eq. (16)}$
		\STATE$\text{Preserve } w_{\text{sp}}^{i} \text{ in sparse form.}$
		\STATE $ \textbf{Return } w_{\text{sp}}^{i}$
	\end{algorithmic}
	\label{alg1}
\end{algorithm}

\begin{algorithm}[tp]
	\caption{RST Inference.}\label{alg:alg2}
	\begin{algorithmic}
		\STATE 
		\STATE $ \textbf{Input: } \text{task } \emph{i} \text{, pre-trained weight } {{w}_{0}} \text{, sparse series } \{w_{\text{sp}}^{k}\}_{k=1}^{n}$
		\STATE$\text{Retrieve } w_{\text{sp}}^{i} \text{ from the series } \{w_{\text{sp}}^{k}\}_{k=1}^{n}$
		\STATE$\text{Transform to dense form.}$
		\STATE$\text{Obtain task weight with addition:}$
		\[w_{\theta }^{i}={{w}_{0}}+w_{sp}^{i}\]
		\STATE $ \textbf{Return } w_{\theta }^{i}$
	\end{algorithmic}
	\label{alg2}
\end{algorithm}

\subsection{Residual Sparse Training}
During the adaptation of LLM to downstream tasks, LoRA \cite{ref14} successfully achieves remarkable generalization with minimal parameter storage and negligible performance degradation. Due to the low intrinsic dimension of LLM \cite{ref39}, Hu et al. \cite{ref14} put forward the hypothesis that updates to the weights during adaptation also exhibit a low intrinsic rank and propose LoRA, formulated as:
\begin{eqnarray}
	h={{w}_{0}}\cdot x+\Delta w\cdot x={{w}_{0}}\cdot x+B\cdot A\cdot x
\end{eqnarray}
where ${{w}_{0}}\in {{\mathbb{R}}^{d\times k}}$ is a pre-trained weight matrix. The update of the weight is constrained and update by representing the latter with a low-rank decomposition ${{w}_{0}}+\Delta w={{w}_{0}}+B\cdot A$, where $B\in {{\mathbb{R}}^{d\times r}}$, $A\in {{\mathbb{R}}^{r\times k}}$, and the rank $r\ll \min (d,k)$.

However, the assumption that low intrinsic dimension of updated weights may not be applicable to VLP models due to the relatively small size of the models, such as ResNet50 \cite{ref17} and ViT-B \cite{ref18}. The theory of structural neuroplasticity in brain neuroscience \cite{ref40} provides valuable insight into the dynamic nature of neuronal connections. While neurons in the brain are predominantly hard-wired, localized regions have the capacity to remodel when learning or adapting to new environments. This phenomenon allows frequently used neurons to strengthen their connections, facilitating the brain's adaptation to changing circumstances. Inspired by this theory, we draw an analogy between the neural network fine-tuning process in downstream tasks and the brain's adaptation to new environments. During fine-tuning, weight connections exhibiting significant variations are deemed beneficial for learning in the downstream task, resembling the remodeling process of neuronal links in the brain. Conversely, weights with minor variations are considered irrelevant to the downstream task, akin to the hardwired regions of the brain that remain fixed as pre-training weights. This approach, termed Residual Sparse Training (RST), entails modifying only a minimal fraction of weights in the pre-trained model to fine-tune the parameters, mirroring the selective adaptation observed in the brain's structural neuroplasticity. Moreover, the index of the weights to be altered should be determined by the requirements of the UDA tasks. The RST is given as:
\begin{eqnarray}
	h={{w}_{\theta }}\cdot x={{w}_{0}}\cdot x+{{w}_{\text{sp}}}\cdot x
\end{eqnarray}
where ${{w}_{\theta }}\in {{\mathbb{R}}^{d\times k}}$ is a task weight during UDA training. Upon completion of the training process, the sparse weights ${{w}_{\text{sp}}}$ can be obtained using the residual paradigm, as expressed by the following equation:
\begin{eqnarray}
	{{w}_{\text{sp}}}={{w}_{\theta }}-{{w}_{0}}
\end{eqnarray}

Therefore, the focus shifts from learning a sparse weight ${{w}_{\text{sp}}}$ to learning a weight that closely resembles ${{w}_{0}}$. We employ a straightforward method to determine whether to retain the pre-trained weights by comparing the magnitude of the weight change with a predefined threshold $\tau $, written as:
\begin{eqnarray}
	{{w}_{\theta }}=\left\{ \begin{matrix}
		{{w}_{\theta }},\left| {{w}_{0}}-{{w}_{\theta }} \right|>\tau   \\
		{{w}_{0}},\left| {{w}_{0}}-{{w}_{\theta }} \right|\le \tau   \\
	\end{matrix} \right.
\end{eqnarray}

When considering $n$ downstream tasks, after conducting RST, a sparse series $\{w_{\text{sp}}^{k}\}_{k=1}^{n}$ is generated. This series is then summed with the pre-training weights ${w}_{0}$ to derive the task-specific sequence $\{w_{\theta}^{k}\}_{k=1}^{n}$. The entire process of RST is remarkably simple, requiring no modification to the network structure. As a result, it can be seamlessly applied to both CNNs and Transformers. We illustrate RST in Figure 2(b) and propose a pseudo-code implementation of training and inference in Algorithm 1 and Algorithm 2 respectively.

\section{Experiments}
We evaluate CMKD and RST against competitive UDA baseline on popular benchmarks. These datasets include \textbf{Office-Home}, \textbf{Office-31}, \textbf{Visda-2017}, \textbf{ImageCLEF-DA}, \textbf{DomainNet}. Apart from the datasets, digits classification constructed from \textbf{MNIST}, \textbf{USPS} and \textbf{SVHN} is taken into account. Besides, RST will be compared with the popular PEFT to explore the superiority of our method in VLP. In ablation studies, the various settings of our methods are investigated.
\subsection{Setup}
\textbf{Office-Home} \cite{ref41} contains 15,588 images, which consists of images from 4 different domains: Artistic images (Ar), Clip Art (Cl), Product images (Pr) and Real-World images (Re). Collected in office and home settings, the dataset contains images of 65 object categories for each domain. We use all domain combinations and construct 12 transfer tasks.

\textbf{Office-31} \cite{ref42} is a benchmark dataset for domain adaptation which collected from three distinct domains: Amazon (A), Webcam (W), and DSLR (D). The dataset comprises 4,110 images in 31 classes, taken by web camera and digital SLR camera with different photographical settings, respectively. 6 transfer tasks are performed to enable unbiased evaluation.

\textbf{Digits} is a common UDA benchmark for digit recognition, utilizing three subsets: SVHN (S) \cite{ref44}, MNIST (M) \cite{ref45}, and USPS (U) \cite{ref46}. We use the training sets from each domain to train our model, and publish the recognition results using the standard test set from the target domain.

\textbf{ImageCLEF-DA} \cite{ref47} is a benchmark dataset for the ImageCLEF 2014 domain adaptation challenge, constructed by picking 12 common categories shared by the three public datasets, each of which is designated a domain: Caltech-256 (C), ImageNet ILSVRC 2012 (I), and Pascal VOC 2012 (P). Each category has 50 photographs, and each domain has 600 images. We develop six transfer tasks using all domain combinations.

\textbf{VisDA-2017} \cite{ref43} is a difficult simulation-to-real dataset with two very separate domains: Synthetic, renderings of 3D models from various perspectives and under various lighting conditions; Real, natural images. Over 280K photos from 12 different classes make up its training, validation, and test domains.

\textbf{DomainNet} \cite{ref48} is a dataset that encompasses a wide range of common objects across six distinct domains. Each domain consists of 345 categories of objects. These domains encompass Clipart (clp), Real (rel), Sketch (skt), Infograph (inf), painting (pnt), and quickdraw (qdr). We develop 30 transfer tasks using all domain combinations. 

\textbf{Baseline Methods for UDA}. We compare our method with state-of-the-art methods on UDA tasks, including MinEnt \cite{ref49}, DANN \cite{ref5}, DSAN \cite{ref6}, CDAN+E \cite{ref50}, CDAN+BSP \cite{ref51}, CDAN+TN \cite{ref52}, rRGrad+CAT \cite{ref22}, SWD \cite{ref54}, MSTN+DSBN \cite{ref55}, BNM \cite{ref57}, DCAN \cite{ref58}, SHOT \cite{ref59}, ATDOC-NA \cite{ref60}, CGDM \cite{ref61}, TVT \cite{ref24}, CDTrans \cite{ref7}, ADDA \cite{ref62}, ADR \cite{ref63}, CyCADA \cite{ref64}, SSRT \cite{ref65}, CDAN+MCC \cite{ref66}, DAN \cite{ref4}, D-CORAL \cite{ref67}, MADA \cite{ref68}, SDAT \cite{ref20}, DAPL \cite{ref27}, and PMTrans \cite{ref26}. 

\textbf{Baseline Methods for PEFT}. We compare our method with SoTA PEFT methods on UDA tasks, including LoRA \cite{ref14}, BitFit \cite{ref70}, AdaLoRA \cite{ref71}. 

\begin{table*}[t]
	\setlength{\abovecaptionskip}{0cm}  
	\setlength{\belowcaptionskip}{-0.2cm} 
	\centering
	\caption{Comparison with SoTA UDA methods on Office-Home. $^o$ implies its pre-trained from on ImageNet-21K instead of ImageNet-1K. $^\star$ is pre-trained from CLIP. The best performance is marked as bold.}
	\renewcommand{\arraystretch}{0.9}
	\setlength{\tabcolsep}{0.5mm}{
		\begin{tabular}{c|cccccccccccccc}
			\toprule
			Method & {Ar→Cl} & {Ar→Pr} & {Ar→Re} & {Cl→Ar} & {Cl→Pr} & {Cl→Re} & {Pr→Ar} & {Pr→Cl} & {Pr→Re} & {Re→Ar} & {Re→Cl} & {Re→Pr} & {Avg.} & {DSP(M)}\\
			\midrule
			\makecell[l]{\textit{ResNet50:}} &       &       &       &       &       &       &       &       &       &       &       &       &  \\
			DCAN  & 54.5  & 75.7  & 81.2  & 67.4  & 74.0    & 76.3  & 67.4  & 52.7  & 80.6  & 74.1  & 59.1  & 83.5  & 70.5 & 283.60\\
			BNM   & 52.3  & 73.9  & 80.0    & 63.3  & 72.9  & 74.9  & 61.7  & 49.5  & 79.7  & 70.5  & 53.6  & 82.2  & 67.9 & 283.60\\
			ATDOC-NA & 58.3  & 78.8  & 82.3  & 69.4  & 78.2  & 78.2  & 67.1  & 56.0    & 82.7  & 72.0    & 58.2  & 85.5  & 72.2 & 283.60\\
			DSAN  & 54.4  & 70.8  & 75.4  & 60.4  & 67.8  & 68,0    & 62.6  & 55.9  & 78.5  & 73.8  & 60.6  & 83.1  & 67.6 & 283.60\\
			SHOT  & 57.1  & 78.1  & 81.5  & 68.0    & 78.2  & 78.1  & 67.4  & 54.9  & 82.2  & 73.3  & 58.8  & 84.3  & 71.8 & 283.60\\
			DAPL$^\star$  & 54.1  & 84.3  & 84.8  & 74.4  & 83.7  & 85.0    & 74.5  & 54.6  & 84.8  & 75.2  & 54.7  & 83.8  & 74.5 & \textbf{0.00}\\
			\midrule
			CLIP$^\star$  & 51.6  & 81.9  & 82.6  & 71.9  & 81.9  & 82.6  & 71.9  & 51.6  & 82.6  & 71.9  & 51.6  & 81.9  & 72.0 & \textbf{0.00}\\
			Baseline$^\star$ & 52.1  & 69.7  & 75.2  & 52.5  & 64.3  & 63.9  & 56.2  & 48.4  & 74.9  & 68.9  & 54.4  & 82.7  & 63.7 & 460.40\\
			CMKD$^\star$  & 65.9      & 86.6       & 87.3      &   74.4    &  87.7    & 85.8      & 75.9       & 64.4      &  87.9    & 79.1      & 67.2      & 90.0      &79.3 & 460.40\\
			CMKD+RST$^\star$ & 64.9       & \textbf{87.3}      & 87.1      & 74.2     & 87.5      &85.4       & 74.5      & 63.0      & 88.0      &   78.1    &  66.5     &89.6       &78.8  &\textbf{1.23}\\
			CMKD+FixMatch$^\star$  & \textbf{67.2}       & \textbf{87.3}      &\textbf{87.7}       & \textbf{76.7}      & \textbf{88.2}      & \textbf{86.0}      &\textbf{76.8}       &\textbf{66.8}       &\textbf{88.4}       & \textbf{79.5}      &  \textbf{68.4}     &\textbf{90.4}       &\textbf{80.3}  &460.40\\
			\midrule
			\midrule
			\makecell[l]{\textit{ViT-B-16:}} &       &       &       &       &       &       &       &       &       &       &       &       &  \\
			CDTrans & 68.8  & 85.0    & 86.9  & 81.5  & 87.1  & 87.3  & 79.6  & 63.3  & 88.2  & 82.0    & 66.0    & 90.6  & 80.5 & 1030.20 \\
			TVT$^o$  & 74.9  & 86.8  & 89.4  & 82.8  & 88.0    & 88.3  & 79.8  & 71.9  & 90.1  & 85.5  & 74.6  & 90.6  & 83.6 & 1030.20\\
			SDAT$^o$ & 70.8  & 87.0   & 90.5  & 85.2  & 87.3  & 89.7  & 84.1  & 70.7  & 90.6  & 88.3  & 75.5  & 92.1  & 84.3 & 1030.20\\
			SSRT$^o$ & 75.2  & 89.0    & 91.1  & 85.1  & 88.3  & 90.0    & 85.0    & 74.2  & 91.3  & 85.7  & 78.6  & 91.8  & 85.5 & 1030.20\\
			PMTrans$^o$ & \textbf{81.2} & 91.6  & 92.4  & \textbf{88.9} & 91.6  & \textbf{93.0} & \textbf{88.5} & 80.0    & \textbf{93.4} & \textbf{89.5} & \textbf{82.4} & 94.5  & 88.9 & 1030.20\\
			\midrule
			CLIP$^\star$  & 67.8  & 89.0    & 89.8  & 82.9  & 89.0    & 89.8  & 82.9  & 67.8  & 89.8  & 82.9  & 67.8  & 89.0    & 82.4 & \textbf{0.00}\\
			Baseline$^\star$ & 71.2  & 83.6  & 88.5  & 78.9  & 85.8  & 85.5  & 71.9  & 71.2  & 86.3  & 81.8  & 73.2  & 90.7  & 80.7 & 1034.80\\
			CMKD$^\star$  & 79.4  & 94.2  & 92.7  & 86.3  & 93.4  & 92.2  & 86.7  & 79.5  & 92.1  & 88.2  & 81.2  & 94.5  & 88.4 & 1034.80\\
			CMKD+RST$^\star$ & 78.9  & 93.8  & 92.7  & 85.3  & 93.0    & 91.7  & 85.3  & 78.4  & 92.0    & 87.2  & 80.3  & 94.1  & 87.7 & \textbf{1.02}\\
			CMKD+FixMatch$^\star$ & 80.9  & \textbf{94.6} & \textbf{92.7} & 87.1  & \textbf{93.7} & 92.5  & 87.8  & \textbf{80.2} & 92.3  & 89.2  & 82.1  & \textbf{94.6} & \textbf{89.0} & 1034.80\\
			\bottomrule
		\end{tabular}%
	}
	\label{tabl}%
\end{table*}%

\textbf{Implementation Details}. We use a ResNet50 \cite{ref17} backbone for Office-Home, Office-31, Digits, ImageCLEF experiments and a ResNet-101 \cite{ref17} backbone for VisDA-2017. Additionally, we also report the performance of ViT-B-16 \cite{ref18} backbone on all 6 benchmarks. The models used in our experiments are pre-trained from CLIP \cite{ref8}. The ``Baseline'' indicates directly training a backbone on the source domain and testing on the target domain. For all tasks, mini-batch stochastic gradient descent (SGD) with momentum of 0.9 and the weight decay ratio 5e-4 are adopted to optimize the training process. During training, the text encoder and ``Batch Normalization'' in ResNet \cite{ref17} is frozen, and for ViT-B-16, we set the learning rate of the visual encoder to 3e-6 for most of our tasks, while the learning rate for VisDA is set to 3e-7. The learning rate of the task head is set to 1,000 times the visual encoder. For ResNet, we set the initial learning rate to 3e-9 on the VisDA dataset and 3e-7 for the remaining datasets. Additionally, the learning rate for the task head is configured to be 10,000 times that of the visual encoder. The tradeoff ${{\lambda }_{2}}$ is set to 0.1, while ${{\lambda }_{1}}$ and ${{\lambda }_{3}}$ adopt dynamic mechanism like DSAN to suppress noisy activations at the early stages of training: $\lambda =\frac{2\cdot \beta}{\exp \left( -1.0\cdot \mu  \right)}-1$. $\mu$ is the ratio of the current training iteration discourse to the maximum number of rounds set to 1e4. a of ${{\lambda }_{1}}$ is set to 0.25 and of ${{\lambda }_{3}}$ to 0.025. The batch size is set to 32 while Digits task is set to 64, total epoch set to 20, the random seed set to 42, the factor of label smoothing set to 0.1. During CMKD training, the prompt we applied in most benchmarks is ``a photo of [CLASS]'', while in digits benchmark, the prompt is designed as ``a photo of the number: [CLASS]''. When utilizing RST, the value of $\tau$  is set to 1e-6 and 1e-5 for ViT and ResNet respectively. All the experiments are conducted with 1x NVIDIA GTX 3090 with 24G RAM and the PyTorch framework is implemented to perform our experiments. For a more comprehensive implementation, please refer to Appendix.A.

\textbf{Evaluation}. In addition to the commonly used Accuracy metrics for UDA, we propose the inclusion of \textbf{D}own\textbf{S}tream \textbf{P}arameters (\textbf{DSP}) as a measure to assess the model's storage overhead during deployment for downstream tasks. The calculation of DSP involves multiplying the number of migration tasks by the total number of parameters to be stored. To illustrate, let's consider the Office-Home benchmark as an example. In the case of training the ViT-B model with all parameters, the visual encoder comprises 86.2 million parameters, the task header includes 0.03 million parameters, and the downstream task model contains 86.23 million parameters. Given that there are 12 sub-tasks on Office-Home, the corresponding DSP equals 1034.80 ($86.23 * 12\approx1034.80$). When applying RST, we calculate the corresponding DSP by sum up the number of non-zero elements in the visual encoder and the classifier parameters. For the other PEFT methods, the corresponding DSP is calculated by sum up the number of additional parameters and classifier parameters. See Appendix.C for details on the parameters of the backbones.

\subsection{Comparison with SoTA UDA Methods}
We compared our approach with the state-of-the-art unsupervised domain adaptation (UDA) method on both ResNet and Transformer models to validate its effectiveness and flexibility. Additionally, we discovered that combining CMKD with FixMatch \cite{ref69} can yield further performance improvements, and the implementation details are described in Appendix.B. Moreover, when CMKD is integrated with RST, it significantly reduces the number of parameters required for downstream tasks while only experiencing a slight decrease in performance. 

\textbf{Results on Office-Home}. Table I shows the quantitative results of methods. As expected, CMKD achieves comparable performance \textbf{(88.4\%)} with SoTA PMTrans \textbf{(88.9\%)}. By incorporating CMKD with FixMatch, our approach surpasses PMTrans and emerges as the new SoTA \textbf{(89.0\%)}. Our method demonstrates remarkable results on the CNN architecture as well, surpassing DAPL by a significant margin of \textbf{+5.8\%} and outperforming zero-shot performance of CLIP with a large gap of \textbf{+8.3\%} in terms of average accuracy. Although the DSP of CMKD significantly surpasses that of CLIP and DAPL, when integrated with RST, a mere \textbf{1.23M} parameters count is sufficient to yield remarkable outcomes across 12 downstream tasks.

\textbf{Results on Office-31}. Table II shows the quantitative comparison with various UDA algorithms. The adaptability of CMKD and the parameter-efficient learning capability of RST have been duly confirmed. In the case of CMKD, it achieved comparable performance to the SoTA on both CNN and Transformer architectures, with average accuracy of \textbf{90.9\%} and \textbf{93.4\%} respectively. With the utilization of RST (\textbf{90.8\%}, \textbf{93.2\%}), a mere \textbf{0.31M} and \textbf{0.29M} parameters are needed to uphold comparable performance to full parametric training.

\begin{table}[t]
	\setlength{\abovecaptionskip}{0cm}  
	\setlength{\belowcaptionskip}{-0.2cm} 
	\centering
	\caption{Comparison with SoTA UDA methods on Office-31.}
	\renewcommand{\arraystretch}{0.9}
	\setlength{\tabcolsep}{0.35mm}{
		\begin{tabular}{c|cccccccc}
			\toprule
			Method & A→D & A→W & D→A & D→W & W→A & W→D & Avg. & DSP(M) \\
			\midrule
			\makecell[l]{\textit{ResNet50:}} &       &       &       &       &       &       &  \\
			CDAN+E & 92.9  & 94.1  & 71.0  & 98.6  & 69.3  & \textbf{100.0}  & 87.7  & 141.40\\
			rRGrad+CAT & 90.8  & 94.4  & 72.2  & 98.0  & 70.2  & \textbf{100.0}  & 87.6  & 141.40\\
			CDAN+BSP & 93.0  & 93.3  & 73.6  & 98.2  & 72.6  & \textbf{100.0}  & 88.5  & 141.40\\
			CDAN+TN & 94.0  & \textbf{95.7}  & 73.4  & 98.7  & 74.2  & \textbf{100.0}  & 89.3 & 141.40\\
			DSAN  & 90.2  & 93.6  & 73.5  & 98.3  & 74.8  & \textbf{100.0}  & 88.4  & 141.40\\
			BNM   & 90.3  & 91.5  & 70.9  & 98.5  & 71.6  & \textbf{100.0}  & 86.8  & 141.40\\
			SHOT  & 94.0  & 90.1  & 74.7  & 98.7  & 74.3  & 99.8  & 88.6  & 141.40\\
			\midrule
			CLIP$^\star$  & 74.3  & 67.8  & 72.6  & 67.8  & 72.6  & 74.3  & 71.6  & \textbf{0.00}\\
			Baseline$^\star$ & 85.1  & 77.1  & 70.1  & 97.1  & 67.7  & 99.6  & 82.8  & 229.99\\
			CMKD$^\star$  & 95.0  & 91.8  & 79.9  & 98.6  & 80.5  & \textbf{100.0}  & 90.9  & 229.99\\
			CMKD+RST$^\star$ & 95.0  & 90.9  & 80.7  & 97.4  & 81.0  & \textbf{100.0}  & 90.8  & \textbf{0.31}\\
			CMKD+FixMatch$^\star$ & \textbf{95.8}  & 93.6  & \textbf{81.3}  & \textbf{99.1}  & \textbf{81.6}  & \textbf{100.0}  & \textbf{91.9}  & 229.99\\
			\midrule
			\midrule
			\makecell[l]{\textit{ViT-B-16:}} &       &       &       &       &       &       &  \\
			CDTrans & 97.0  & 96.7  & 81.1  & 99.0  & 81.9  & \textbf{100.0}  & 92.6  & 514.94\\
			TVT$^o$  & 96.4  & 96.4  & 84.9  & 99.4  & 86.1  & \textbf{100.0}  & 93.8  & 514.94\\
			SSRT$^o$ & 98.6  & 97.7  & 83.5  & 99.2  & 82.2  & \textbf{100.0}  & 93.5  & 514.94\\
			PMTrans$^o$ & \textbf{99.4}  &\textbf{ 99.1}  & \textbf{85.7}  & \textbf{99.6}  & \textbf{86.3}  & \textbf{100.0}  & \textbf{95.0}  & 514.94\\
			\midrule
			CLIP$^\star$  & 79.9  & 76.9  & 78.9  & 76.9  & 78.9  & 79.9  & 78.6  & \textbf{0.00}\\
			Baseline$^\star$ & 87.3    &85.8       & 76.3      &98.7       & 75.9      & \textbf{100.0}      & 87.3 & 517.30\\
			CMKD$^\star$  & 96.6  & 97.4  & 84.4  & 99.1  & 84.5  & \textbf{100.0}  & 93.4  & 517.30\\
			CMKD+RST$^\star$ & 95.2  & 96.9  & 84.0  & 99.1  & 84.2  & \textbf{100.0}  & 93.2  & \textbf{0.29}\\
			CMKD+FixMatch$^\star$ & 98.0  & 98.4  & 85.3  & 99.4  & 85.4  & \textbf{100.0}  & 94.4  & 517.30\\
			\bottomrule
		\end{tabular}%
	}
	\label{tab2}%
\end{table}%

\textbf{Results on Digits}. In this benchmark, the previous methods have all employed LeNet-5 as the backbone and achieved favorable outcomes for digits adaptation tasks. Intriguingly, as shown in Table III, despite undergoing large-scale visual-language pre-training, the zero-shot inference of CLIP falls significantly short of the previous methods and even lags behind the baseline performance. Interestingly, when employing CMKD based on ViT-B, the model attains state-of-the-art performance (\textbf{98.3\%}). Moreover, when combined with FixMatch, the performance is further enhanced to \textbf{98.5\%}, which represents a remarkable improvement of \textbf{+40.5\%} compared to CLIP. This clearly demonstrates the superiority of CMKD, even in cases where the guidance from the teacher's perspective is insufficient. CMKD solely considers the supervised signals from the teacher as a reference rather than a learning target.

\begin{table}[t]
	\setlength{\abovecaptionskip}{0.cm}
	\setlength{\belowcaptionskip}{-0.cm}
	\centering
	\caption{Comparison with SoTA UDA methods on Digits.}
	\renewcommand{\arraystretch}{0.9}
	\setlength{\tabcolsep}{2.0mm}{
		\begin{tabular}{c|ccccc}
			\toprule
			Method & S→M & U→M & M→U & Avg. & DSP(M)\\
			\midrule
			\makecell[l]{\textit{LetNet5:}} &       &       &       &  \\
			ADDA  & 76.0  & 90.1  & 89.4  & 85.2  & \textbf{0.03}\\
			ADR   & 95.0  & 93.1  & 93.2  & 93.8  &\textbf{0.03}\\
			CDAN+E & 89.2  & 98.0  & 95.6  & 94.3  &\textbf{0.03}\\
			CyCADA  & 90.4  & 96.5  & 95.6  & 94.2  &\textbf{0.03}\\
			rRGrad+CAT & 98.8  & 96.0  & 94.0  & 96.3  &\textbf{0.03}\\
			DSAN  & 90.1  & 95.3  & 96.9  & 94.1  &\textbf{0.03}\\
			SWD   & 98.9  & 97.1  & \textbf{98.1} & 98.0  &\textbf{0.03}\\
			SHOT  & \textbf{98.9} & \textbf{98.0} & 97.9  & \textbf{98.3} &\textbf{0.03}\\
			\midrule
			\midrule
			\makecell[l]{\textit{ResNet50:}} &       &       &       &  \\
			CLIP$^\star$  & 58.4  & 58.4  & 47.6  & 54.8 &\textbf{0.00} \\
			Baseline$^\star$ & 81.8  & 93.1  & 94.6  & 89.8 &114.93\\
			CMKD$^\star$  & 96.4 & \textbf{97.8} & \textbf{97.1} & \textbf{97.1} &114.93\\
			CMKD+RST$^\star$ &  95.6     &   97.1    & 95.7      & 96.1 & \textbf{0.23}\\
			CMKD+Fixmatch$^\star$ & 97.0      &  97.2     &   96.1    &  96.8 &114.93\\
			\midrule
			\midrule
			\makecell[l]{\textit{ViT-B-16:}} &       &       &       &  \\
			CLIP$^\star$  & 55.5  & 55.5  & 62.9  & 58.0  & \textbf{0.00}\\
			Baseline$^\star$ & 85.7  & 96.8  & \textbf{97.6} & 93.4  & 258.62\\
			CMKD$^\star$  & 98.3  & 98.9  & 97.8  & 98.3  & 258.62\\
			CMKD+RST$^\star$ & 97.3  & 96.8  & 96.1  & 96.7  & \textbf{0.72}\\
			CMKD+Fixmatch$^\star$ & \textbf{98.6} & \textbf{99.1} & \textbf{97.8} & \textbf{98.5} & 258.62\\
			\bottomrule
		\end{tabular}%
	}
	\label{tab3}%
\end{table}%

\textbf{Results on ImageCLEF-DA}. Table IV shows the quantitative results with the CNN-based and ViT-based methods. Overall, CMKD achieves the best performance on each task with \textbf{94.3\%} accuracy and outperforms the SoTA methods. Numerically, under the ResNet50, CMKD surpasses the SoTA methods with an increase of \textbf{+0.8\%} accuracy over CLIP, \textbf{+1.5\%} accuracy over DSAN, respectively.

\begin{table}[t]
	\setlength{\abovecaptionskip}{0.cm}
	\setlength{\belowcaptionskip}{-0.cm}
	\centering
	\caption{Comparison with SoTA UDA methods on ImageCLEF-DA.}
	\renewcommand{\arraystretch}{0.9}
	\setlength{\tabcolsep}{0.8mm}{
		\begin{tabular}{c|cccccccc}
			\toprule
			Method &I→P & I→C & P→I & P→C & C→I & C→P & Avg. &DSP(M) \\
			\midrule
			\makecell[l]{\textit{ResNet50:}} &       &       &       &       &       &       &  \\
			DAN   & 75.0  & 93.3  & 86.2  & 91.3  & 84.1  & 69.8  & 83.3  &141.14\\
			DANN  & 75.0  & 96.2  & 86.0  & 91.5  & 87.0  & 74.3  & 85.0  &141.14\\
			D-CORAL  & 76.9  & 93.6  & 88.5  & 91.6  & 86.8  & 74.0  & 85.2  &141.14\\
			MADA  & 75.0  & 96.0  & 87.9  & 92.2  & 88.8  & 75.2  & 85.8  &141.14\\
			CDAN+E & 77.7  & 97.7 & 90.7  & 94.3  & 91.3  & 74.2  & 87.7  &141.14\\
			DSAN  & 80.2 & 97.2  & 93.3  & 95.9  & 93.8  & 80.8 & 90.2 &141.14\\
			\midrule
			CLIP$^\star$  & 81.3  & 96.5  & 94.8 & 96.5 & 94.8 & 81.3  & 90.9  &\textbf{0.00}\\
			Baseline$^\star$ & 78.2  & 95.8  & 90.3 & 93.5 & 91.0 & 75.8  & 87.4  &229.87\\
			CMKD$^\star$ & 81.5  & 97.8  & 94.5 & \textbf{98.6} & 96.0 & \textbf{81.5}  & 91.7  &229.87\\
			CMKD+RST$^\star$ & \textbf{83.3}   &  \textbf{98.8}     & \textbf{ 97.0}     & 97.0      &  \textbf{96.7}     & 80.5      & \textbf{92.2} &  \textbf{0.21}\\
			CMKD+Fixmatch$^\star$ &   81.0    &  98.7            &  95.8     &  97.8     &   95.7 & 81.3 & 91.7 &  229.87\\
			\midrule
			\midrule
			\makecell[l]{\textit{ViT-B-16:}}  &       &       &       &       &       &       &  \\
			CLIP$^\star$  & 80.7  & 93.3  & 96.3  & 93.3  & 96.3  & 80.7  & 90.1  & \textbf{0.00}\\
			Baseline$^\star$ & 82.3  & 97.5  & 93.8  & 94.5  & 90.5  & 75.6  & 89.0  &517.23\\
			CMKD$^\star$  & 85.3  & 99.5  & 98.0  & 99.2  & \textbf{98.3}  & \textbf{85.0}  & 94.2  &517.23\\
			CMKD+RST$^\star$ & 85.0  & 99.5  & 98.0  & 99.0  & 98.2  & 83.8  & 93.9  &\textbf{0.08}\\
			CMKD+Fixmatch$^\star$ & \textbf{85.5} & \textbf{99.7} & \textbf{98.5} & \textbf{99.3} & 98.0 & \textbf{85.0} & \textbf{94.3} &517.23\\
			\bottomrule
		\end{tabular}%
	}
	\label{tab4}%
\end{table}%

\textbf{Results on VisDA-2017}. As shown in Table V, due to the similar distribution between the pre-training data of CLIP and the target domain (real) of VisDA, CLIP's zero-shot inference capability under the ViT-B architecture reached \textbf{88.0\%}, significantly surpassing the baseline (\textbf{78.6\%}). In contrast to the digits benchmark, representing a scenario where teacher vision is superior, CMKD enables the student perspective to achieve further performance improvement beyond the limits set by the teacher vision. It achieves a performance of \textbf{90.3\%}, outperforming the baseline and CLIP by \textbf{+11.5\%} and \textbf{+2.3\%}, respectively. When combined with FixMatch, the performance even reached \textbf{91.8\%}. The validity is also demonstrated in the CNN structure, which achieved a score of \textbf{87.8\%}, surpassing the baseline and CLIP by \textbf{+21.3\%} and \textbf{+4.4\%} respectively. When CMKD is combined with RST, only need \textbf{0.06 M} parameters can reach \textbf{89.1\%}, achieving approximate results with SDAT (\textbf{89.9\%}) while the latter requires \textbf{85.81 M}.

\begin{table*}[t]
	\setlength{\abovecaptionskip}{0.cm}
	\setlength{\belowcaptionskip}{-0.cm}
	\centering
	\caption{Comparison with SoTA UDA methods on VisDA-2017.}
	\renewcommand{\arraystretch}{0.9}
	\setlength{\tabcolsep}{1.8mm}{
		\begin{tabular}{c|cccccccccccccc}
			\toprule
			Method & plane & bcycl & bus & car & horse & knife & mcycl & person & plant & sktbrd & train & truck & Avg. & DSP(M)\\
			\midrule
			\makecell[l]{\textit{ResNet101:}} &       &       &       &       &       &       &       &       &       &       &       &       &  \\
			CDAN+E & 85.2  & 66.9  & 83.0  & 50.8  & 84.2  & 74.9  & 88.1  & 74.5  & 83.4  & 76.0  & 81.9  & 38.0  & 73.9  & 42.52\\
			DSAN  & 90.9  & 66.9  & 75.7  & 62.4  & 88.9  & 77.0  & 93.7  & 75.1  & 92.8  & 67.6  & 89.1  & 39.4  & 75.1  & 42.52\\
			BNM   & 89.6  & 61.5  & 76.9  & 55.0  & 89.3  & 69.1  & 81.3  & 65.5  & 90.0  & 47.3  & 89.1  & 30.1  & 70.4  & 42.52\\
			MSTN+DSBN & 94.7  & 86.7  & 76.0  & 72.0  & 95.2  & 75.1  & 87.9  & 81.3  & 91.1  & 68.9  & 88.3  & 45.5  & 80.2  & 42.52\\
			CGDM  & 92.8  & 85.1  & 76.3  & 64.5  & 91.0  & 93.2  & 81.3  & 79.3  & 92.4  & 83.0  & 85.6  & 44.8  & 80.8  & 42.52\\
			SHOT  & 95.5  & 87.5  & 80.1  & 54.5  & 93.6  & 94.2  & 80.2  & 80.9  & 90.0  & 89.9  & 87.1  & 58.4  & 82.7  & 42.52\\
			CDAN+MCC & 94.5  & 80.8  & 78.4  & 65.3  & 90.6  & 79.4  & 87.5  & \textbf{82.2}  &\textbf{94.7}  & 81.0  & 86.0  & 44.6  & 80.4  & 42.52\\
			DAPL$^\star$  & 97.8  & 83.1  & 88.8  & 77.9  & 97.4  & 91.5  & 94.2  & 79.7  & 88.6  & 89.3  & 92.5  & 62.0  & 86.9  & \textbf{0.00}\\
			\midrule
			CLIP$^\star$  & \textbf{98.2}  & 83.9  & \textbf{90.5}  & 73.5  & 97.2  & 84.0  & \textbf{95.3} & 65.7  & 79.4  & 89.9  & 91.8  & 63.3  & 84.4  & \textbf{0.00}\\
			Baseline$^\star$ & 92.6  & 39.1  & 73.3  & 78.5  & 83.1  & 54.1  & 90.6  & 29.0  & 78.8  & 67.3  & \textbf{97.5}  & 14.4  & 66.5  & 56.31\\
			CMKD$^\star$  & 97.8  & 87.6  & 87.8  & 75.4  & 96.8  & 95.7  & 87.4  & 79.3  & 91.2  & 89.8  & 92.0  & \textbf{63.4}  & 87.0  &56.31\\
			CMKD+RST$^\star$ & 97.5  & 86.1  & 87.1  & 78.1  & 96.8  & 95.1  & 90.2  & \textbf{82.2}  & 87.3  & 91.8  & 94.5  & 53.0  & 86.6  &\textbf{0.03}\\
			CMKD+FixMatch$^\star$ & \textbf{98.2}  & \textbf{88.0}  & 89.1  & \textbf{79.3}  & \textbf{98.3}  & \textbf{96.5}  & 92.8  & 78.8  & 91.9  & \textbf{94.7}  & 94.7  & 50.9  & \textbf{87.8}  & 56.31\\
			\midrule
			\midrule
			\makecell[l]{\textit{ViT-B-16:}} &       &       &       &       &       &       &       &       &       &       &       &       &  \\
			CDTrans & 97.1  & 90.5  & 82.4  & 77.5  & 96.6  & 96.1  & 93.6  & \textbf{88.6}  & \textbf{97.9}  & 86.9  & 90.3  & 62.8  & 88.4  &85.81\\
			TVT$^o$   & 92.9  & 85.6  & 77.5  & 60.5  & 93.6  & 98.2  & 89.4  & 76.4  & 93.6  & 92.0  & 91.7  & 55.7  & 83.9  &85.81\\
			SDAT$^o$  & 98.4  & 90.9  & 85.4  & 82.1  & 98.5  & 97.6  & \textbf{96.3}  & 86.1  & 96.2  & 96.7  & 92.9  & 56.8  & 89.8  &85.81\\
			SSRT-B$^o$ & 98.9  & 87.6  & 89.1  & \textbf{84.7}  & 98.3  & \textbf{98.7}  & 96.2  & 81.0  & 94.8  & 97.9  & 94.5  & 43.1  & 88.7  &85.81\\
			PMTrans$^o$ & 98.9  & 93.7  & 84.5  & 73.3  & 99.0  & 98.0  & 96.2  & 67.8  & 94.2  & 98.4  & 96.6  & 49.0  & 87.5 &85.81 \\
			\midrule
			CLIP$^\star$  & 99.3  & 91.7  & 93.9  & 74.3  & 98.4  & 94.3  & 90.3  & 78.2  & 78.3  & 97.3  & 95.2  & 64.8  & 88.0  &\textbf{0.00}\\
			Baseline$^\star$ & \textbf{99.5}  & 83.0  & 84.4  & 76.3  & 97.4  & 78.1  & 95.9  & 20.8  & 97.2  & 94.0  & \textbf{98.6}  & 17.6  & 78.6  &86.21\\
			CMKD$^\star$  & 99.4  & 94.6  & \textbf{91.5}  & 78.9  & 98.7  & 97.3  & 93.3  & 81.3  & 91.8  & 97.9  & 96.9  & 61.7  & 90.3  &86.21\\
			CMKD+RST$^\star$ & 99.3  & 91.3  & 91.0  & 77.7  & 99.0  & 96.8  & 94.0  & 81.8  & 91.1  & 97.1  & 96.8  & 52.8  & 89.1  &\textbf{0.06}\\
			CMKD+FixMatch$^\star$ & \textbf{99.5}  & \textbf{95.2}  & 91.3  & 81.5  & \textbf{99.4}  & 98.0  & 95.2  & 83.5  & 95.5  & \textbf{98.5}  & 96.8  & \textbf{67.0}  & \textbf{91.8}  &86.21\\
			\bottomrule
		\end{tabular}%
	}
	\label{tab5}%
\end{table*}%

\textbf{Results on DomainNet}. CMKD and CMKD+FixMatch emerged as the new state-of-the-art models, achieving respective accuracies of \textbf{53.9\%} and \textbf{54.3\%}, surpassing the previous SoTA model, PMTrans, which achieved \textbf{52.4\%}, as shown in Table VI. While CLIP does not rely on downstream task parameters, its average accuracy of \textbf{6.6\%} falls significantly behind other methods. By combining CMKD with RST, a mere \textbf{15.71M} DSP is needed to achieve a remarkable accuracy of \textbf{52.2\%}, yielding results on par with PMTrans.

\textbf{Analysis}. The experimental results provide several insightful observations. Firstly, despite the extensive visual-language pre-training of CLIP, its zero-shot performance still significantly lags behind that of previous SoTA. Secondly, CMKD demonstrates remarkable performance, achieving SoTA on five benchmarks when combined with FixMatch. This indicates that future research could leverage CMKD as a baseline to explore more effective techniques for enhancing the potential of VLP models on UDA tasks. Thirdly, both CMKD and RST exhibit great flexibility and applicability across ResNet and ViT structures. Fourthly, despite a slight performance degradation with RST, the deployment cost, as measured by DSP, drops by more than 99\%, demonstrating its efficiency in reducing resource requirements. Fifthly, regardless of the sufficiency of guidelines from the teacher model (e.g., based on the ViT-B model, CLIP's zero-shot inference result on Office-Home is 82.4\%, while it is only 6.6\% on DomainNet), CMKD effectively balances the perspectives of both the student and the teacher, leading to improved model performance.

\begin{table*}[t]
	\setlength{\abovecaptionskip}{0.cm}
	\setlength{\belowcaptionskip}{-0.cm}
	\centering
	\renewcommand{\arraystretch}{0.8}
	\caption{Comparison with SoTA UDA methods on DomainNet.}
	\setlength{\tabcolsep}{0.8mm}{
		\begin{tabular}{c|ccccccc||c|ccccccc||c|ccccccc}
			\toprule
			SVD  & clp & inf & pnt & qdr & rel & skt &Avg. & BNM  & clp & inf & pnt & qdr & rel & skt &Avg. & CGDM& clp & inf & pnt & qdr & rel & skt &Avg. \\
			\midrule
			\midrule
			clp   & - & 14.7    & 31.9    & 10.1    & 45.3    & 36.5    & 27.7  & clp   & - & 12.1    & 33.1    & 6.2   & 50.8    & 40.2    & 28.5          & clp   & -     & 16.9     & 35.3   & 10.8    & 53.5    & 36.9    & 30.7 \\
			inf   & 22.9    & - & 24.2    & 2.5   & 33.2    & 21.3    & 20.0    & inf   & 26.6    & - & 28.5    & 2.4   & 38.5    & 18.1    & 22.8          & inf   & 27.8    & -     & 28.2    & 4.4   & 48.2    & 22.5    & 26.2 \\
			pnt   & 33.6    & 1530    & - & 4.4   & 46.1    & 30.7    & 26.0    & pnt   & 39.9    & 12.2    & - & 3.4   & 54.5    & 36.2    & 29.2          & pnt   & 37.7    & 14.5    & -     & 4.6   & 59.4    & 33.5    & 30.0 \\
			qdr   & 15.5    & 2.2   & 6.4   & - & 11.1    & 10.2    & 9.1   & qdr   & 17.8    & 1.0     & 3.6   & - & 9.2   & 8.3   & 8.0                   & qdr   & 14.9    & 1.5     & 6.2   & -     & 10.9    & 10.2    & 8.7 \\
			rel   & 41.2    & 18.1    & 44.0    & 4.6   & - & 31.6   & 27.9  & rel   & 48.6    & 13.2    & 49.7    & 3.6   & - & 33.9    & 29.8             & rel   & 49.4    & 20.8    & 47.2   & 4.8   & -     & 38.2    & 32.0 \\
			skt   & 44.2    & 15.2    & 37.0    & 10.3    & 44.7    & - & 30.3  & skt   & 54.9    & 12.8    & 42.3    & 5.4   & 51.3    & - & 33.3          & skt   & 50.1    & 16.5    & 43.7    & 11.1    & 55.6    & -     & 35.4 \\
			Avg   & 31.5    & 13.1   & 29.0    & 6.4   & 36.1    & 26.1    & 23.6 & Avg   & 37.6    & 10.3    & 31.4    & 4.2   & 40.9    & 27.3    & 25.3  & Avg    & 36.0    & 14.0    & 32.1    & 7.1   & 45.5    & 28.3    & 27.2 \\
			DSP(M) & \multicolumn{7}{c||}{726.20} & DSP(M) & \multicolumn{7}{c||}{726.20} & DSP(M)&  \multicolumn{7}{c}{726.20}\\
			\midrule
			\midrule
			MDD   & clp & inf & pnt & qdr & rel & skt &Avg. & SCDA  & clp & inf & pnt & qdr & rel & skt &Avg. & CDTrans& clp & inf & pnt & qdr & rel & skt &Avg. \\
			\midrule
			\midrule
			clp   &-	&20.5	&40.7	&6.2	&52.5	&42.1	&32.4  & clp  & -	&18.6	&39.3	&5.1	&55.0	&44.1	&32.4  & clp   &-	&29.4	&57.2	&26.0	&72.6	&58.1	&48.7 \\
			inf   &33.0	 &-	 &33.8	&2.6	&46.2	&24.5	&28.0    & inf   &29.6	&-	&34.0	&1.4	&46.3	&25.4	&27.3  & inf   &57.0	&-	&54.4	&12.8	&69.5	&48.4	&48.4 \\
			pnt   &43.7	 &20.4	&-	&2.8	&51.2	&41.7	&32.0    & pnt   & 44.1	&19.0	&-	& 2.6	&56.2	&42.0	&32.8    & pnt   &62.9	&27.4	&-	&15.8	&72.1	&53.9	&46.4 \\
			qdr   &18.4	 &3.0	&8.1	&-	&12.9	&11.8	&10.8  & qdr   & 30.0	&4.9	&15.0	&-	&25.4	&19.8	&19.0    & qdr   &44.6	&8.9	&29.0	&-	&42.6	&28.5	&30.7 \\
			rel   &52.8	&21.6	&47.8	&4.2	&-	&41.2	&33.5  & rel   & 54.0	&22.5	&51.9	&2.3	&-	&42.5	&34.6  & rel   &66.2	&31.0	&61.5	&16.2	&-	&52.9	&45.6 \\
			skt   &54.3	&17.5	&43.1	&5.7	&54.2	&-	&35.0    & skt   &55.6	&18.5	&44.7	&6.4	&53.2	&-	&35.7  & skt   &69.0	&29.6	&59.0	&27.2	&72.5	&-	&51.5 \\
			Avg   &40.4	&16.6	&34.7	&4.3	&43.4	&32.3	&28.6 & Avg  &42.6	&16.7	&37.0	&3.6	&47.2	&34.8	&30.3 & Avg   &59.0	&25.3	&52.2	&19.6	&65.9	&48.4	&45.2 \\
			DSP(M) & \multicolumn{7}{c||}{726.20} & DSP(M) & \multicolumn{7}{c||}{726.20} & DSP(M)&  \multicolumn{7}{c}{2581.94}\\
			\midrule
			\midrule
			SSRT$^o$  & clp & inf & pnt & qdr & rel & skt &Avg. & PMTRans$^o$ & clp & inf & pnt & qdr & rel & skt &Avg. & CLIP$^\star$  & clp & inf & pnt & qdr & rel & skt &Avg. \\
			\midrule
			\midrule
			clp   &-	&33.8	&60.2	&19.4	&75.8	&59.8	&49.8  & clp   &-	&34.2	&62.7	&\textbf{32.5}	&\textbf{79.3}	&63.7	&54.5  & clp   & - & 9.6   & 5.4   & 1.7   & 8.0     & 6.9   & 6.3 \\
			inf  &55.5	&-	&54.0	&9.0	&68.2	&44.7	&46.3  & inf   &67.4	&-	&61.1	&\textbf{22.2}	&\textbf{78.0}	&57.6	&57.3  & inf   & 9.8   & - & 5.4   & 1.7   & 8.0     & 6.9   & 6.3 \\
			pnt   &61.7	&28.5	&-	&8.4	&71.4	&55.2	&45.0    & pnt  &69.7	&33.5	&-	&\textbf{23.9}	&\textbf{79.8}	&61.2	&53.6  & pnt   & 9.8   & 9.6   & - & 1.7   & 8.0     & 6.9   & 6.3 \\
			qdr   &42.5	&8.8	&24.2	&-	&37.6	&33.6	&29.3  & qdr   &54.6	&17.4	&\textbf{38.9}	&-	&\textbf{49.5}	&41.0	&40.3  & qdr   & 9.8   & 9.6   & 5.4   & - & 8.0     & 6.9   & 6.3 \\
			rel   &69.9	&37.1	&66.0	&10.1	&-	&58.9	&48.4  & rel  &74.1	&35.3	&\textbf{70.0}	&\textbf{25.4}	&-	&61.1	&53.2  & rel   & 9.8   & 9.6   & 5.4   & 1.7   & - & 6.9   & 6.3 \\
			skt   &70.6	&32.8	&62.2	&21.7	&73.2	&-	&52.1  & skt  &73.8	&33.0	&62.6	&\textbf{30.9}	&77.5	&-	&55.6  & skt   & 9.8   & 9.6   & 5.4   & 1.7   & 8.0     & - & 6.3 \\
			Avg  &60.0	&28.2	&53.3	&13.7	&65.3	&50.4	&45.2  & Avg   &67.9	&30.7	&59.1	&\textbf{27.0}	&\textbf{72.8}	&56.9	&52.4 & Avg   & 9.8   & 9.6   & 5.4   & 1.7   & 8.0     & 6.9   & 6.6 \\
			DSP(M) & \multicolumn{7}{c||}{2581.94} & DSP(M) & \multicolumn{7}{c||}{2581.94} & DSP(M)&  \multicolumn{7}{c}{\textbf{0.00}}\\
			\midrule
			\midrule
			CMKD$^\star$ & clp & inf & pnt & qdr & rel & skt &Avg. & +RST$^\star$  & clp & inf & pnt & qdr & rel & skt &Avg. & +FixMatch$^\star$ & clp & inf & pnt & qdr & rel & skt &Avg. \\
			\midrule
			\midrule
			clp   & -	&43.5	&65.9	&22.8	&78.2	&66.7	&55.4  & clp   & -	&43.2	&64.3	&20.7	&77.5	&65.5	&54.2  & clp   & -	&\textbf{44.3}	&\textbf{66.2}	&22.6	&78.4	&\textbf{66.9}	&\textbf{55.7} \\
			inf   & 71.3	&-	&64.5	&19.8	&78.0	&61.0	&58.9  & inf   &70.6	&-	&62.9	&17.7	&77.2	&60.3   &  57.7  & inf   & \textbf{71.6}	&-	&\textbf{64.7}	&18.5	&\textbf{78.0}	&\textbf{61.0}	&\textbf{58.8} \\
			pnt   & \textbf{70.6}	&42.3	&-	&19.1	&76.8	&65.0	&54.8  & pnt   & 68.4	&40.5	&-	&18.1	&75.9	&63.8    &  53.3     & pnt   & 70.5	&\textbf{43.4}	&-	&19.5	&77.0	&\textbf{65.3}	&\textbf{55.1} \\
			qdr   & 50.3	&27.2	&37.1	&-	&46.4	&41.5	&40.5  & qdr   & 48.5      & 22.5      &  34.3     & - & 41.2      & 37.6      &   36.8  & qdr   & \textbf{51.6}	&\textbf{29.3}	&38.0	&-	&46.5	&\textbf{43.3}	&\textbf{41.7} \\
			rel   & 77.0	&48.0	&69.2	&18.5	&-	&66.0	&55.7  & rel   &75.8       &47.2       &  67.9     & 16.8      & - & 64.8      &   54.5    & rel   & \textbf{77.1}	&\textbf{48.7}	&69.6	&19.1	&-	&\textbf{66.2}	&\textbf{56.1} \\
			skt   & 76.1	&44.7	&67.5	&22.9	&78.0	&-	&57.8  & skt   & 74.8      & 43.1      &    65.9   &   21.6    &      77.4 & - &  56.6     & skt   & \textbf{76.2}	&\textbf{45.2}	&\textbf{68.1}	&23.0	&\textbf{78.9}	&-	&\textbf{58.3} \\
			Avg   & 69.1	&41.1	&60.8	&20.6	&71.5	&60.0	&53.9 & Avg   &  67.7     & 39.3      &  59.1     & 19.0      & 69.8      &  58.4     &  52.2 & Avg   & \textbf{69.4}	&\textbf{42.2}	&\textbf{61.3}	&20.5	&71.8	&\textbf{60.5}	&\textbf{54.3} \\
			DSP(M) & \multicolumn{7}{c||}{2591.30} & DSP(M) & \multicolumn{7}{c||}{\textbf{15.71}} & DSP(M)&  \multicolumn{7}{c}{2591.30}\\
			\bottomrule
		\end{tabular}%
	}
	\label{tab:addlabel}%
\end{table*}%

\begin{table*}[htbp]
	\setlength{\abovecaptionskip}{0cm}  
	\setlength{\belowcaptionskip}{-0.2cm} 
	\centering
	\caption{Comparison with SoTA PEFT methods on Office-Home.}
	\renewcommand{\arraystretch}{0.95}
	\setlength{\tabcolsep}{0.5mm}{
		\begin{tabular}{p{2cm}|cccccccccccccc}
			\toprule
			\centering Method & {Ar→Cl} & {Ar→Pr} & {Ar→Re} & {Cl→Ar} & {Cl→Pr} & {Cl→Re} & {Pr→Ar} & {Pr→Cl} & {Pr→Re} & {Re→Ar} & {Re→Cl} & {Re→Pr} & {Avg.} & {DSP(M)}\\
			\midrule
			\makecell[l]{\textit{ResNet50:}} &       &       &       &       &       &       &       &       &       &       &       &       & &  \\
			\centering \textcolor{gray}{Fine-Tuning$^\star$}   & \textcolor{gray}{65.9}      & \textcolor{gray}{86.6}       & \textcolor{gray}{87.3}      & \textcolor{gray}{74.4}   &  \textcolor{gray}{87.7}    & \textcolor{gray}{85.8}      & \textcolor{gray}{75.9}       & \textcolor{gray}{64.4}      &  \textcolor{gray}{87.9}    & \textcolor{gray}{79.1}      & \textcolor{gray}{67.2}      & \textcolor{gray}{90.0}     &\textcolor{gray}{79.3}   & \textcolor{gray}{460.40}\\
			\centering \textcolor{gray}{Linear-Probe$^\star$} & \textcolor{gray}{59.2}      & \textcolor{gray}{86.3}      &  \textcolor{gray}{87.1}     & \textcolor{gray}{73.7}      & \textcolor{gray}{87.6}      & \textcolor{gray}{85.4}      & \textcolor{gray}{72.4}      & \textcolor{gray}{58.8}      & \textcolor{gray}{87.2}      & \textcolor{gray}{75.6}      & \textcolor{gray}{60.0}      & \textcolor{gray}{89.5}       & \textcolor{gray}{76.9}  & \textcolor{gray}{0.80}\\
			\centering $\text{LoRA}_{r=1}^\star$  &59.1	&85.2   &87.1   &72.5   & 86.2       & 84.6      & 73.0      &60.7    & 87.1      &76.5     &61.1   & 88.7 & 76.8 & 1.76\\
			\centering $\text{LoRA}_{r=4}^\star$  &60.6	&86.1	&\textbf{87.4}	&72.5	&87.2	&\textbf{85.4}	&73.2	&60.8	&87.8	&75.9	&61.6	&89.4	&77.3	&4.04\\
			\centering BitFit$^\star$  & 60.1       & 86.2       &  87.1     &  73.8     & 87.8      & 85.1       &  73.1     & 60.4      & 87.4      &76.5       & 61.4 & \textbf{89.6}       &77.4  & \textbf{1.12} \\
			\centering RST$^\star$  & \textbf{64.9}       & \textbf{87.3}      & 87.1      & \textbf{74.2}      & \textbf{87.5}      &\textbf{85.4}       & \textbf{74.5}      & \textbf{63.0}      & \textbf{88.0}      &   \textbf{78.1}    &  \textbf{66.5}     &\textbf{89.6}       &\textbf{78.8}  &1.23  \\
			\midrule
			\midrule
			\makecell[l]{\textit{ViT-B-16:}} &       &       &       &       &       &       &       &       &       &       &       &       &  \\
			\centering \textcolor{gray}{Fine-Tuning$^\star$}  & \textcolor{gray}{79.4}  & \textcolor{gray}{94.2}  & \textcolor{gray}{92.7}  & \textcolor{gray}{86.3}  & \textcolor{gray}{93.4}  & \textcolor{gray}{92.2}  & \textcolor{gray}{86.7}  & \textcolor{gray}{79.5}  & \textcolor{gray}{92.1}  & \textcolor{gray}{88.2}  & \textcolor{gray}{81.2}  & \textcolor{gray}{94.5}  & \textcolor{gray}{88.4} & \textcolor{gray}{1034.80}\\
			\centering \textcolor{gray}{Linear-Probe$^\star$} & \textcolor{gray}{76.5}      &\textcolor{gray}{93.0}       & \textcolor{gray}{92.2}      &  \textcolor{gray}{83.9}     & \textcolor{gray}{93.0}       &  \textcolor{gray}{90.5}     & \textcolor{gray}{83.2}      & \textcolor{gray}{76.7}      & \textcolor{gray}{91.9}      & \textcolor{gray}{85.2}       & \textcolor{gray}{78.6}       & \textcolor{gray}{93.5}      & \textcolor{gray}{86.5} & \textcolor{gray}{0.40} \\
			\centering $\text{LoRA}_{r=1}^\star$  &76.1	& 93.0     & 92.1      & 84.3     &92.9       &   91.0    & 83.3      & 76.8      & \textbf{92.0}      &  85.0     &78.6       &93.4      & 86.5  & 3.40\\
			\centering $\text{LoRA}_{r=4}^\star$  &75.7	&93.0	&92.3	&83.8	&\textbf{93.2}	&90.0	&83.8	&77.1	&91.8	&85.1	&78.3	&93.4	&86.5	& 12.0\\
			\centering BitFit$^\star$  &77.4       &93.0       & 92.4      &  84.3     & 93.0      & 91.0      & 83.8      &77.8       &  91.9     & 85.7      &  79.7     & 93.7      & 87.0  &1.40 \\
			\centering $\text{AdaLoRA}_{r=2\text{→}1}^\star$ & 76.2      & 93.0      & 92.3     & 84.5      &93.1       &91.0       &83.6       &  77.1     & \textbf{92.0}      & 85.2      & 78.5      &93.3       & 86.7 &2.56  \\
			\centering $\text{AdaLoRA}_{r=8\text{→}4}^\star$ &  76.4     & 93.1      & 92.2       & 84.2      & 93.0      &  90.6     & 83.6      &  76.7     & 91.9      & 85.4      &  78.8     & 93.4      & 86.6  & 6.09\\
			\centering RST$^\star$ & \textbf{78.9}  & \textbf{93.8}  & \textbf{92.7}  & \textbf{85.3}  & 93.0    & \textbf{91.7}  & \textbf{85.3}  & \textbf{78.4}  & \textbf{92.0}    & \textbf{87.2}  & \textbf{80.3}  & \textbf{94.1}  & \textbf{87.7} & \textbf{1.02}\\
			\bottomrule
		\end{tabular}%
	}
	\label{tabl}%
\end{table*}%
\begin{table*}[htbp]
	\setlength{\abovecaptionskip}{0cm}  
	\setlength{\belowcaptionskip}{-0.2cm} 
	\centering
	\caption{Baseline fine-tuning results with different configurations on Office-Home.}
	\renewcommand{\arraystretch}{0.95}
	\setlength{\tabcolsep}{0.5mm}{
		\begin{tabular}{p{2cm}|ccccccccccccc}
			\toprule
			\centering Method & {Ar→Cl} & {Ar→Pr} & {Ar→Re} & {Cl→Ar} & {Cl→Pr} & {Cl→Re} & {Pr→Ar} & {Pr→Cl} & {Pr→Re} & {Re→Ar} & {Re→Cl} & {Re→Pr} & {Avg.}\\
			\midrule
			\makecell[l]{\textit{ResNet50:}} &       &       &       &       &       &       &       &       &       &       &       &       &  \\
			\centering ImageNet-1K   &  51.0 & 	68.2 & 	74.8 & 	\textbf{54.2} & 	63.6 & 	\textbf{66.8} & 	53.6 & 	45.4 & 	74.6 & 	65.6 & 	53.6 & 	79.3 & 	62.5\\
			\centering CLIP-1$^\star$ & 24.7&	25.1&	36.1&	14.8&	29.4&	25.9&	17.1&	28.2&	38.5&	45.5&	45.3&	59.9&	32.6 \\
			\centering CLIP-2$^\star$ &  \textbf{52.1}&	\textbf{69.7}&	\textbf{75.2}&	52.5&	\textbf{64.3}&	63.9&	\textbf{56.2}&	\textbf{48.4}&	\textbf{74.9}&	\textbf{68.9}&	\textbf{54.4}&	\textbf{82.7}&	\textbf{63.7}\\
			\midrule
			\midrule
			\makecell[l]{\textit{ViT-B-16:}} &       &       &       &       &       &       &       &       &       &       &       &       &  \\
			\centering ImageNet-21K$^o$   & 64.6  &\textbf{86.0}    & \textbf{88.6}      & \textbf{81.4}      & \textbf{87.4}       & \textbf{88.5}       & \textbf{79.2}      & 60.7      & \textbf{88.8}      &  81.6     & 61.4      & 89.3      & 79.8      \\
			\centering CLIP-1$^\star$ & 40.4&	65.5&	75.1&	41.0&	60.0&	62.0&	44.7&	49.3&	73.1&	67.2&	48.6&	81.6&	59.0\\
			\centering CLIP-2$^\star$ &\textbf{71.2}	&83.6 &	88.5&	78.9&	85.8&	85.5&	71.9&	\textbf{71.2}&	86.3&	\textbf{81.8}&	73.2&	\textbf{90.7}&	\textbf{80.7} \\
			\bottomrule
		\end{tabular}%
	}
	\label{tabl}%
\end{table*}%

\subsection{Comparison with SoTA PEFT Methods}
RST can be regarded as a type of PEFT designed specifically for unsupervised domain adaptation. Therefore, in this subsection, we will conduct a fair comparison with advanced PEFT methods. In our experiments, we consistently utilize the CMKD approach proposed in this paper as the UDA training method, combined with various PEFT techniques. We compare the average accuracy and DSP of these approaches. This section of the experiments is performed on the Office-Home benchmark. All configurations will remain the same except for PEFT-related hyperparameters. To ensure that the DSPs of the related methods are comparable, we set the ranks of our LoRAs \cite{ref14} as 1 and 4, represented as $\text{LoRA}_{r=1}$ and $\text{LoRA}_{r=4}$ respectively. For AdaLoRA \cite{ref71}, the initial ranks are set as 8 and 2, and the corresponding target ranks as 4 and 1, denoted as $\text{AdaLoRA}_{r=2\text{→}1}$ and $\text{AdaLoRA}_{r=8\text{→}4}$. Furthermore, we also compare ``Fine-Tuning'' with ``Linear-Probe''. The former refers to training the entire backbone and task-head, whereas the latter only trains the task-head. As AdaLoRA \cite{ref71} and Adapter do not offer an implementation of CNNs, comparisons can only be conducted within the framework of the ViT architecture.

As shown in Table VII, our method demonstrates outstanding performance. With the ResNet50 architecture, we achieve an average accuracy of \textbf{78.8\%} using only \textbf{1.23M} DSP, resulting in a mere \textbf{-0.7\%} accuracy drop compared to Fine-Tuning. Moreover, the DSP employed is only \textbf{0.2\%} of that used in Fine-Tuning, highlighting the efficiency of our approach. The effectiveness of our method is further confirmed with the ViT-B architecture. We attain an average accuracy of \textbf{87.7\%} with just \textbf{1.02M} DSP, a significant reduction in downstream task model storage compared to Fine-Tuning. Furthermore, our approach significantly outperforms the state-of-the-art method of PEFT in the LLM domain. For instance, with the ViT-B architecture, RST outperforms BitFiT, $\text{LoRA}_{r=1}$, and $\text{AdaLoRA}_{r=2\text{→}1}$ by \textbf{+0.7\%}, \textbf{+1.2\%}, and \textbf{+1.0\%}, respectively. Additionally, RST requires \textbf{-0.38}, \textbf{-2.38}, and \textbf{-1.54} less DSP than BitFiT, $\text{LoRA}_{r=1}$, and $\text{AdaLoRA}_{r=2\text{→}1}$, respectively. Surprisingly, we have observed that the advanced LoRA series does not appear to outperform in the case of large data trained but small-scale models like CLIP. In addition to RST we proposed, even a simple method like BitFIT, which only trains Bias, outperforms the entire LoRA family, both in terms of average accuracy and DSP.

Throughout the experiments, we did not observe that the PEFT method outperformed Fine-Tuning in terms of average accuracy, which differs from the LLM domain. Therefore, the pursuit of efficient fine-tuning on large data with small models presents a promising and worthwhile problem to explore, particularly in scenarios like the UDA domain with numerous downstream tasks.

\subsection{Ablation Study}
\textbf{Baseline Configuration}. To the best of our knowledge, we are the first to fine-tuning with visual-language pre-training model in the UDA domain. Therefore, it is necessary to explore the effect of hyperparameters on the Baseline performance. In this section, we focus on investigating the impact of the learning rate. Following the previous UDA approaches, for the Office-Home dataset, we set the initial learning rate of the backbone to 3e-4 and the task head learning rate to 10 times that of the backbone (referred to as Configuration1). However, due to CLIP being pre-trained on a large-scale dataset, the model parameters are particularly sensitive to the learning rate. As a result, in Configuration2, we set the initial learning rate of the backbone to 3e-7 for ResNet50 and 3e-6 for ViT-B, while the task head learning rate is set to 10,000 and 1,000 times that of the backbone. Besides, ``Batch Normalization'' in RseNet is Frozen. The remaining hyperparameter details are kept the same, as detailed in the ``Implementation Details'' of Section IV.A. In addition to exploring the hyperparameter settings, we also aim to understand the impact of different pre-training models on the baseline performance. Therefore, we conduct experiments using Configuration1 on common pre-training models (e.g., ResNet50 pre-trained on ImageNet-1k and ViT-B pre-trained on ImageNet-21k), denoted as ``ImageNet-1k'' or ``ImageNet-21k'' on the Table VIII. We also utilized configuration1 and configuration2 for the pre-training model of CLIP, denoted as ``CLIP-1'' and ``CLIP-2'', respectively. This allows us to compare the contribution of the pre-training datasets from CLIP to that from ImageNet to the Baseline.

As shown in Table VIII, the configuration used in the previous method is not suitable for the VLP model, and the performance of the CLIP-based Baseline is notably inferior to that of the ImageNet pre-training model. For instance, based on ResNet50, there is a difference of \textbf{-29.9\%} between CLIP-1 and ImageNet-1k. However, when Configuration2 is applied, the performance of the Baseline significantly improves, with a \textbf{+31.1\%} and \textbf{+21.7\%} enhancement in CLIP-2 compared to CLIP-1 for ResNet50 and ViT-B architectures, respectively. Nevertheless, based on the experimental results, CLIP's dataset does not lead to significant improvements, and the average accuracy of CLIP-2 is only \textbf{+1.2\%} and \textbf{+0.9\%} higher than that of ImageNet.

\begin{table*}[htbp]
	\setlength{\abovecaptionskip}{0cm}  
	\setlength{\belowcaptionskip}{-0.2cm} 
	\centering
	\caption{Contribution of each part in CMKD based on Office-Home using ResNet50.}
	\renewcommand{\arraystretch}{1.0}
	\setlength{\tabcolsep}{0.5mm}{
		\begin{tabular}{cccc|ccccccccccccc}
			\toprule
			 ${{\mathcal{L}}_{\text{cls}}}$ & ${{\mathcal{L}}_{\text{task}}}$ & ${{\mathcal{L}}_{\text{distill}}}$ & ${{\mathcal{L}}_{\text{reg}}}$ & {Ar→Cl} & {Ar→Pr} & {Ar→Re} & {Cl→Ar} & {Cl→Pr} & {Cl→Re} & {Pr→Ar} & {Pr→Cl} & {Pr→Re} & {Re→Ar} & {Re→Cl} & {Re→Pr} & {Avg.}\\
			\midrule
			\ding{52}  & &   & &  52.1&	69.7&	75.2&	52.5&	64.3&	63.9&	56.2&	48.4&	74.9&	68.9&	54.4&	82.7&	63.7\\
			\ding{52}  & \ding{52} &   & &64.2  & 85.6	& 86.9	& 73.0	& 86.4	&84.6	&75.1	&63.4	&87.0	&78.4	&66.4	&89.3	&78.4	\\
			\ding{52}  &  & \ding{52}  & & 63.3  &83.9	&85.3	&71.5	&85.7	&82.9	&73.4	&61.5	&86.0	&77.7	&64.3	&88.6	&77.0	\\
			\ding{52}  &  \ding{52} & \ding{52}   &  & 65.1  &85.9	&87.1	&73.9	&87.1	&\textbf{85.8}	&75.3	&63.9	&\textbf{87.9}	&78.7	&66.9	&89.8	&79.0	\\
			\ding{52}  & \ding{52} & \ding{52} & \ding{52} & \textbf{65.9}      & \textbf{86.6}       & \textbf{87.3}      &   \textbf{74.4}    &  \textbf{87.7}    & \textbf{85.8}      & \textbf{75.9}       & \textbf{64.4}      &  \textbf{87.9}    & \textbf{79.1}      & \textbf{67.2}      & \textbf{90.0}      &\textbf{79.3}\\
			\bottomrule
		\end{tabular}%
	}
	\label{tabl}%
\end{table*}%

\begin{table*}[htbp]
	\setlength{\abovecaptionskip}{0cm}  
	\setlength{\belowcaptionskip}{-0.0cm} 
	\centering
	\caption{Comparison with different ${\alpha}$ in CMKD based on Office-Home using ResNet50}
	\renewcommand{\arraystretch}{1.2}
	\setlength{\tabcolsep}{0.5mm}{
		\begin{tabular}{c|ccccccccccccc}
			\toprule
			${\alpha}$& {Ar→Cl} & {Ar→Pr} & {Ar→Re} & {Cl→Ar} & {Cl→Pr} & {Cl→Re} & {Pr→Ar} & {Pr→Cl} & {Pr→Re} & {Re→Ar} & {Re→Cl} & {Re→Pr} & {Avg.} \\
			\midrule
			$0.5$ &64.7 & 85.6 & 86.8 &72.6 &87.4  & 85.2 &74.5  &63.7  & 87.0 &78.5  &66.6  &89.4  &78.5 \\
			$\emph{sg}(\exp \left(-\mathrm{GE}\left(p_{h}^{t}\right)\right))$ &65.7 &85.8  &87.2  & \textbf{75.0}& \textbf{88.3} &85.7  &\textbf{76.3}  &\textbf{64.5}  &87.6  & \textbf{79.2} &67.0  &89.9  &79.1\\
			$\emph{sg}(\exp (-\text{KL}(p_{h}^{t}\parallel p_{\text{g}}^{t})))$	& \textbf{65.9}      & \textbf{86.6}       & \textbf{87.3}      &   74.4    &  87.7    & \textbf{85.8}      & 75.9       & 64.4      &  \textbf{87.9}    & 79.1      & \textbf{67.2}      & \textbf{90.0}      &\textbf{79.3}\\
			\bottomrule
		\end{tabular}%
	}
	\label{table8}%
\end{table*}%

\textbf{Contribution of components in CMKD}. Contribution of the components. CMKD comprises three components: task-term ${{\mathcal{L}}_{\text{task}}}$, distillation term ${{\mathcal{L}}_{\text{distill}}}$ and regularization ${{\mathcal{L}}_{\text{reg}}}$. To evaluate the individual contributions of each component to the overall approach, we conducted an additional experiment to assess the effectiveness of average accuracy on the Office-Home dataset using ResNet50. The classification loss ${{\mathcal{L}}_{\text{cls}}}$ is used throughout the experiment. As shown in Table IX, ${{\mathcal{L}}_{\text{task}}}$ alone achieves \textbf{78.4\%} and ${{\mathcal{L}}_{\text{distill}}}$ achieves \textbf{77.0\%}, respectively, outperforming the baseline by \textbf{+14.7\%} and \textbf{+13.3\%}. The former represents the performance of the model learning on the target domain from the student's perspective, while the latter is the result of learning with the teacher's perspective (text encoder) as an aid. When the two are used in combination, the performance reaches \textbf{79.0\%}, demonstrating that the student perspective and the teacher perspective complement each other. Additionally, when regularization was applied to prevent the teacher perspective from being biased, the average accuracy reached \textbf{79.3\%}.

\textbf{Trade-off ${\alpha}$ in CMKD}. CMKD is a KD-inspired UDA method. In KD \cite{ref28}, a trade-off parameter ``${\alpha}$'' is used as an artificially designed hyperparameter to balance ${{\mathcal{L}}_{\text{task}}}$ and ${{\mathcal{L}}_{\text{distill}}}$ losses. However, in the UDA domain, where real labels for knowledge transfer are lacking, we propose using a dynamic mechanism to generate a different trade-off for each sample. In Table X, we compare the experimental results of the artificial selection with the two dynamic mechanisms, namely $\emph{sg}(\exp \left(-\mathrm{GE}\left(p_{h}^{t}\right)\right))$ and $\emph{sg}(\exp (-\text{KL}(p_{h}^{t}\parallel p_{\text{g}}^{t})))$. The first dynamic mechanism uses entropy to estimate the uncertainty of the current task head, and when the uncertainty is large, the teacher's perspective is used as the dominant one. The second mechanism considers the consistency between the student's perspective and the teacher's perspective, and when they are consistent, the student's perspective is dominant; otherwise, the teacher's perspective is dominant. In Table X, the experimental results show that the manually designed trade-off performs poorly, achieving only \textbf{78.5\%} average accuracy. In contrast, the two dynamic mechanisms outperform the former by \textbf{+0.6\%} and \textbf{+0.8\%}, respectively. Ultimately, we choose $\emph{sg}(\exp (-\text{KL}(p_{h}^{t}\parallel p_{\text{g}}^{t})))$ as the trade-off.

\begin{figure}[t]
	\setlength{\abovecaptionskip}{0cm}  
	\setlength{\belowcaptionskip}{-0.2cm} 
	\centering
	\includegraphics[width=3.5in]{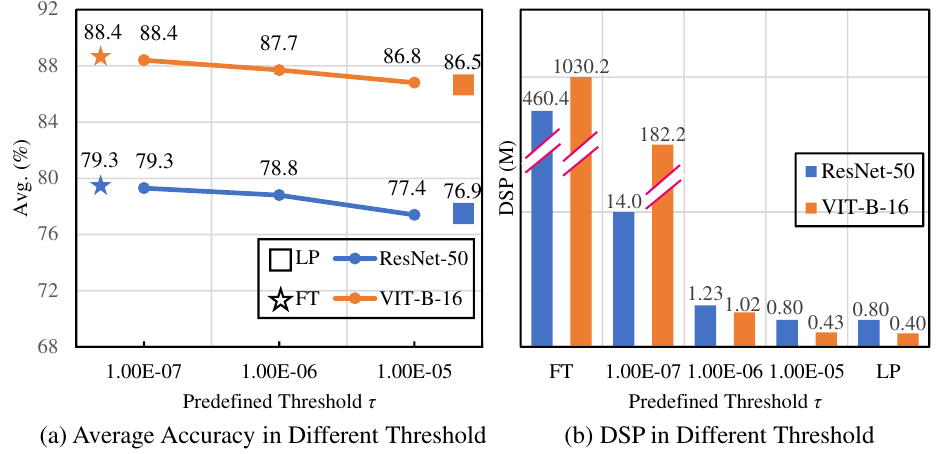}
	\caption{Average accuracy and DSP comparison in different predefined threshold on Office-Home. LP and FT stand for Linear-Probe and Fine-Tuning, respectively.}
	\label{fig_4}
\end{figure}

\textbf{Different Predefined Threshold $\tau$ in RST}. Similar to the related work of PEFT, RST can control the number of downstream task parameters by adjusting the hyperparameter $\tau$. Generally, a larger DSP (closer to Fine-Tuning) leads to better performance, while a smaller DSP (closer to Linear-Probe) results in less effectiveness. To strike a better balance between DSP and performance, we set different values of $\tau$ for residual sparse training on the Office-Home benchmark. As shown in Figure 4, when the  $\tau$ is chosen as 1e-6, a better trade-off can be achieved. Specifically, for ResNet50, it achieves \textbf{78.8\%} accuracy with a DSP of \textbf{1.23}, while for ViT-B-16, it achieves \textbf{87.7\%} accuracy with a DSP of \textbf{1.02}. In comparison, the average accuracy only drops by \textbf{-0.7\%} and \textbf{-0.5\%} while DSP drops by \textbf{-99.61\%} and \textbf{-99.90\%}, when compared to Fine-Tuning. The flexibility and effectiveness of tau selection allow it to be chosen according to the specific task requirements. 

\textbf{[Additional ablation studies and analyses have been included in the Appendix. Furthermore, the Appendix showcases the performance of the proposed method on various other tasks.]}

\section{Conclusion}
In this work, we have presented a comprehensive study on unsupervised domain adaptation tasks, focusing on two significant challenges that have been underexplored in the literature. Firstly, we identified the potential of large-scale visual-language pre-training models for UDA and proposed a novel method, cross-modal knowledge distillation, which effectively harnesses VLP models as teachers to guide the target domain model's learning process. Our experiments demonstrate that CMKD outperforms traditional UDA methods based on ImageNet pre-trained models, achieving state-of-the-art results in various UDA tasks. Secondly, we addressed the issue of storage overhead in UDA by introducing residual sparse training, which efficiently reduces the parameter count of VLP models while preserving competitive performance with Fine-Tuning. The RST approach allows for the deployment of UDA models across multiple transfer tasks with significantly reduced storage requirements, making it more practical and scalable for real-world applications. We believe that our findings and methodologies will be beneficial to researchers and practitioners working on UDA tasks, facilitating the development of resource-efficient domain adaptation techniques.

\bibliographystyle{IEEEtran}
\bibliography{reference}

\begin{thebibliography}{10}
\providecommand{\url}[1]{#1}
\csname url@samestyle\endcsname
\providecommand{\newblock}{\relax}
\providecommand{\bibinfo}[2]{#2}
\providecommand{\BIBentrySTDinterwordspacing}{\spaceskip=0pt\relax}
\providecommand{\BIBentryALTinterwordstretchfactor}{4}
\providecommand{\BIBentryALTinterwordspacing}{\spaceskip=\fontdimen2\font plus
\BIBentryALTinterwordstretchfactor\fontdimen3\font minus
  \fontdimen4\font\relax}
\providecommand{\BIBforeignlanguage}[2]{{%
\expandafter\ifx\csname l@#1\endcsname\relax
\typeout{** WARNING: IEEEtran.bst: No hyphenation pattern has been}%
\typeout{** loaded for the language `#1'. Using the pattern for}%
\typeout{** the default language instead.}%
\else
\language=\csname l@#1\endcsname
\fi
#2}}
\providecommand{\BIBdecl}{\relax}
\BIBdecl

\bibitem{ref1}
R.~Xia, G.~Li, Z.~Huang, H.~Meng, and Y.~Pang, ``Cbash: Combined backbone and
  advanced selection heads with object semantic proposals for weakly supervised
  object detection,'' \emph{IEEE Trans. Circuits Syst. Video Technol.},
  vol.~32, no.~10, pp. 6502--6514, 2022.

\bibitem{ref2}
J.~Ji, R.~Shi, S.~Li, P.~Chen, and Q.~Miao, ``Encoder-decoder with cascaded
  crfs for semantic segmentation,'' \emph{IEEE Trans. Circuits Syst. Video
  Technol.}, vol.~31, no.~5, pp. 1926--1938, 2020.

\bibitem{ref3}
W.~Dong, T.~Yang, J.~Qu, T.~Zhang, S.~Xiao, and Y.~Li, ``Joint contextual
  representation model-informed interpretable network with dictionary aligning
  for hyperspectral and lidar classification,'' \emph{IEEE Trans. Circuits
  Syst. Video Technol.}, 2023.

\bibitem{ref4}
M.~Long, Y.~Cao, J.~Wang, and M.~Jordan, ``Learning transferable features with
  deep adaptation networks,'' in \emph{Proc. Int. Conf. Mach. Learn}, 2015, pp.
  97--105.

\bibitem{ref5}
Y.~Ganin and V.~Lempitsky, ``Unsupervised domain adaptation by
  backpropagation,'' in \emph{Proc. Int. Conf. Mach. Learn.}, 2015, pp.
  1180--1189.

\bibitem{ref6}
Y.~Zhu, F.~Zhuang, J.~Wang, G.~Ke, J.~Chen, J.~Bian, H.~Xiong, and Q.~He,
  ``Deep subdomain adaptation network for image classification,'' \emph{IEEE
  Trans. Neural Net. Learn. Syst.}, vol.~32, no.~4, pp. 1713--1722, 2020.

\bibitem{ref7}
T.~Xu, W.~Chen, P.~Wang, F.~Wang, H.~Li, and R.~Jin, ``Cdtrans: Cross-domain
  transformer for unsupervised domain adaptation,'' in \emph{Proc. Int. Conf.
  Learn. Represent.}, 2022.

\bibitem{ref8}
A.~Radford, J.~W. Kim, C.~Hallacy, A.~Ramesh, G.~Goh, S.~Agarwal, G.~Sastry,
  A.~Askell, P.~Mishkin, J.~Clark \emph{et~al.}, ``Learning transferable visual
  models from natural language supervision,'' in \emph{Proc. Int. Conf. Mach.
  Learn}, 2021, pp. 8748--8763.

\bibitem{ref9}
Y.~Li, H.~Fan, R.~Hu, C.~Feichtenhofer, and K.~He, ``Scaling language-image
  pre-training via masking,'' in \emph{Proc. IEEE Conf. Comput. Vis. Pattern
  Recognit.}, 2023, pp. 23\,390--23\,400.

\bibitem{ref10}
J.~Deng, W.~Dong, R.~Socher, L.-J. Li, K.~Li, and L.~Fei-Fei, ``Imagenet: A
  large-scale hierarchical image database,'' in \emph{Proc. IEEE Conf. Comput.
  Vis. Pattern Recognit.}, 2009, pp. 248--255.

\bibitem{ref11}
O.~Russakovsky, J.~Deng, H.~Su, J.~Krause, S.~Satheesh, S.~Ma, Z.~Huang,
  A.~Karpathy, A.~Khosla, M.~Bernstein \emph{et~al.}, ``Imagenet large scale
  visual recognition challenge,'' \emph{Int. J. Comput. Vis.}, vol. 115, pp.
  211--252, 2015.

\bibitem{ref27}
C.~Ge, R.~Huang, M.~Xie, Z.~Lai, S.~Song, S.~Li, and G.~Huang, ``Domain
  adaptation via prompt learning,'' \emph{arXiv preprint arXiv:2202.06687},
  2022.

\bibitem{ref12}
X.~Dong, J.~Bao, T.~Zhang, D.~Chen, S.~Gu, W.~Zhang, L.~Yuan, D.~Chen, F.~Wen,
  and N.~Yu, ``Clip itself is a strong fine-tuner: Achieving 85.7\% and 88.0\%
  top-1 accuracy with vit-b and vit-l on imagenet,'' \emph{arXiv preprint
  arXiv:2212.06138}, 2022.

\bibitem{ref13}
K.~Zhou, J.~Yang, C.~C. Loy, and Z.~Liu, ``Learning to prompt for
  vision-language models,'' \emph{Int. J. Comput. Vis.}, vol. 130, no.~9, pp.
  2337--2348, 2022.

\bibitem{ref80}
L.~Yan, C.~Han, Z.~Xu, D.~Liu, and Q.~Wang, ``Prompt learns prompt: exploring
  knowledge-aware generative prompt collaboration for video captioning,'' in
  \emph{Proc. IJCAI}, 2023, pp. 1622--1630.

\bibitem{ref14}
E.~J. Hu, Y.~Shen, P.~Wallis, Z.~Allen-Zhu, Y.~Li, S.~Wang, L.~Wang, and
  W.~Chen, ``Lora: Low-rank adaptation of large language models,'' in
  \emph{Proc. Int. Conf. Learn. Represent.}, 2022.

\bibitem{ref15}
H.~Touvron, T.~Lavril, G.~Izacard, X.~Martinet, M.-A. Lachaux, T.~Lacroix,
  B.~Rozi{\`e}re, N.~Goyal, E.~Hambro, F.~Azhar \emph{et~al.}, ``Llama: Open
  and efficient foundation language models,'' \emph{arXiv preprint
  arXiv:2302.13971}, 2023.

\bibitem{ref16}
X.~Ding, X.~Zhang, N.~Ma, J.~Han, G.~Ding, and J.~Sun, ``Repvgg: Making
  vgg-style convnets great again,'' in \emph{Proc. IEEE Conf. Comput. Vis.
  Pattern Recognit.}, 2021, pp. 13\,733--13\,742.

\bibitem{ref17}
K.~He, X.~Zhang, S.~Ren, and J.~Sun, ``Deep residual learning for image
  recognition,'' in \emph{Proc. IEEE Conf. Comput. Vis. Pattern Recognit.},
  2016, pp. 770--778.

\bibitem{ref18}
A.~Dosovitskiy, L.~Beyer, A.~Kolesnikov, D.~Weissenborn, X.~Zhai,
  T.~Unterthiner, M.~Dehghani, M.~Minderer, G.~Heigold, S.~Gelly \emph{et~al.},
  ``An image is worth 16x16 words: Transformers for image recognition at
  scale,'' in \emph{Proc. Int. Conf. Learn. Represent.}, 2021.

\bibitem{ref34}
W.~Su, X.~Zhu, Y.~Cao, B.~Li, L.~Lu, F.~Wei, and J.~Dai, ``Vl-bert:
  Pre-training of generic visual-linguistic representations,'' in \emph{Proc.
  Int. Conf. Learn. Represent.}, 2020, pp. 2790--2799.

\bibitem{ref35}
W.~Kim, B.~Son, and I.~Kim, ``Vilt: Vision-and-language transformer without
  convolution or region supervision,'' in \emph{Proc. Int. Conf. Mach. Learn.},
  2021, pp. 5583--5594.

\bibitem{ref86}
Z.~Wang, J.~Yu, A.~W. Yu, Z.~Dai, Y.~Tsvetkov, and Y.~Cao, ``Simvlm: Simple
  visual language model pretraining with weak supervision,'' in \emph{ICLR},
  2022.

\bibitem{ref88}
J.~Ke, K.~Ye, J.~Yu, Y.~Wu, P.~Milanfar, and F.~Yang, ``Vila: Learning image
  aesthetics from user comments with vision-language pretraining,'' in
  \emph{Proc. IEEE Conf. Comput. Vis. Pattern Recognit.}, 2023, pp.
  10\,041--10\,051.

\bibitem{ref19}
H.~Li, N.~Dong, Z.~Yu, D.~Tao, and G.~Qi, ``Triple adversarial learning and
  multi-view imaginative reasoning for unsupervised domain adaptation person
  re-identification,'' \emph{IEEE Trans. Circuits Syst. Video Technol.},
  vol.~32, no.~5, pp. 2814--2830, 2021.

\bibitem{ref20}
H.~Rangwani, S.~K. Aithal, M.~Mishra, A.~Jain, and V.~B. Radhakrishnan, ``A
  closer look at smoothness in domain adversarial training,'' in \emph{Proc.
  Int. Conf. Mach. Learn.}, 2022, pp. 18\,378--18\,399.

\bibitem{ref21}
H.~Liu, J.~Wang, and M.~Long, ``Cycle self-training for domain adaptation,'' in
  \emph{Proc. Adv. Neural Inf. Process. Syst.}, 2021, pp. 22\,968--22\,981.

\bibitem{ref22}
Z.~Deng, Y.~Luo, and J.~Zhu, ``Cluster alignment with a teacher for
  unsupervised domain adaptation,'' in \emph{Proc. IEEE Int. Conf. Comput.
  Vis.}, 2019, pp. 9944--9953.

\bibitem{ref23}
A.~Vaswani, N.~Shazeer, N.~Parmar, J.~Uszkoreit, L.~Jones, A.~N. Gomez,
  {\L}.~Kaiser, and I.~Polosukhin, ``Attention is all you need,'' in
  \emph{Proc. Adv. Neural Inf. Process. Syst.}, 2017.

\bibitem{ref24}
J.~Yang, J.~Liu, N.~Xu, and J.~Huang, ``Tvt: Transferable vision transformer
  for unsupervised domain adaptation,'' in \emph{Proc. WACV}, 2023, pp.
  520--530.

\bibitem{ref26}
J.~Zhu, H.~Bai, and L.~Wang, ``Patch-mix transformer for unsupervised domain
  adaptation: A game perspective,'' in \emph{Proc. IEEE Conf. Comput. Vis.
  Pattern Recognit.}, 2023, pp. 3561--3571.

\bibitem{ref28}
G.~Hinton, O.~Vinyals, and J.~Dean, ``Distilling the knowledge in a neural
  network,'' \emph{arXiv preprint arXiv:1503.02531}, 2015.

\bibitem{ref29}
S.~Yang, L.~Xu, M.~Zhou, X.~Yang, J.~Yang, and Z.~Huang, ``Skill-transferring
  knowledge distillation method,'' \emph{IEEE Trans. Circuits Syst. Video
  Technol.}, 2023.

\bibitem{ref30}
X.~Zhang, X.~Wang, and P.~Cheng, ``Unsupervised hashing retrieval via efficient
  correlation distillation,'' \emph{IEEE Trans. Circuits Syst. Video Technol.},
  2023.

\bibitem{ref76}
C.~Han, Q.~Wang, Y.~Cui, Z.~Cao, W.~Wang, S.~Qi, and D.~Liu, ``E\^{} 2vpt: An
  effective and efficient approach for visual prompt tuning,'' in \emph{Proc.
  IEEE Int. Conf. Comput. Vis.}, 2023.

\bibitem{ref77}
C.~Han, Q.~Wang, Y.~Cui, W.~Wang, L.~Huang, S.~Qi, and D.~Liu, ``Facing the
  elephant in the room: Visual prompt tuning or full finetuning?'' \emph{Proc.
  Int. Conf. Learn. Represent.}, 2024.

\bibitem{ref31}
X.~L. Li and P.~Liang, ``Prefix-tuning: Optimizing continuous prompts for
  generation,'' in \emph{Proc. ACL-IJCNLP}, 2021, pp. 4582--4597.

\bibitem{ref32}
B.~Lester, R.~Al-Rfou, and N.~Constant, ``The power of scale for
  parameter-efficient prompt tuning,'' in \emph{Proc. ACL-IJCNLP}, 2021, pp.
  4582--4597.

\bibitem{ref33}
N.~Houlsby, A.~Giurgiu, S.~Jastrzebski, B.~Morrone, Q.~De~Laroussilhe,
  A.~Gesmundo, M.~Attariyan, and S.~Gelly, ``Parameter-efficient transfer
  learning for nlp,'' in \emph{Proc. Int. Conf. Mach. Learn.}, 2019, pp.
  2790--2799.

\bibitem{ref78}
L.~Yang, Q.~Wang, J.~Wang, X.~Quan, F.~Feng, Y.~Chen, M.~Khabsa, S.~Wang,
  Z.~Xu, and D.~Liu, ``Mixpave: Mix-prompt tuning for few-shot product
  attribute value extraction,'' in \emph{Proc. ACL}, 2023, pp. 9978--9991.

\bibitem{ref79}
F.~Ma, C.~Zhang, L.~Ren, J.~Wang, Q.~Wang, W.~Wu, X.~Quan, and D.~Song,
  ``Xprompt: Exploring the extreme of prompt tuning,'' \emph{arXiv preprint
  arXiv:2210.04457}, 2022.

\bibitem{ref87}
S.~Dong, Y.~Feng, Q.~Yang, Y.~Huang, D.~Liu, and H.~Fan, ``Efficient multimodal
  semantic segmentation via dual-prompt learning,'' \emph{arXiv preprint
  arXiv:2312.00360}, 2023.

\bibitem{ref37}
D.-H. Lee, ``Pseudo-label: The simple and efficient semi-supervised learning
  method for deep neural networks,'' in \emph{Proc. Int. Conf. Mach. Learn.},
  vol.~3, no.~2, 2013, pp. 896--902.

\bibitem{ref39}
A.~Aghajanyan, L.~Zettlemoyer, and S.~Gupta, ``Intrinsic dimensionality
  explains the effectiveness of language model fine-tuning,'' \emph{arXiv
  preprint arXiv:2012.13255}, 2020.

\bibitem{ref40}
P.~Mateos-Aparicio and A.~Rodr{\'\i}guez-Moreno, ``The impact of studying brain
  plasticity,'' \emph{Frontiers in cellular neuroscience}, vol.~13, no.~66,
  2019.

\bibitem{ref41}
H.~Venkateswara, J.~Eusebio, S.~Chakraborty, and S.~Panchanathan, ``Deep
  hashing network for unsupervised domain adaptation,'' in \emph{Proc. IEEE
  Conf. Comput. Vis. Pattern Recognit.}, 2017, pp. 5018--5027.

\bibitem{ref42}
K.~Saenko, B.~Kulis, M.~Fritz, and T.~Darrell, ``Adapting visual category
  models to new domains,'' in \emph{Proc. Eur. Conf. Comput. Vis.}, 2010, pp.
  213--226.

\bibitem{ref44}
Y.~Netzer, T.~Wang, A.~Coates, A.~Bissacco, B.~Wu, and A.~Y. Ng, ``Reading
  digits in natural images with unsupervised feature learning,'' in \emph{Proc.
  Adv. Neural Inf. Process. Syst.}, 2011.

\bibitem{ref45}
L.~Deng, ``The mnist database of handwritten digit images for machine learning
  research [best of the web],'' \emph{IEEE signal processing magazine},
  vol.~29, no.~6, pp. 141--142, 2012.

\bibitem{ref46}
J.~J. Hull, ``A database for handwritten text recognition research,''
  \emph{IEEE Trans. Pattern Anal. Mach. Intell}, vol.~16, no.~5, pp. 550--554,
  1994.

\bibitem{ref47}
M.~Long, H.~Zhu, J.~Wang, and M.~I. Jordan, ``Deep transfer learning with joint
  adaptation networks,'' in \emph{Proc. Int. Conf. Mach. Learn.}, 2017, pp.
  2208--2217.

\bibitem{ref43}
X.~Peng, B.~Usman, N.~Kaushik, J.~Hoffman, D.~Wang, and K.~Saenko, ``Visda: The
  visual domain adaptation challenge,'' \emph{arXiv preprint arXiv:1710.06924},
  2017.

\bibitem{ref48}
X.~Peng, Q.~Bai, X.~Xia, Z.~Huang, K.~Saenko, and B.~Wang, ``Moment matching
  for multi-source domain adaptation,'' in \emph{Proc. IEEE Int. Conf. Comput.
  Vis.}, 2019, pp. 1406--1415.

\bibitem{ref49}
Y.~Grandvalet and Y.~Bengio, ``Semi-supervised learning by entropy
  minimization,'' in \emph{Proc. Adv. Neural Inf. Process. Syst.}, 2005, pp.
  281--296.

\bibitem{ref50}
M.~Long, Z.~Cao, J.~Wang, and M.~I. Jordan, ``Conditional adversarial domain
  adaptation,'' in \emph{Proc. Adv. Neural Inf. Process. Syst.}, 2018, pp.
  1647--1657.

\bibitem{ref51}
X.~Chen, S.~Wang, M.~Long, and J.~Wang, ``Transferability vs. discriminability:
  Batch spectral penalization for adversarial domain adaptation,'' in
  \emph{Proc. Int. Conf. Mach. Learn.}, 2019, pp. 1081--1090.

\bibitem{ref52}
X.~Wang, Y.~Jin, M.~Long, J.~Wang, and M.~I. Jordan, ``Transferable
  normalization: Towards improving transferability of deep neural networks,''
  in \emph{Proc. Adv. Neural Inf. Process. Syst.}, 2019.

\bibitem{ref54}
C.-Y. Lee, T.~Batra, M.~H. Baig, and D.~Ulbricht, ``Sliced wasserstein
  discrepancy for unsupervised domain adaptation,'' in \emph{Proc. IEEE Conf.
  Comput. Vis. Pattern Recognit.}, 2019, pp. 10\,285--10\,295.

\bibitem{ref55}
W.-G. Chang, T.~You, S.~Seo, S.~Kwak, and B.~Han, ``Domain-specific batch
  normalization for unsupervised domain adaptation,'' in \emph{Proc. IEEE Conf.
  Comput. Vis. Pattern Recognit.}, 2019, pp. 7354--7362.

\bibitem{ref57}
S.~Cui, S.~Wang, J.~Zhuo, L.~Li, Q.~Huang, and Q.~Tian, ``Towards
  discriminability and diversity: Batch nuclear-norm maximization under label
  insufficient situations,'' in \emph{Proc. IEEE Conf. Comput. Vis. Pattern
  Recognit.}, 2020, pp. 3941--3950.

\bibitem{ref58}
S.~Li, C.~Liu, Q.~Lin, B.~Xie, Z.~Ding, G.~Huang, and J.~Tang, ``Domain
  conditioned adaptation network,'' in \emph{Proc. AAAI}, 2020, pp.
  11\,386--11\,393.

\bibitem{ref59}
J.~Liang, D.~Hu, and J.~Feng, ``Do we really need to access the source data?
  source hypothesis transfer for unsupervised domain adaptation,'' in
  \emph{Proc. Int. Conf. Mach. Learn.}, 2020, pp. 6028--6039.

\bibitem{ref60}
{J. Liang, D. Hu, and J. Feng}, ``Domain adaptation with auxiliary target
  domain-oriented classifier,'' in \emph{Proc. IEEE Conf. Comput. Vis. Pattern
  Recognit.}, 2021, pp. 16\,632--16\,642.

\bibitem{ref61}
Z.~Du, J.~Li, H.~Su, L.~Zhu, and K.~Lu, ``Cross-domain gradient discrepancy
  minimization for unsupervised domain adaptation,'' in \emph{Proc. IEEE Conf.
  Comput. Vis. Pattern Recognit.}, 2021, pp. 3937--3946.

\bibitem{ref62}
A.~Chadha and Y.~Andreopoulos, ``Improved techniques for adversarial
  discriminative domain adaptation,'' \emph{IEEE Trans. Image Process.},
  vol.~29, pp. 2622--2637, 2019.

\bibitem{ref63}
K.~Saito, Y.~Ushiku, T.~Harada, and K.~Saenko, ``Adversarial dropout
  regularization,'' in \emph{Proc. Int. Conf. Learn. Represent.}, 2017.

\bibitem{ref64}
J.~Hoffman, E.~Tzeng, T.~Park, J.-Y. Zhu, P.~Isola, K.~Saenko, A.~Efros, and
  T.~Darrell, ``Cycada: Cycle-consistent adversarial domain adaptation,'' in
  \emph{Proc. Int. Conf. Mach. Learn.}, 2018, pp. 1989--1998.

\bibitem{ref65}
T.~Sun, C.~Lu, T.~Zhang, and H.~Ling, ``Safe self-refinement for
  transformer-based domain adaptation,'' in \emph{Proc. IEEE Conf. Comput. Vis.
  Pattern Recog.}, 2022, pp. 7191--7200.

\bibitem{ref66}
Y.~Jin, X.~Wang, M.~Long, and J.~Wang, ``Minimum class confusion for versatile
  domain adaptation,'' in \emph{Proc. Eur. Conf. Comput. Vis.}, 2020, pp.
  464--480.

\bibitem{ref67}
B.~Sun and K.~Saenko, ``Deep coral: Correlation alignment for deep domain
  adaptation,'' in \emph{Proc. Eur. Conf. Comput. Vis.}, 2016, pp. 443--450.

\bibitem{ref68}
Z.~Pei, Z.~Cao, M.~Long, and J.~Wang, ``Multi-adversarial domain adaptation,''
  in \emph{Proc. AAAI}, 2018, pp. 7618--7625.

\bibitem{ref70}
E.~B. Zaken, S.~Ravfogel, and Y.~Goldberg, ``Bitfit: Simple parameter-efficient
  fine-tuning for transformer-based masked language-models,'' in \emph{Proc.
  ACL}, 2022, pp. 1--9.

\bibitem{ref71}
Q.~Zhang, M.~Chen, A.~Bukharin, P.~He, Y.~Cheng, W.~Chen, and T.~Zhao,
  ``Adaptive budget allocation for parameter-efficient fine-tuning,'' in
  \emph{Proc. Int. Conf. Learn. Represent.}, 2023.

\bibitem{ref69}
K.~Sohn, D.~Berthelot, N.~Carlini, Z.~Zhang, H.~Zhang, C.~A. Raffel, E.~D.
  Cubuk, A.~Kurakin, and C.-L. Li, ``Fixmatch: Simplifying semi-supervised
  learning with consistency and confidence,'' in \emph{Proc. Adv. Neural Inf.
  Process. Syst.}, 2020, pp. 596--608.

\end{thebibliography}


\begin{thebibliography}{10}
\providecommand{\url}[1]{#1}
\csname url@samestyle\endcsname
\providecommand{\newblock}{\relax}
\providecommand{\bibinfo}[2]{#2}
\providecommand{\BIBentrySTDinterwordspacing}{\spaceskip=0pt\relax}
\providecommand{\BIBentryALTinterwordstretchfactor}{4}
\providecommand{\BIBentryALTinterwordspacing}{\spaceskip=\fontdimen2\font plus
\BIBentryALTinterwordstretchfactor\fontdimen3\font minus
  \fontdimen4\font\relax}
\providecommand{\BIBforeignlanguage}[2]{{%
\expandafter\ifx\csname l@#1\endcsname\relax
\typeout{** WARNING: IEEEtran.bst: No hyphenation pattern has been}%
\typeout{** loaded for the language `#1'. Using the pattern for}%
\typeout{** the default language instead.}%
\else
\language=\csname l@#1\endcsname
\fi
#2}}
\providecommand{\BIBdecl}{\relax}
\BIBdecl

\bibitem{ref41}
H.~Venkateswara, J.~Eusebio, S.~Chakraborty, and S.~Panchanathan, ``Deep
  hashing network for unsupervised domain adaptation,'' in \emph{Proc. IEEE
  Conf. Comput. Vis. Pattern Recognit.}, 2017, pp. 5018--5027.

\bibitem{ref42}
K.~Saenko, B.~Kulis, M.~Fritz, and T.~Darrell, ``Adapting visual category
  models to new domains,'' in \emph{Proc. Eur. Conf. Comput. Vis.}, 2010, pp.
  213--226.

\bibitem{ref47}
M.~Long, H.~Zhu, J.~Wang, and M.~I. Jordan, ``Deep transfer learning with joint
  adaptation networks,'' in \emph{Proc. Int. Conf. Mach. Learn.}, 2017, pp.
  2208--2217.

\bibitem{ref44}
Y.~Netzer, T.~Wang, A.~Coates, A.~Bissacco, B.~Wu, and A.~Y. Ng, ``Reading
  digits in natural images with unsupervised feature learning,'' in \emph{Proc.
  Adv. Neural Inf. Process. Syst.}, 2011.

\bibitem{ref45}
L.~Deng, ``The mnist database of handwritten digit images for machine learning
  research [best of the web],'' \emph{IEEE signal processing magazine},
  vol.~29, no.~6, pp. 141--142, 2012.

\bibitem{ref46}
J.~J. Hull, ``A database for handwritten text recognition research,''
  \emph{IEEE Trans. Pattern Anal. Mach. Intell}, vol.~16, no.~5, pp. 550--554,
  1994.

\bibitem{ref43}
X.~Peng, B.~Usman, N.~Kaushik, J.~Hoffman, D.~Wang, and K.~Saenko, ``Visda: The
  visual domain adaptation challenge,'' \emph{arXiv preprint arXiv:1710.06924},
  2017.

\bibitem{ref48}
X.~Peng, Q.~Bai, X.~Xia, Z.~Huang, K.~Saenko, and B.~Wang, ``Moment matching
  for multi-source domain adaptation,'' in \emph{Proc. IEEE Int. Conf. Comput.
  Vis.}, 2019, pp. 1406--1415.

\bibitem{ref8}
A.~Radford, J.~W. Kim, C.~Hallacy, A.~Ramesh, G.~Goh, S.~Agarwal, G.~Sastry,
  A.~Askell, P.~Mishkin, J.~Clark \emph{et~al.}, ``Learning transferable visual
  models from natural language supervision,'' in \emph{Proc. Int. Conf. Mach.
  Learn}, 2021, pp. 8748--8763.

\bibitem{ref69}
K.~Sohn, D.~Berthelot, N.~Carlini, Z.~Zhang, H.~Zhang, C.~A. Raffel, E.~D.
  Cubuk, A.~Kurakin, and C.-L. Li, ``Fixmatch: Simplifying semi-supervised
  learning with consistency and confidence,'' in \emph{Proc. Adv. Neural Inf.
  Process. Syst.}, 2020, pp. 596--608.

\bibitem{ref72}
E.~D. Cubuk, B.~Zoph, J.~Shlens, and Q.~V. Le, ``Randaugment: Practical
  automated data augmentation with a reduced search space,'' in \emph{Proc.
  IEEE Conf. Comput. Vis. Pattern Recog.}, 2020, pp. 702--703.

\bibitem{ref17}
K.~He, X.~Zhang, S.~Ren, and J.~Sun, ``Deep residual learning for image
  recognition,'' in \emph{Proc. IEEE Conf. Comput. Vis. Pattern Recognit.},
  2016, pp. 770--778.

\bibitem{ref10}
J.~Deng, W.~Dong, R.~Socher, L.-J. Li, K.~Li, and L.~Fei-Fei, ``Imagenet: A
  large-scale hierarchical image database,'' in \emph{Proc. IEEE Conf. Comput.
  Vis. Pattern Recognit.}, 2009, pp. 248--255.

\bibitem{ref12}
X.~Dong, J.~Bao, T.~Zhang, D.~Chen, S.~Gu, W.~Zhang, L.~Yuan, D.~Chen, F.~Wen,
  and N.~Yu, ``Clip itself is a strong fine-tuner: Achieving 85.7\% and 88.0\%
  top-1 accuracy with vit-b and vit-l on imagenet,'' \emph{arXiv preprint
  arXiv:2212.06138}, 2022.

\bibitem{ref18}
A.~Dosovitskiy, L.~Beyer, A.~Kolesnikov, D.~Weissenborn, X.~Zhai,
  T.~Unterthiner, M.~Dehghani, M.~Minderer, G.~Heigold, S.~Gelly \emph{et~al.},
  ``An image is worth 16x16 words: Transformers for image recognition at
  scale,'' in \emph{Proc. Int. Conf. Learn. Represent.}, 2021.

\bibitem{ref11}
O.~Russakovsky, J.~Deng, H.~Su, J.~Krause, S.~Satheesh, S.~Ma, Z.~Huang,
  A.~Karpathy, A.~Khosla, M.~Bernstein \emph{et~al.}, ``Imagenet large scale
  visual recognition challenge,'' \emph{Int. J. Comput. Vis.}, vol. 115, pp.
  211--252, 2015.

\bibitem{ref7}
T.~Xu, W.~Chen, P.~Wang, F.~Wang, H.~Li, and R.~Jin, ``Cdtrans: Cross-domain
  transformer for unsupervised domain adaptation,'' in \emph{Proc. Int. Conf.
  Learn. Represent.}, 2022.

\bibitem{ref74}
S.~Ioffe and C.~Szegedy, ``Batch normalization: Accelerating deep network
  training by reducing internal covariate shift,'' in \emph{Proc. Int. Conf.
  Mach. Learn.}, 2015, pp. 448--456.

\bibitem{ref75}
J.~L. Ba, J.~R. Kiros, and G.~E. Hinton, ``Layer normalization,'' \emph{arXiv
  preprint arXiv:1607.06450}, 2016.

\bibitem{ref73}
K.~You, Z.~Kou, M.~Long, and J.~Wang, ``Co-tuning for transfer learning,'' in
  \emph{Proc. Adv. Neural Inf. Process. Syst.}, 2020, pp. 17\,236--17\,246.

\bibitem{ref70}
E.~B. Zaken, S.~Ravfogel, and Y.~Goldberg, ``Bitfit: Simple parameter-efficient
  fine-tuning for transformer-based masked language-models,'' in \emph{Proc.
  ACL}, 2022, pp. 1--9.

\bibitem{ref14}
E.~J. Hu, Y.~Shen, P.~Wallis, Z.~Allen-Zhu, Y.~Li, S.~Wang, L.~Wang, and
  W.~Chen, ``Lora: Low-rank adaptation of large language models,'' in
  \emph{Proc. Int. Conf. Learn. Represent.}, 2022.

\bibitem{ref89}
S.~R. Richter, V.~Vineet, S.~Roth, and V.~Koltun, ``Playing for data: Ground
  truth from computer games,'' in \emph{Eur. Conf. Comput. Vis.}\hskip 1em plus
  0.5em minus 0.4em\relax Springer, 2016, pp. 102--118.

\bibitem{ref90}
M.~Cordts, M.~Omran, S.~Ramos, T.~Rehfeld, M.~Enzweiler, R.~Benenson,
  U.~Franke, S.~Roth, and B.~Schiele, ``The cityscapes dataset for semantic
  urban scene understanding,'' in \emph{IEEE Conf. Comput. Vis. Pattern
  Recognit.}, 2016, pp. 3213--3223.

\bibitem{ref91}
{L. Chen, G. Papandreou, I. Kokkinos, K. Murphy and A. Yuille}, ``Deeplab:
  Semantic image segmentation with deep convolutional nets, atrous convolution,
  and fully connected crfs,'' \emph{IEEE Trans. Pattern Anal. Mach. Intell.},
  vol.~40, no.~4, pp. 834--848, 2017.

\bibitem{ref92}
L.~Hoyer, D.~Dai, and L.~Van~Gool, ``Daformer: Improving network architectures
  and training strategies for domain-adaptive semantic segmentation,'' in
  \emph{IEEE Conf. Comput. Vis. Pattern Recognit.}, 2022, pp. 9924--9935.

\bibitem{ref81}
Z.~Cao, M.~Long, J.~Wang, and M.~I. Jordan, ``Partial transfer learning with
  selective adversarial networks,'' in \emph{Proc. IEEE Conf. Comput. Vis.
  Pattern Recognit.}, 2018, pp. 2724--2732.

\bibitem{ref59}
J.~Liang, D.~Hu, and J.~Feng, ``Do we really need to access the source data?
  source hypothesis transfer for unsupervised domain adaptation,'' in
  \emph{Proc. Int. Conf. Mach. Learn.}, 2020, pp. 6028--6039.

\bibitem{ref82}
S.~Li, C.~H. Liu, Q.~Lin, Q.~Wen, L.~Su, G.~Huang, and Z.~Ding, ``Deep residual
  correction network for partial domain adaptation,'' \emph{IEEE Trans. Pattern
  Anal. Mach. Intell.}, vol.~43, no.~7, pp. 2329--2344, 2020.

\bibitem{ref83}
J.~Liang, Y.~Wang, D.~Hu, R.~He, and J.~Feng, ``A balanced and
  uncertainty-aware approach for partial domain adaptation,'' in \emph{Proc.
  Eur. Conf. Comput. Vis.}, 2020, pp. 123--140.

\bibitem{ref84}
C.-X. Ren, P.~Ge, P.~Yang, and S.~Yan, ``Learning target-domain-specific
  classifier for partial domain adaptation,'' \emph{IEEE Trans. Neural Net.
  Learn. Syst}, vol.~32, no.~5, pp. 1989--2001, 2020.

\bibitem{ref85}
J.~Liang, D.~Hu, Y.~Wang, R.~He, and J.~Feng, ``Source data-absent unsupervised
  domain adaptation through hypothesis transfer and labeling transfer,''
  \emph{IEEE Trans. Pattern Anal. Mach. Intell.}, vol.~44, no.~11, pp.
  8602--8617, 2021.

\bibitem{ref93}
B.~Cheng, I.~Misra, A.~G. Schwing, A.~Kirillov, and R.~Girdhar,
  ``Masked-attention mask transformer for universal image segmentation,'' in
  \emph{IEEE Conf. Comput. Vis. Pattern Recognit.}, 2022, pp. 1290--1299.

\bibitem{ref94}
C.~Sakaridis, D.~Dai, and L.~Van~Gool, ``Acdc: The adverse conditions dataset
  with correspondences for semantic driving scene understanding,'' in
  \emph{IEEE Conf. Comput. Vis. Pattern Recognit.}, 2021, pp. 10\,765--10\,775.

\bibitem{ref95}
A.~Tarvainen and H.~Valpola, ``Mean teachers are better role models:
  Weight-averaged consistency targets improve semi-supervised deep learning
  results,'' \emph{Proc. Adv. Neural Inf. Process. Syst.}, vol.~30, 2017.

\bibitem{ref100}
W.~Tranheden, V.~Olsson, J.~Pinto, and L.~Svensson, ``Dacs: Domain adaptation
  via cross-domain mixed sampling,'' in \emph{Proc. WACV}, 2021, pp.
  1379--1389.

\bibitem{ref101}
Y.~Yang and S.~Soatto, ``Fda: Fourier domain adaptation for semantic
  segmentation,'' in \emph{IEEE Conf. Comput. Vis. Pattern Recognit.}, 2020,
  pp. 4085--4095.

\bibitem{ref96}
Z.~Teed and J.~Deng, ``Raft: Recurrent all-pairs field transforms for optical
  flow,'' in \emph{Eur. Conf. Comput. Vis.}\hskip 1em plus 0.5em minus
  0.4em\relax Springer, 2020, pp. 402--419.

\bibitem{ref98}
A.~Dosovitskiy, P.~Fischer, E.~Ilg, P.~Hausser, C.~Hazirbas, V.~Golkov, P.~Van
  Der~Smagt, D.~Cremers, and T.~Brox, ``Flownet: Learning optical flow with
  convolutional networks,'' in \emph{IEEE Int. Conf. Comput. Vis.}, 2015, pp.
  2758--2766.

\bibitem{ref99}
N.~Mayer, E.~Ilg, P.~Hausser, P.~Fischer, D.~Cremers, A.~Dosovitskiy, and
  T.~Brox, ``A large dataset to train convolutional networks for disparity,
  optical flow, and scene flow estimation,'' in \emph{IEEE Conf. Comput. Vis.
  Pattern Recognit.}, 2016, pp. 4040--4048.

\bibitem{ref97}
M.~Menze, C.~Heipke, and A.~Geiger, ``Object scene flow,'' \emph{ISPRS Journal
  of Photogrammetry and Remote Sensing}, 2018.

\bibitem{ref102}
D.~J. Butler, J.~Wulff, G.~B. Stanley, and M.~J. Black, ``A naturalistic open
  source movie for optical flow evaluation,'' in \emph{European Conf. on
  Computer Vision (ECCV)}, ser. Part IV, LNCS 7577, {A. Fitzgibbon et al.
  (Eds.)}, Ed.\hskip 1em plus 0.5em minus 0.4em\relax Springer-Verlag, Oct.
  2012, pp. 611--625.

\bibitem{ref103}
D.~Kondermann, R.~Nair, K.~Honauer, K.~Krispin, J.~Andrulis, A.~Brock,
  B.~Gussefeld, M.~Rahimimoghaddam, S.~Hofmann, C.~Brenner \emph{et~al.}, ``The
  hci benchmark suite: Stereo and flow ground truth with uncertainties for
  urban autonomous driving,'' in \emph{IEEE Conf. Comput. Vis. Pattern
  Recognit.}, 2016, pp. 19--28.

\end{thebibliography}

\vfill

\end{document}


\title{Unsupervised Domain Adaption Harnessing Vision-Language Pre-training\\
	—Appendix—}

	\markboth{Journal of \LaTeX\ Class Files,~Vol.~14, No.~8, August~2021}%
	{Shell \MakeLowercase{\textit{et al.}}: A Sample Article Using IEEEtran.cls for IEEE Journals}
	
	\IEEEpubid{0000--0000/00\$00.00~\copyright~2021 IEEE}
	
	\maketitle
	
	\subsection{\textbf{Complement Hyper-Parameters Detailed in Different Benchmarks}}
	In this section, we will provide more comprehensive details about the experimental configurations and training tricks. 
	
	\textbf{Training Iterations}. We train 10,000 iterations for Office-Home \cite{ref41}, Office-31 \cite{ref42} with ImageCLEF-DA \cite{ref47} datasets; 20,000 iterations for Digits (including SVHN \cite{ref44}, MNIST \cite{ref45}, and USPS \cite{ref46}) with VISDA-2017 \cite{ref43} benchmark; and 40,000 iterations for DomainNet \cite{ref48} dataset.
	
	\textbf{Data Augmentation}. In our experiments, we adhered to the widely used settings of the relevant UDA methods and employed specific data augmentation designs for different benchmarks. For clarity and convenience, we describe the data augmentation operators using the PyTorch style.
	
	\noindent**************************************************
	
	\begin{center}
		\textbf{Office-Home, Office-31, ImageCLEF-DA and DomainNet}
	\end{center}
	
	\begin{flushleft}
		\textit{Testing: }
	\end{flushleft}
	
	Resize (224) - ToTensor () - Normalize (mean,std)
	
	\begin{flushleft}
		\textit{Training: }
	\end{flushleft}
	
	Resize (256) - RandomCrop (224) - RandomHorizontal-Flip() - ToTensor () - Normalize (mean,std)
	
	\begin{center}
		\textbf{VisDA-2017}
	\end{center}
	
	\begin{flushleft}
		\textit{Testing: }
	\end{flushleft}
	
	Resize (224) - CenterCrop (224) - ToTensor () - Normalize (mean,std)
	
	\begin{flushleft}
		\textit{Training: }
	\end{flushleft}
	
	Resize (224) - RandomHorizontalFlip () - CenterCrop (224) - ToTensor () - Normalize (mean,std)
	
	\begin{center}
		\textbf{Digits}
	\end{center}
	
	\begin{flushleft}
		\textit{Testing: }
	\end{flushleft}
	
	Lambda (lambda x: x.convert("RGB")) - Resize (224) - ToTensor () - Normalize (mean,std)
	
	\begin{flushleft}
		\textit{Training: }
	\end{flushleft}
	\begin{center}
		SVHN→MNIST
	\end{center}
	
	\begin{flushleft}
		\textit{SVHN}
	\end{flushleft}
	
	Lambda (lambda x: x.convert("RGB")) - RandomResizedCrop (size=224, scale=(0.75, 1.2)) - ToTensor () - Normalize (mean,std)
	
	\begin{flushleft}
		\textit{MNIST}
	\end{flushleft}
	
	Lambda (lambda x: x.convert("RGB")) - RandomResizedCrop (size=224, scale=(0.75, 1.2)) - ToTensor () - Normalize (mean,std)
	
	\begin{center}
		USPS→MNIST
	\end{center}
	
	\begin{flushleft}
		\textit{USPS}
	\end{flushleft}
	
	Lambda (lambda x: x.convert("RGB")) - Resize (28) -pad (4) - RandomCrop(28) - RandomRotation(10) - Resize(224) - ToTensor () - Normalize (mean,std)
	
	\begin{flushleft}
		\textit{MNIST}
	\end{flushleft}
	
	Lambda (lambda x: x.convert("RGB")) - RandomResizedCrop (size=224, scale=(0.75, 1.2)) - ToTensor () - Normalize (mean,std)
	
	\begin{center}
		MNIST→USPS
	\end{center}
	
	\begin{flushleft}
		\textit{MNIST}
	\end{flushleft}

	Lambda (lambda x: x.convert("RGB")) - RandomResizedCrop (size=224, scale=(0.75, 1.2)) - ToTensor () - Normalize (mean,std)
	
	\begin{flushleft}
		\textit{USPS}
	\end{flushleft}
	
	Lambda (lambda x: x.convert("RGB")) - RandomResizedCrop (size=224, scale=(0.75, 1.2)) - ToTensor () - Normalize (mean,std)
	~\\
	
	As we utilized CLIP \cite{ref8} as pre-training model, we adopted the mean and standard deviation values provided in the original paper, respectively:
\begin{center}
		$\text{mean} = (0.48145466, 0.4578275, 0.40821073)$
\end{center}
\begin{center}
		$\text{std} = (0.26862954, 0.26130258, 0.27577711)$
\end{center}
\noindent**************************************************
\IEEEpubidadjcol

\textbf{Trade-Off in CMKD}. In the majority of experiments, we set the values of ${{\lambda }_{1}}$ to 0.25, ${{\lambda }_{2}}$ to 0.1, and ${{\lambda }_{1}}$ to 0.025. However, in the case of Office-31, we set both ${{\lambda }_{2}}$ and ${{\lambda }_{1}}$ to 0. For the DomainNet dataset, which contains a large number of categories, requiring a lower weight for self-training, we set ${{\lambda }_{1}}$ to 0.025 and ${{\lambda }_{3}}$ to 0.0025.

\textbf{Predefined Threshold $\tau$ in RST}. The parameter $\tau$ in RST is directly linked to the average accuracy and DSP. Throughout our experiments, we discovered that the optimal value of $\tau$ varies with the backbone learning rate. Hence, in most experiments, we set $\tau$ to 1e-6. However, for the VisDA-2017 benchmark with ViT-B-16, $\tau$ is set to 1e-7, and for ResNet-50, $\tau$ is set to 1e-8.

\begin{figure}[t]
	\setlength{\abovecaptionskip}{0.cm}
	\setlength{\belowcaptionskip}{-0.cm}
	\centering
	\includegraphics[width=3.5in]{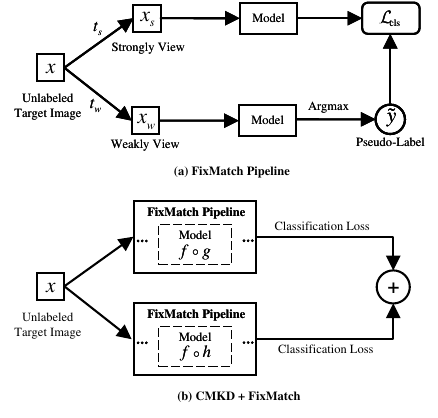}
	\caption{FixMatch pipeline and implementation in our paper. $t_s$ represents strong data augmentation, and $t_w$ represents weak data augmentation. $f$ serves as the visual encoder, $g$ as the text encoder, and $h$ as the task head.}
	\label{fig_1}
\end{figure}

\subsection{\textbf{FixMatch with CMKD}} 
In our experiments, we observe that CMKD, when combined with FixMatch \cite{ref69}, demonstrates more effective performance. We provide a detailed implementations and configurations of FixMatch\cite{ref69} in our method. FixMatch \cite{ref69} is a semi-supervised technique that applies strong and weak augmentations to unlabeled samples separately. Afterward, the model generates pseudo-labels for the weakly augmented samples with higher confidence to guide the training of the strongly augmented samples. Since CMKD requires using the teacher's perspective to guide the student's visual learning, when combined with FixMatch, the training process needs to be performed separately based on the teacher's and student's viewpoints, as illustrated in Figure 1. The objective function of FixMatch can be defined as:
\begin{eqnarray}
	\lambda_{f m} \cdot \mathbbm{1}_{\max \left(p_{w}\right)>\tau_{f m}} \arg \max \left(p_{w}\right) \cdot \log \left(p_{s}\right)
\end{eqnarray}
where $\lambda_{f m}$ represents the trade-off, $\mathbbm{1}$ is the indicator function, $\tau_{f m}$ represents the predefined threshold, $p_{w}$ denotes the prediction of weakly view, and $p_{s}$ denotes the prediction of strongly view. In most benchmark, $\lambda_{f m}$ is set to 0.5 and $\tau_{f m}$ to 0.95, whereas on the VisDA-2017 dataset, $\lambda_{f m}$ is set to 2.0 and $\tau_{f m}$ to 0.80, and on the DomainNet dataset, $\lambda_{f m}$ is set to 0.1 and $\tau_{f m}$ to 0.95.

For weak augmentation, we employ the data augmentation methods referenced in Section A that are specific to the corresponding benchmarks. For strong augmentation, our specific configuration is as follows:

\noindent**************************************************

Lambda (lambda x: x.convert("RGB")) - Resize (224) - CenterCrop(224) - RandomHorizontalFlip() - ColorJitter (brightness = 0.4, contrast = 0.4, saturation = 0.4, hue = 0.0) - RandAugment('rand-n2-m10-mstd0.5') - ToTensor () - Normalize (mean, std)
~\\

``RandAugment'' \cite{ref72} is a method of automated data augmentation, and 'rand-n2-m10-mstd0.5' is the specific hyper-parameter setting.

\noindent**************************************************

\subsection{\textbf{Relevant Model Parameters}} 
In this paper, we consider the number of parameters necessary for deploying downstream tasks for the first time. Therefore, in this subsection, we provide a separate presentation of the parameter counts for both the backbone $f(\cdot )$ and task-head $h(\cdot )$ of the model. This will facilitate the computation of the DSP, as demonstrated in Table I.

\begin{table}[h]
	\setlength{\abovecaptionskip}{0cm}  
	\setlength{\belowcaptionskip}{-0.2cm} 
	\centering
	\caption{Number of parameters for different models.}
	\setlength{\tabcolsep}{2.5mm}{
			\begin{tabular}{c|p{10.5em}|p{10.69em}|p{8.375em}}
				\toprule
				\multicolumn{1}{p{6.69em}|}{\centering Model} &  \multicolumn{1}{c|}{Pre-trained} &  \multicolumn{1}{c|}{Backbone} & \multicolumn{1}{c}{Task-Head} \\
				\midrule
				\multicolumn{1}{c|}{\multirow{2}[4]{*}{ResNet-50 \cite{ref17}}}  & \multicolumn{1}{c|}{ImageNet \cite{ref10}} & \multicolumn{1}{c|}{23508032 (23.5M)} & \multicolumn{1}{c}{2048*c} \\
				\cmidrule{2-4}          & \multicolumn{1}{c|}{CLIP \cite{ref12}}  & \multicolumn{1}{c|}{38316896 (38.3M)} & \multicolumn{1}{c}{1024*c} \\
				\midrule
				\multicolumn{1}{c|}{\multirow{2}[4]{*}{ResNet-101 \cite{ref17}}} & \multicolumn{1}{c|}{ImageNet \cite{ref10}} & \multicolumn{1}{c|}{42500160 (42.5M)} & \multicolumn{1}{c}{2048*c} \\
				\cmidrule{2-4}          & \multicolumn{1}{c|}{CLIP \cite{ref12}}  & \multicolumn{1}{c|}{56259936 (56.3M)} & \multicolumn{1}{c}{512*c} \\
				\midrule
				\multicolumn{1}{c|}{\multirow{2}[4]{*}{ViT-B \cite{ref18}}} & \multicolumn{1}{c|}{ImageNet \cite{ref11}} & \multicolumn{1}{c|}{85798656 (85.8M)} & \multicolumn{1}{c}{768*c} \\
				\cmidrule{2-4}          & \multicolumn{1}{c|}{CLIP \cite{ref12}}  & \multicolumn{1}{c|}{86192640 (86.2M)} & \multicolumn{1}{c}{512*c} \\
				\bottomrule
			\end{tabular}%
	}
	\label{tab2}%
\end{table}%

The design of the task head varies across different papers. Due to the relatively low number of parameters associated with the task head, for the sake of convenience, we assume that all task head are computed as 1-layer fully connected networks. Furthermore, when it comes to the same Backbone, there are slight distinctions between the ImageNet pre-training model and the CLIP pre-training model.

\subsection{\textbf{Classifier Configuration}} 
Our classifier is intentionally designed to be lightweight while maintaining strong performance in comparison to existing classifiers. Taking inspiration from CDTrans \cite{ref7}, we've structured our classifier network as ``BatchNorm(dim)-LayerNorm(dim)-FFN(dim, c)'', where ``BatchNorm'' refers to Batch Normalization \cite{ref74}, ``LayerNorm'' refers to Layer Normalization \cite{ref75}, and ``FFN'' stands for Feed Forward Networks. Regarding the initialization of the classifier, we initialize the FFN using a normal distribution with a standard deviation set to 0.001. For Batch Normalization, we set the weight of the affine layer to 1.0 and the bias to 0.0 during initialization. The bias remains fixed throughout the training process.

\begin{table*}[t]
	\setlength{\abovecaptionskip}{0.cm}
	\setlength{\belowcaptionskip}{-0.cm}
	\centering
	\caption{Comparison with SoTA UDA methods on VisDA-2017 using ImageNet pre-trained models. $^o$ implies its pre-trained from on ImageNet-21K instead of ImageNet-1K. $^\star$ is pre-trained from CLIP. The best performance is marked as bold.}
	\setlength{\tabcolsep}{1.8mm}{
		\begin{tabular}{c|cccccccccccccc}
			\toprule
			Method & plane & bcycl & bus & car & horse & knife & mcycl & person & plant & sktbrd & train & truck & Avg. & DSP(M)\\
			\midrule
			\makecell[l]{\textit{ResNet101:}} &       &       &       &       &       &       &       &       &       &       &       &       &  \\
			CDAN+E & 85.2  & 66.9  & 83.0  & 50.8  & 84.2  & 74.9  & 88.1  & 74.5  & 83.4  & 76.0  & 81.9  & 38.0  & 73.9  & 42.52\\
			DSAN  & 90.9  & 66.9  & 75.7  & 62.4  & 88.9  & 77.0  & 93.7  & 75.1  & 92.8  & 67.6  & 89.1  & 39.4  & 75.1  & 42.52\\
			BNM   & 89.6  & 61.5  & 76.9  & 55.0  & 89.3  & 69.1  & 81.3  & 65.5  & 90.0  & 47.3  & 89.1  & 30.1  & 70.4  & 42.52\\
			MSTN+DSBN & 94.7  & 86.7  & 76.0  & 72.0  & 95.2  & 75.1  & 87.9  & 81.3  & 91.1  & 68.9  & 88.3  & 45.5  & 80.2  & 42.52\\
			CGDM  & 92.8  & 85.1  & 76.3  & 64.5  & 91.0  & 93.2  & 81.3  & 79.3  & 92.4  & 83.0  & 85.6  & 44.8  & 80.8  & 42.52\\
			SHOT  & 95.5  & 87.5  & 80.1  & 54.5  & 93.6  & 94.2  & 80.2  & 80.9  & 90.0  & \textbf{89.9}  & 87.1  & 58.4  & 82.7  & 42.52\\
			CDAN+MCC & 94.5  & 80.8  & 78.4  & 65.3  & 90.6  & 79.4  & 87.5  & 82.2  &\textbf{94.7}  & 81.0  & 86.0  & 44.6  & 80.4  & 42.52\\
			DAPL$^\star$  & \textbf{97.8}  & 83.1  & \textbf{88.8}  & 77.9  & \textbf{97.4}  & 91.5  & \textbf{94.2}  & 79.7  & 88.6  & 89.3  & 92.5  & \textbf{62.0}  & \textbf{86.9}  & 42.52\\
			\midrule
			Baseline  & 90.6      & 64.1      & 51.3      & \textbf{83.7}      & 66.2      & 16.5      & 80.6      & 20.9      & 66.9      & 31.1      & \textbf{94.7}      & 5.7      & 56.0 &42.52\\
			CMKD  &  95.6     & 88.0      & 73.2      & 67.9      & 92.3      & 92.4      & 80.0      & 82.3      & 93.3      & 82.9      & 87.7      & 50.8      &82.2  & 42.52\\
			CMKD+RST & 94.3      & 83.8      & 72.7      & 70.8      & 91.8      & 91.9      & 85.5      & \textbf{83.4}      & 91.0      & 82.1      & 89.3      & 41.5      &  81.5 & \textbf{0.52}\\
			CMKD+FixMatch & 96.4      & \textbf{88.5}      & 82.5      & 75.6      & 96.0      & \textbf{95.7}      & 90.2      & 79.9      & 94.3      & 89.4      & 87.9      & 46.1      & 85.2 & 42.52\\
			\midrule
			\midrule
			\makecell[l]{\textit{ViT-B-16:}} &       &       &       &       &       &       &       &       &       &       &       &       &  \\
			CDTrans & 97.1  & 90.5  & 82.4  & 77.5  & 96.6  & 96.1  & 93.6  & \textbf{88.6}  & \textbf{97.9}  & 86.9  & 90.3  & 62.8  & 88.4  &85.81\\
			TVT$^o$   & 92.9  & 85.6  & 77.5  & 60.5  & 93.6  & 98.2  & 89.4  & 76.4  & 93.6  & 92.0  & 91.7  & 55.7  & 83.9  &85.81\\
			SDAT$^o$  & 98.4  & 90.9  & 85.4  & 82.1  & 98.5  & 97.6  & 96.3  & 86.1  & 96.2  & 96.7  & 92.9  & 56.8  & 89.8  &85.81\\
			SSRT-B$^o$ & 98.9  & 87.6  & \textbf{89.1}  & \textbf{84.7}  & 98.3  & \textbf{98.7}  & 96.2  & 81.0  & 94.8  & 97.9  & 94.5  & 43.1  & 88.7  &85.81\\
			PMTrans$^o$ & 98.9  & 93.7  & 84.5  & 73.3  & \textbf{99.0}  & 98.0  & 96.2  & 67.8  & 94.2  & \textbf{98.4}  & 96.6  & 49.0  & 87.5 &85.81 \\
			\midrule
			Baseline$^o$ & \textbf{99.1}      & 73.6      & 82.5      & 68.6      & 93.5      & 84.5      & \textbf{96.9}      & 13.4      & 70.2      & 96.5      & \textbf{97.5}      & 21.6      & 74.8  &85.81\\
			CMKD $^o$ &98.8  &92.9  &81.7  &74.6 &96.9 &95.5 &91.4 &72.1  &93.5   &96.4   &95.5   &54.7   &87.0  &85.81\\
			CMKD+RST$^o$ & 98.7      & 93.4      & 81.7      & 72.3      & 97.3      & 96.9    & 92.7      & 69.9      & 89.5      & 97.5      & 96.4      & 57.2      & 87.0 & \textbf{0.62} \\
			CMKD+FixMatch$^o$ &  98.8     & \textbf{96.3}      & 85.6      & 78.0      & \textbf{99.0}      &97.4       & 91.5      & 75.3      &  96.7     & 98.0      & 96.2      &  \textbf{70.2}     & \textbf{90.3}   &85.81\\
			\bottomrule
		\end{tabular}%
	}
	\label{tab5}%
\end{table*}%

\subsection{\textbf{CMKD Generalize to ImageNet Pre-trained Models}} 
In this paper, our aim is to harness the capabilities of visual language pre-training models to tackle two significant challenges within unsupervised domain adaptation. Specifically, CMKD employs CLIP's zero-shot inference as a teacher's perspective, guiding the model in learning from unlabeled data, and achieving state-of-the-art results across multiple benchmarks. On the other hand, RST effectively alleviates the deployment burden in UDA by training highly sparse weights for the sub-tasks. Nevertheless, despite numerous experiments that validate the efficacy of the proposed methods, their performance under a conventionally pre-trained ImageNet model remains uncertain. To delve deeper into the effectiveness and adaptability of our approach, this subsection is dedicated to implementing our method based on the ImageNet pre-trained model.

We first begin by reviewing the CMKD paradigm introduced in this paper, which can be described as follows:
\begin{eqnarray}
	{{\mathcal{L}}_{\text{cmkd}}}={{\lambda }_{1}}\cdot (\alpha \cdot {{\mathcal{L}}_{\text{task}}}+(1-a)\cdot {{\mathcal{L}}_{\text{distill}}})+{{\mathcal{L}}_{\text{reg}}}
\end{eqnarray}
\begin{eqnarray}
	{{\mathcal{L}}_{\text{task}}}=\text{GI}(p_{h}^{t})=1-\sum_{i=1}^{c}{[{{p_{h}^{t}(y=i|{{x}^{t}})}]^{2}}}
\end{eqnarray}
\begin{eqnarray}
	\mathcal{L}_{\text {distill }}=\mathrm{GI}\left(p_{m}^{t}\right)=1-\sum_{i=1}^{c}\left[p_{m}^{t}\left(y=i \mid x^{t}\right)\right]^{2}
\end{eqnarray}
\begin{eqnarray}
	p_{m}^{t}=0.5 \cdot\left(p_{h}^{t}+\textit{\text{sg}}\left(p_{\text{g}}^{t}\right)\right)
\end{eqnarray}
\begin{eqnarray}
	\mathcal{L}_{\text {reg }}=\lambda_{2} \cdot \operatorname{KL}\left(y|| p_{\text{g}}^{s}\right)+\lambda_{3} \cdot \operatorname{GI}\left(p_{\text{g}}^{t}\right)
\end{eqnarray}
Based on the equation above, it becomes evident that the primary challenge when applying CMKD using ImageNet as a foundation is the absence of the teacher's perspective $p_{\text{g}}$ provided by the pre-trained model. Taking inspiration from Co-tuning \cite{ref73}, we adopt a strategy wherein we learn the class cluster centers from the output of the ImageNet classifier, resulting in the acquisition of class cluster centers denoted as $M$. In particular, we partition the source domain dataset into a training set and a validation set following a 1:1 ratio. Subsequently, we derive the prototype $M$ through the process of category relationship learning \cite{ref73}. Subsequently, we compute the sample probabilities based on the discrepancy associated with each class cluster center, effectively using it as the teacher's viewpoint. This enables the realization of CMKD on the ImageNet pre-trained model. Thus, $p_{\text{g}}$ can be written as:
\begin{eqnarray}
	\begin{aligned}
		p_{\text{g}}(y=i|x)=\frac{\exp \left(-\mathrm{KL}\left(M_{i} \| l(f(x)) \right)\right)}{\sum_{k=1}^{c} \exp \left(-\mathrm{KL}\left(M_{k} \| l(f(x))\right)\right)}
	\end{aligned}
\end{eqnarray}
where $l(\cdot)$ denotes ImageNet classifier.

Based on the ImageNet pre-trained model, we verify the effectiveness of CMKD with RST on VisDA benchmark. Regarding the hyperparameter configuration, we set ${\lambda}_{1}$ to 0.25, ${\lambda}_{2}$ to 0.0, ${\lambda}_{3}$ to 0.0 and $\tau$ to 1e-6. For normalization in the preprocessing step, the mean is set to (0.485, 0.456, 0.406) and std is set to (0.229, 0.224, 0.225). As evident from Table II, our approach can be effectively applied to ImageNet pre-trained models as well. When employed in conjunction with the ViT-B model, we achieved a result of \textbf{90.3\%} when combined with FixMatch \cite{ref69}, surpassing the previous SoTA (SDAT) by an increment of \textbf{+0.5\%}. Moreover, when integrated with RST, merely \textbf{0.62M} parameters are required to attain equivalent performance as CMKD, thereby significantly reducing the parameter count necessary for downstream task deployment. Regarding ResNet-50, even though it trails behind the current state-of-the-art (DAPL) by \textbf{-1.7\%}, it still demonstrates remarkable performance when contrasted with ImageNet-based pre-trained modeling approaches. It outperforms SHOT, CGDM, and CDAN+MCC by margins of \textbf{+2.5\%}, \textbf{+4.4\%}, and \textbf{+4.8\%}, respectively. Furthermore, with the application of RST, a mere \textbf{0.52M} parameters are required to attain performance comparable to CMKD, resulting in a negligible drop of just \textbf{-0.7\%}.

In the experiments, the CMKD paradigm showcased its ability to generalize to UDA tasks, offering promising outcomes. Conversely, RST yielded unexpected results. Initially, we held the belief that RST could be applied to downstream tasks by acquiring knowledge from ultra-sparse weights due to the extensive pre-training of visual language pre-training models. However, when put into practice using the ImageNet pre-trained model, RST also led to a notable reduction in the DSP. This prompts us to eagerly anticipate the future trajectory of RST and its potential advancements. We're hopeful that RST's paradigm can be further expanded to encompass additional tasks. Its ability to demonstrate effectiveness across diverse domains such as object detection, semantic segmentation, etc., solely utilizing the ImageNet pre-trained model, is a tantalizing prospect.

\begin{table}[t]
	\setlength{\abovecaptionskip}{0cm}  
	\setlength{\belowcaptionskip}{-0.0cm} 
	\centering
	\caption{UDA methods performed with image encoder of VLP on VisDA-2017}
	\renewcommand{\arraystretch}{1.0}
	\setlength{\tabcolsep}{2.0 mm}{
		\begin{tabular}{c|ccccccc}
			\toprule
			Method & Origin Hyperparameters & Ours Hyperparameters \\
			\midrule
			\makecell[l]{\textit{ResNet101:}} &       &  \\
			BNM$^\star$   & 61.1 (\textcolor{green}{-9.3})  & 68.5 (\textcolor{green}{-1.9})  \\
			CGDM$^\star$  & 73.1 (\textcolor{green}{-7.3})  & 80.9 (\textcolor{red}{+0.1})  \\
			SHOT$^\star$  & 72.3 (\textcolor{green}{-10.4})  & 81.6 (\textcolor{green}{-1.1})\\
			CMKD$^\star$ (Ours) & - & \textbf{87.0} \\
			\midrule
			\makecell[l]{\textit{ViT-B-16}}  &       &   \\
			CDTrans$^\star$ & 76.4 (\textcolor{green}{-12.0}) & 86.5 (\textcolor{green}{-1.9}) \\
			SSRT$^\star$ & 77.1 (\textcolor{green}{-11.6}) & 87.0 (\textcolor{green}{-1.7})\\
			PMTrans$^\star$ &77.8 (\textcolor{green}{-9.7}) & 85.0 (\textcolor{green}{-2.5}) \\
			CMKD$^\star$ (Ours) & - & \textbf{90.3} \\
			\bottomrule
		\end{tabular}%
	}
	\label{table6}%
\end{table}%

\subsection{\textbf{Other UDA methods using VLP pre-trained models}}
In prior Unsupervised Domain Adaptation tasks, the conventional practice involves utilizing ImageNet-1k or ImageNet-21k as initial weights for training. However, this paper introduces the VLP model into UDA training. To ensure a fair comparison, we train based on CLIP's Image Encoder and commonly used UDA methods to validate the effectiveness of CMKD. Throughout the experiment, we maintain two sets of hyperparameters, where ``Origin Hyperparameter'' represents the original paper's parameters. Notably, due to CLIP's sensitivity to hyperparameters, we additionally employ the ``Ours Hyperparameter'' explored in this paper (refer to Section IV in the main text for detailed insights). 

From Table III, it's evident that CMKD effectively leverages both the visual and text encoders of the CLIP model, substantially improving UDA performance and achieving notable results. However, when applying other methods to CLIP's visual encoder, using both hyperparameter sets 1 and 2, we observed a performance decrease compared to models trained on ImageNet pre-training weights. Our analysis suggests two potential reasons: firstly, it could be attributed to the challenge of finding optimal hyperparameters; secondly, traditional UDA methods focus on aligning feature spaces between the source and target domains, potentially disrupting the distribution of the pre-trained model's feature space. Consequently, there remains significant room for exploration in leveraging VLP models to further enhance UDA tasks.

\begin{table}[t]
	\setlength{\abovecaptionskip}{0cm}  
	\setlength{\belowcaptionskip}{-0.0cm} 
	\centering
	\caption{UDA methods combined with RST on Office-Home}
	\renewcommand{\arraystretch}{1.0}
	\setlength{\tabcolsep}{2.0 mm}{
		\begin{tabular}{c|ccccccc}
			\toprule
			Method & Avg. & DSP (M) \\
			\midrule
			\makecell[l]{\textit{ResNet101:}} &       &  \\
			BNM   & 68.2 (\textcolor{red}{+0.3})  & 2.11 (\textcolor{green}{-99.25\%})  \\
			DSAN  & 67.1 (\textcolor{green}{-0.5})  & 1.98 (\textcolor{green}{-99.3\%})  \\
			SHOT  & 71.3 (\textcolor{green}{-0.5})  & 2.83 (\textcolor{green}{-99.0\%})\\
			\midrule
			\makecell[l]{\textit{ViT-B-16}}  &       &   \\
			CDTrans$^o$ & 80.1 (\textcolor{green}{-0.4}) & 10.3 (\textcolor{green}{-99.0\%}) \\
			SSRT$^o$ & 84.7 (\textcolor{green}{-0.8}) & 9.27 (\textcolor{green}{-99.1\%})\\
			PMTrans$^o$ &87.3 (\textcolor{green}{-1.4}) & 6.18 (\textcolor{green}{-99.4\%}) \\
			\bottomrule
		\end{tabular}%
	}
	\label{table6}%
\end{table}%

\begin{table}[t]
	\setlength{\abovecaptionskip}{0cm}  
	\setlength{\belowcaptionskip}{-0.0cm} 
	\centering
	\caption{Different distillation terms on Office-Home using ResNet-50.}
	\renewcommand{\arraystretch}{1.5}
	\setlength{\tabcolsep}{6.0 mm}{
		\begin{tabular}{c|ccccccc}
			\toprule
			Distillation term & Avg. \\
			\midrule
			$\text{KL(\emph{sg}(}p_{\text{g}}^{t})||p_{h}^{t}\text{)}$ & 76.5 \\
			$\mathrm{GI}\left(p_{m}^{t}\right)$   & \textbf{79.3}  \\
			\bottomrule
		\end{tabular}%
	}
	\label{table6}%
\end{table}%

\subsection{\textbf{Other UDA methods using RST}}
RST offers schemes aimed at significantly diminishing the model storage overhead. In this subsection, we validate the compatibility of RST with other UDA methods. The pre-defined thresholds $\tau$ associated with RST are intricately linked to the learning rate, thus, we set tau to 1e-5 in this section. To visually highlight the parameter reduction differences, we'll opt for the office-home benchmark, which encompasses a broader range of tasks for our experiments. From Table IV, it's evident that despite varying degrees of model performance reduction post-RST implementation, there's a notable decrease in DSP. This reduction effectively alleviates the storage strain on the model.

\begin{table}[t]
	\setlength{\abovecaptionskip}{0cm}  
	\setlength{\belowcaptionskip}{-0.0cm} 
	\centering
	\caption{Different trade-off of CMKD on Office-Home using ResNet-50.}
	\renewcommand{\arraystretch}{1.2}
	\setlength{\tabcolsep}{6.0 mm}{
		\begin{tabular}{c|ccccccc}
			\toprule
			$({\lambda }_{1}, {\lambda }_{2}, {\lambda }_{3})$ & Avg. \\
			\midrule
			(0.05, 0.02, 0.005) & 78.1 \\
			(0.25, 0.1, 0.025) & \textbf{79.3} \\
			(0.50, 0.2, 0.05) & 77.3 \\
			\bottomrule
		\end{tabular}%
	}
	\label{table6}%
\end{table}%

\subsection{\textbf{Further Hyperparameters Discussion}}
In this section, our primary focus was on examining CMKD's distillation term and the associated trade-off configurations. We contend in the paper that the teacher model (CLIP, in this case) doesn't consistently outperform, and the alignment of student and teacher distributions via KL divergence might negatively impact model performance. This is validated in the experimental results showcased in Table V, where the performance experiences a considerable drop compared to $\mathrm{GI}\left(p_{m}^{t}\right)$.

Furthermore, various hyperparameter configurations in CMKD are deliberated upon. Table VI illustrates that the most optimal performance is attained at the settings of (0.25, 0.1, 0.025).

\subsection{\textbf{Convergence and Stability of Empirical Results}}
\begin{figure*}[t]
	\setlength{\abovecaptionskip}{0.cm}
	\setlength{\belowcaptionskip}{-0.cm}
	\centering
	\includegraphics[width=7in]{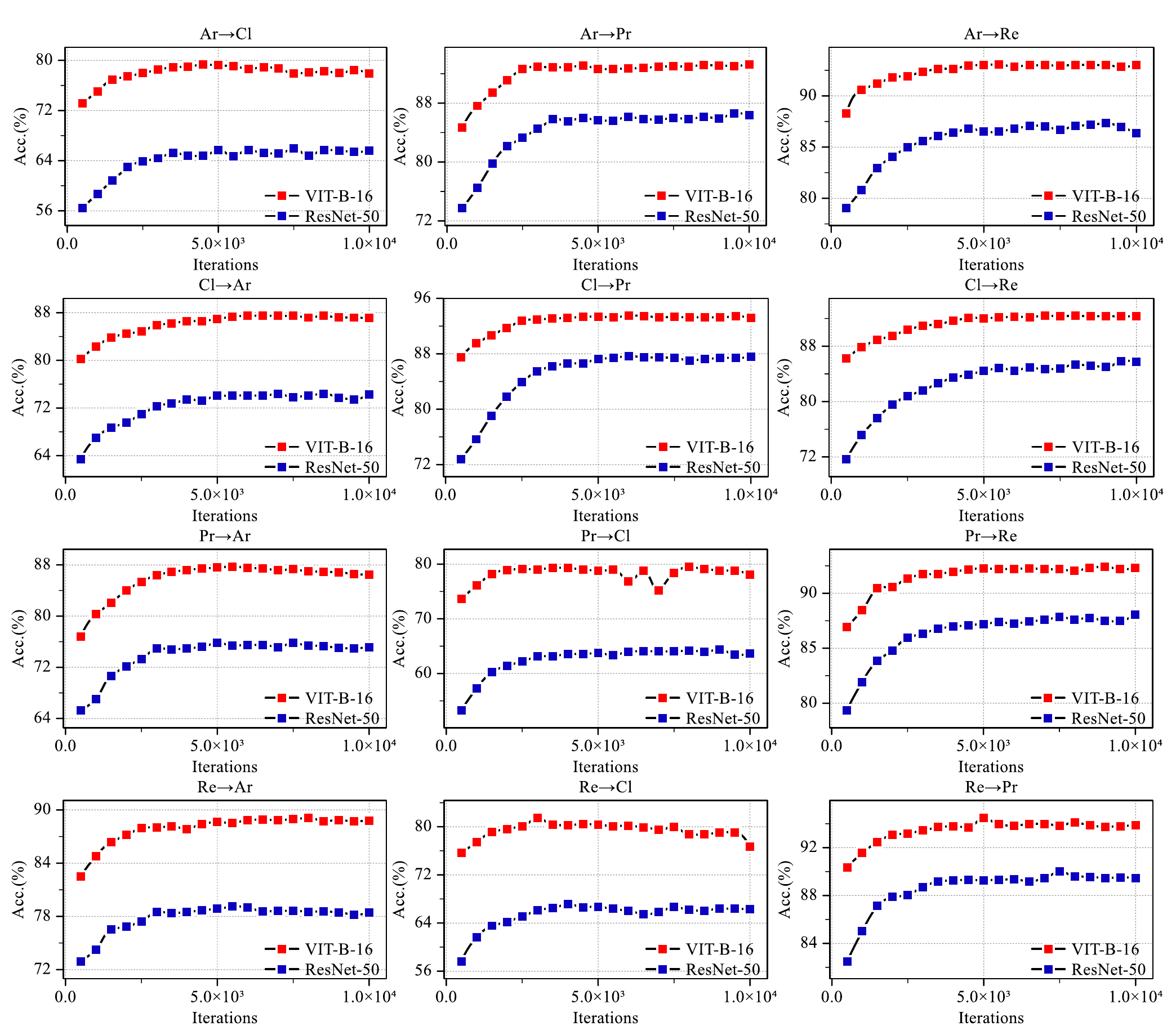}
	\caption{Validation Accuracy across iterations on different splits of Office-Home.}
	\label{fig_2}
\end{figure*}
To further substantiate the training stability and convergence speed of our approach, we measure the accuracy of their training processes using the ResNet-50 and ViT-B-16 architectures on the Office-Home Benchmark. As depicted in Figure 2, the training process for all 12 subtasks displays remarkable smoothness, and the validation results throughout the training exhibit minimal fluctuations. This consistency suggests a high level of training stability inherent in our method. Additionally, the curves within the graph demonstrate the impressive convergence capability of our approach, characterized by swift convergence followed by a sustained smooth phase during the initial training stages.
\subsection{\textbf{RST with ImageNet Pre-trained Model on Segmentation}}
\begin{table*}[t]
	\setlength{\abovecaptionskip}{0.cm}
	\setlength{\belowcaptionskip}{-0.cm}
	\centering
	\caption{Comparison with SoTA PEFT methods on GTA$\rightarrow$Cityscape.}
	\renewcommand{\arraystretch}{1.0}
	\setlength{\tabcolsep}{0.5mm}{
		\begin{tabular}{c|ccccccccccccccccccccc}
			\toprule
			Method &Road &S.walk &Build. &Wall &Fence &Pole &Tr.Light &Sign &Veget. &Terrain &Sky &Person &Rider &Car &Truck &Bus &Train &M.bike &Bike &mIOU & DSP(M)\\
			\midrule
			Fine-Tuning & \textbf{95.7} & \textbf{70.2} & 87.8 &36.2 & 35.2 & \textbf{38.3}& 50.2 & 53.8 &88.1 & 45.5 & \textbf{88.3} &\textbf{69.6} &\textbf{43.7} &87.7 &49.4 &50.2 &0.0 &24.9 &48.3 &56.0 &42.51 \\
			Linear-Probe & 76.6 & 11.1 & 60.7 & 1.67 & 0.0 & 15.3 & 0.9 & 0.0 & 59.6 & 6.0 & 59.9 & 0.8 & 2.4 & 43.9 & 1.3 & 1.4 & 0.1 & 4.6 & 3.8 & 18.4 & \textbf{0.70}\\
			BitFit & 83.5 & 8.2 & 75.9 & 11.8 & 1.5 & 21.4 & 8.6 & 3.2 & 75.7 & 14.9 &79.0 & 29.8 & 10.8 & 72.4 & 18.8 & 5.7 & 0.1 & 10.8 & 4.0 & 28.3 & \textbf{0.75}\\ 
			LoRA & 85.6 & 34.5 & 80.8 & 24.8 & 20.7 & 25.6 & 26.6 & 38.8 & 83.6 & 35.1 & 81.3 & 58.1 & 28.2 & 75.5 & 31.8 & 37.9 & 23.4 & \textbf{26.1} & 45.8 & 45.5 & 5.39\\
			\midrule
			RST$_{config1}$ & 93.8  & 58.9 & 86.8& 32.0& 32.6 & 35.8 &42.6 & 49.2 &87.4 & 41.8 &87.5 &65.1 &33.1 &87.0 &49.5 &49.0  &0.7  &25.7 &\textbf{50.8} &53.1& 5.88\\
			RST$_{config2}$ & 94.0 & 57.3 & \textbf{88.4} & \textbf{39.2} & \textbf{36.6} & 37.5 & 49.4 & 54.3 & \textbf{88.4} & \textbf{47.5} & 88.1 & 69.4 & 43.2 & \textbf{88.7} &\textbf{58.2} & \textbf{52.8} & \textbf{13.7} & 24.6 & 45.2 & \textbf{56.7} & 12.6 \\
			\bottomrule
		\end{tabular}%
	}
	\label{tab3}%
\end{table*}%

Unsupervised Domain Adaptive Semantic Segmentation (UDASS) presents a common challenge in unsupervised domain adaptation, alongside classification tasks. This task mirrors the setup of classification tasks within UDA, where the model is trained on labeled source domain data to facilitate object segmentation in the unlabeled target domain. In order to further demonstrate the versatility of RST, we opted to compare it with various PEFT methods, notably BitFit \cite{ref70}, LoRA \cite{ref14}, etc., on the GTA\cite{ref89}$\rightarrow$Cityscape\cite{ref90} benchmark, which is a standard task in UDASS. Given that in the UDASS task, the pre-training weights are derived from the ImageNet classification task, and these pre-training features may not adequately align with the representations required for downstream dense detection tasks. Therefore, by setting approximately 10\% of the trainable parameters and setting the rank of LoRA to 16, we aimed to address this mismatch.

For network architecture, we utilize the DeepLab-V2 \cite{ref91} with ResNet101 \cite{ref17} as the backbone. As a standard practice, we adhere to the self-training approach outlined in DAFormer \cite{ref92} along with its associated training parameters. Specifically, we employ AdamW optimization with a learning rate of $6 \times 10^{-5}$ for the encoder and $6 \times 10^{-4}$ for the decoder, utilizing a batch size of 2. Additionally, we incorporate linear learning rate warmup, and maintain a momentum parameter 0.999. For RST, we have introduced a slight modification. Instead of using a stable predefined threshold, we now use a value that changes over time. We set up two sets of configurations, with the threshold for \textit{config1} ranging from $5 \times 10^{-4}$ to $3.5 \times 10^{-2}$ as the number of training rounds increases, and the threshold for \textit{config2} ranging from $5 \times 10^{-4}$ to $2 \times 10^{-2}$. The trainable parameters for \textit{config1} are approximately 10\%, mainly for fair comparison with LoRA; \textit{Config2}'s trainable parameters are about 20\%, mainly for comparison with Fine-Tuning. This dynamic adjustment mechanism mirrors the approach outlined in the implementation details provided in the manuscript. The rationale behind this adjustment lies in the divergence between the pre-training task and the downstream task. Given the significant differences between the two, we believe that involving more parameters in the initial stages of training allows for a more comprehensive assessment of which parameters should be retained and which ones should be reset back to the pre-training weights throughout the training process. This adaptive approach ensures better alignment between the model's learned representations and the requirements of the downstream task.

Table VII presents intriguing experimental findings:

(1) In the context of Unsupervised Domain Adaptive Segmentation, utilizing ImageNet classification pre-training models leads to a significant gap between the pre-training task and the downstream task. Consequently, methodologies like Linear-Probe and BitFit \cite{ref70}, despite training very few parameters, exhibit a substantial performance degradation compared to Fine-Tuning. However, approaches like LoRA and RST can adjust the trainable parameters based on task difficulty, resulting in superior performance.

(2) Under Configuration 1, both RST and LoRA train approximately 10\% of the parameters. Remarkably, RST surpasses LoRA by +7.6\% mIOU, with only a slight difference of -2.8\% mIOU compared to Fine-Tuning. This highlights two key points: firstly, despite a significant feature gap between the pre-trained and downstream tasks, a substantial number of parameters in the model remain beneficial for the downstream task. Secondly, in the scenario of smaller models, RST outperforms LoRA \cite{ref14} with the same number of trainable parameters, showcasing the versatility of RST across tasks like sense prediction.

(3) An intriguing experiment involves increasing the number of trainable parameters based on Config2, resulting in the model training 20\% of the parameters compared to full parameter fine-tuning. Surprisingly, RST outperforms Fine-Tuning by +0.7 mIOU despite a greater number of trainable parameters. This suggests that RST not only excels in reducing model deployment complexity but also holds potential as a novel regularization technique capable of enhancing downstream task performance.

\subsection{\textbf{Partial-set Domain Adaptation}}
\begin{table*}[htbp]
	\setlength{\abovecaptionskip}{0cm}  
	\setlength{\belowcaptionskip}{-0.0cm} 
	\centering
	\caption{Comparison with SoTA methods on Office-Home (65→25) for partial-set UDA.}
	\renewcommand{\arraystretch}{0.9}
	\setlength{\tabcolsep}{0.5mm}{
		\begin{tabular}{c|cccccccccccccc}
			\toprule
			Method & {Ar→Cl} & {Ar→Pr} & {Ar→Re} & {Cl→Ar} & {Cl→Pr} & {Cl→Re} & {Pr→Ar} & {Pr→Cl} & {Pr→Re} & {Re→Ar} & {Re→Cl} & {Re→Pr} & {Avg.}  & DSP(M) \\
			\midrule
			\makecell[l]{\textit{ResNet50:}} &    &    &     &   &       &       &       &       &       &       &       &       &  &\\
			DRCN  &54.0 &76.4 &83.0 &62.1 &64.5 &71.0 &70.8 &49.8 &80.5 &77.5 &59.1 &79.9 &69.0 & 283.60\\
			BA$^{\text{3}}$US  &60.6 &83.2 &88.4 &71.8 &72.8 &83.4 &75.5 &61.6 &86.5 &79.3 &62.8 &86.1 &76.0 & 283.60\\
			TSCDA  &63.6 &82.5 &89.6 &73.7 &73.9 &81.4 &75.4 &61.6 &87.9 &83.6 &67.2 &88.8 &77.4 & 283.60\\
			SHOT 	 &64.6 &85.1 &92.9 &78.4 &76.8 &86.9 &79.0 &65.7 &89.0 &81.1 &67.7 &86.4 &79.5& 283.60\\
			SHOT++ 	&65.0 &85.8 &\textbf{93.4} &78.8 &77.4 &\textbf{87.3} &79.3 &66.0 &89.6 &81.3 &68.1 &86.8 &79.9 & 283.60\\
			\midrule
			CLIP$^\star$ &  63.2  & 84.3   &  86.9   & 76.4  &  84.3     & 86.9      & 76.4    & 63.2      & 86.9      &  76.4     & 63.2      &  84.3     & 77.7  & \textbf{0.00}\\
			Baseline$^\star$ & 58.4    & 75.2    & 82.7     & 61.9   & 67.7      &  72.5     & 66.0       & 56.0       & 80.5       &  75.4     & 60.0       & 81.5      & 69.8  & 460.4\\
			CMKD$^\star$ & 72.4   & 87.5    & 90.9    & 78.1   & 80.0      &  85.4     & 79.2      &  70.3     & 89.8      & 87.6      &71.9       & 89.6      & 81.9  & 460.4\\
			CMKD+RST$^\star$ & 68.3    &\textbf{89.9}    & 91.4     & 78.0  &  \textbf{81.5}    & 85.3      &77.8       & 70.0       & \textbf{91.5}      &85.7       & 70.4      &    89.5   & 81.6  & \textbf{0.98}\\
			CMKD+FixMatch$^\star$ & \textbf{73.1}    & 87.7    &  91.8   & \textbf{80.3}   & 79.8      &  86.1     &  \textbf{80.0}     & \textbf{74.4}      & 91.1      & \textbf{88.7}      &\textbf{74.9}       &  \textbf{89.9}     & \textbf{83.1}  & 460.40\\
			\midrule
			\midrule
			\makecell[l]{\textit{ViT-B-16:}} &       &       &       &       &       &       &       &       &       &       &       &       &  &\\
			CLIP$^\star$ & 79.5    & 89.3   &  91.3   & 84.3  & 89.3       &  91.3     & 84.3      & 79.5      &   91.3    & 84.3      & 79.5      &89.3        & 86.1  & \textbf{0.00}\\
			Baseline$^\star$ & 78.1    & 84.1    & 89.8    & 79.3   & 82.6      &  84.1     & 77.1      & 77.8      & 87.5      &  81.5     & 79.2      & 87.8      & 82.4 & 1034.80\\
			CMKD$^\star$ & 83.4   & 89.6   & \textbf{95.4}    &86.5   & 91.2     &  91.5    &  \textbf{89.4}     & 89.9      & 94.4       & 90.9      &  87.3     & 93.3      & 90.2 & 1034.80\\
			CMKD+RST$^\star$ & 82.3   &89.1    & 94.6     & 85.5  & 91.0      & 91.3       & 85.1      & 89.4      & 94.4      &  89.0     & 86.0      & \textbf{93.6}       & 89.3   & \textbf{0.88}\\
			CMKD+FixMatch$^\star$ & \textbf{85.0}   & \textbf{90.1}   & \textbf{95.4}    & \textbf{87.1}  & \textbf{93.4}      & \textbf{91.8}      & 86.1      &   \textbf{91.2}    & \textbf{95.2}      & \textbf{91.0}      & \textbf{90.3}      & 93.5      & \textbf{90.8} & 1034.80\\
			\bottomrule
		\end{tabular}%
	}
	\label{table11}%
\end{table*}%
In real-world scenarios, it is common for the categories in the target domain to be a subset of the source domain, which can be regarded as a special case of the traditional UDA, called Partial-set Domain Adaptation (PDA) \cite{ref81}. To further validate the validity and flexibility of our method, we try to apply CMKD to the PDA task while only making simple modifications to the original method. When utilizing CMKD and RST for PDA, the key aspect is to estimate the categories for the target dataset. This prevents the model from adapting to classes that are not part of the target domain, thereby avoiding potential performance degradation. Inspired by SHOT \cite{ref59}, we count the number of predicted categories and estimate the categories on the target domain using a thresholding method. The counting of category $i$ can be formulated as follows:
\begin{eqnarray}
	n_{i}=\sum_{j=1}^{n_{t}} \mathbbm{1}_{\left[\text{argmax} \left(p_{h j}^{t}(y|{{x}_{j}^{t}})\right)=i\right]}
\end{eqnarray}
Next, PDA can be implemented with CMKD by making a simple modification to the predicted probability of the samples $x_{j}^t$, represented as follows:
\begin{eqnarray}
	\hat{p}_{h j}^{t}(y=c|{{x}_{j}^{t}})= \mathbbm{1}_{\left[n_{c}\ge T_{pda} \right]}p_{h j}^{t}(y=c|{{x}_{j}^{t}})
\end{eqnarray}
where $\hat{p}_{h j}^{t}(y=c|{{x}_{j}^{t}})$ represents the modified probability of category $c$, $p_{h j}^{t}(y=c|{{x}_{j}^{t}})$ denotes the original probability of category $c$, $n_{c}$ signifies the counting of categories $c$, and $T_{pda}$ is a predefined threshold. The PDA task can be realized by employing CMKD with the modified probability instead of the original probability. In the PDA task, the weight of the classification loss is set to one-half of that in the UDA task, while the transfer loss is reduced to one-tenth of the UDA task's weight. $T_{pda}$ is selected to be 14.

For the experiment settings, we adopt the protocol outlined in \cite{ref82} for the Office-Home \cite{ref41} dataset. Specifically, the target domain in Office-Home \cite{ref41} consists of a total of 65 classes, and we focus on 25 classes (the first 25 in alphabetical order) for our analysis. CMKD is compared with the SoTA method for PDA, respectively:	DRCN \cite{ref82}, BA$^{\text{3}}$US \cite{ref83}, TSCDA \cite{ref84}, SHOT \cite{ref59}, SHOT++ \cite{ref85}. As observed from Table VIII, it is evident that CMKD and RST demonstrates exceptional performance in the PDA task, achieving state-of-the-art results. Moreover, this highlights the versatility of CMKD and RST in the PDA task, as it remains effective across different architectures.

\subsection{\textbf{Generalized to Universal Image Segmentation Model}}

\begin{table*}[t]
	\setlength{\abovecaptionskip}{0.cm}
	\setlength{\belowcaptionskip}{-0.cm}
	\centering
	\caption{Using Mask2Former for UDASS on Cityscape$\rightarrow$ACDC. }
	\renewcommand{\arraystretch}{1.0}
	\setlength{\tabcolsep}{0.5mm}{
		\begin{tabular}{c|ccccccccccccccccccccc}
			\toprule
			Method &Road &S.walk &Build. &Wall &Fence &Pole &Tr.Light &Sign &Veget. &Terrain &Sky &Person &Rider &Car &Truck &Bus &Train &M.bike &Bike &mIOU & DSP(M)\\
			\midrule
			\makecell[l]{\textit{Deeplabv2:}} &    &    &     &   &       &       &       &       &       &       &       &       &  & &&&&&&&\\
			Deeplabv2 & 71.9& 26.2& 51.1& 18.8& 22.5& 19.7& 33.0& 27.7& 67.9& 28.6& 44.2& 43.1& 22.1& 71.2& 29.8& 33.3& 48.4& 26.2& 35.8& 38.0 & 42.51\\
			DACS & 58.5& \textbf{34.7}& \textbf{76.4}& 20.9& 22.6& \textbf{31.7}& 32.7& \textbf{46.8}& 58.7& \textbf{39.0}& 36.3& 43.7& 20.5& 72.3& 39.6& 34.8& 51.1& 24.6& 38.2& 41.2&42.51\\
			FDA & \textbf{73.2}& \textbf{34.7}& 59.0& \textbf{24.8}& \textbf{29.5}& 28.6& \textbf{43.3}& 44.9& \textbf{70.1}& 28.2& \textbf{54.7}& \textbf{47.0}& \textbf{28.5}& \textbf{74.6}& \textbf{44.8}& \textbf{52.3}& \textbf{63.3}& \textbf{28.3}& \textbf{39.5}& \textbf{45.7}&42.51\\
			\midrule
			\midrule
			\makecell[l]{\textit{Mask2Former:}} &    &    &     &   &       &       &       &       &       &       &       &       &  & &&&&&&&\\
			Mask2Former & 85.4 & 55.7 & 52.4 & 23.3 & 23.0 & 41.2 & 68.4 & 52.6 & 71.8 & 30.2 & 68.9 & 48.1 & 21.4 & 74.5 & 30.7 & 19.4 & 14.5 & 24.4 & 23.2 & 43.6 &44.0\\
			CMKD  & \textbf{86.2}    & \textbf{62.0}   & \textbf{77.3}    & \textbf{26.9}   & \textbf{29.4}      & \textbf{45.3}      & \textbf{70.6}      & \textbf{56.4}      & \textbf{76.1}      & \textbf{31.6}      & \textbf{74.4}      & \textbf{50.6}      & 20.0  &\textbf{78.0} & 36.4 &\textbf{35.5} &\textbf{32.3} &\textbf{38.8} & 40.4 & \textbf{51.0} & 44.0 \\
			CMKD+RST & \textbf{86.2}   & 61.4   & 77.0 & 24.9  & 27.6  & 43.5      & 69.0      & 55.6      &  75.5     & 30.3      &74.0       & 48.6      & \textbf{25.1}      & 77.2 & \textbf{39.1} & 32.4 & 29.2 & 35.2 & \textbf{45.5} & 50.4 & \textbf{5.52}  \\
			\bottomrule
		\end{tabular}%
	}
	\label{tab3}%
\end{table*}%

In this section, we opt for the Universal Image Segmentation Model to delve into the Unsupervised Domain Adaptive Segmentation (UDASS) task, specifically utilizing the Mask2Former \cite{ref93} architecture. Typically, in UDASS experiments, the Deeplabv2 \cite{ref91} model with ResNet101 \cite{ref17} backbone is commonly employed. However, as Mask2Former \cite{ref93} contains more parameters in the decoder, to ensure a fair comparison with similar parameter counts, we select the Mask2Former model with ResNet50 backbone for our experiments. Considering that Mask2Former has been supervised trained on the Cityscape \cite{ref90} dataset in the upstream task, we opt for the Cityscape\cite{ref90}$\rightarrow$ACDC\cite{ref94} benchmark for exploration in this subsection, preventing data leakage.

In the classification task, CMKD was primarily designed based on the technique of Gini impurity. However, in the segmentation task, methods related to entropy optimization are not directly applicable to UDASS. Therefore, we need to adapt CMKD to suit the segmentation task. We draw inspiration from the mean-teacher \cite{ref95} approach, which generates pseudo-labels to guide the training of student models using an Exponential Moving Average (EMA) teacher model. As a result, $\mathcal{L}_{\mathrm{task}}$ can be redefined as:
\begin{eqnarray}
	\mathcal{L}_{task}= \mathcal{L}_{m2f}(x^{t}, \tilde{y}^{t}; F)
\end{eqnarray}
\begin{eqnarray}
	\tilde{y}^{t}_{i j}=\underset{c}{argmax}\text{ } \overline{F}(x^t)_{i j c}
\end{eqnarray}
\begin{eqnarray}
	\overline{F}\gets m \overline{F} + (1-m)F_{T}
\end{eqnarray}
where $F(\cdot )=E(\cdot )\circ D_{h}(\cdot )$. $E$ represents the backbone network, while $D_{h}$ encompasses the Mask2Former Head, comprising the Transformer Decoder and Pixel Decoder. $D_{h}$ is the functional equivalent of the task-head in the classification task $\mathcal{L}_{m2f}$ denotes the loss function associated with the Mask2Former, for which no modifications were made to the network structure or the loss function utilized in the original paper. $\overline{F}$ represents the teacher model generated through Exponential Moving Average (EMA) and $F_{T}$ is the model of training step $T$.

For the distillation term $\mathcal{L}_{distill}$, we adhere to the concept of CMKD and generate pseudo-labels by averaging the predictions of the pre-trained head $D_{g}(\cdot )$ and the predictions of the EMA model. During training, the pre-training head does not update the weights. These pseudo-labels are then employed to guide the training of the model.

\begin{eqnarray}
	\mathcal{L}_{distill}= \mathcal{L}_{m2f}(x^{t}, \check{y}^{t}; F)
\end{eqnarray}
\begin{eqnarray}
	\check{y}^{t}_{i j}=\underset{c}{argmax}\text{ } 0.5\cdot(\overline{F}(x^t)_{i j c} + \check{F}(x^t)_{i j c})
\end{eqnarray}
where $\check{F}(\cdot )=E(\cdot )\circ D_{g}(\cdot )$. In the segmentation task, the constraint term is removed, and the modified CMKD formulation is denoted as:
\begin{eqnarray}
	{{\mathcal{L}}_{\text{cmkd}}}=\alpha \cdot {{\mathcal{L}}_{\text{task}}}+(1-a)\cdot {{\mathcal{L}}_{\text{distill}}}
\end{eqnarray}
\begin{eqnarray}
	\alpha =\emph{sg}(\exp (-\text{KL}(\overline{F}(x^t)\parallel \check{F}(x^t))))
\end{eqnarray}
Finally, the total loss is written as:
\begin{eqnarray}
	{{\mathcal{L}}_{\text{total}}}=\lambda_{sup} \cdot{{\mathcal{L}}_{\text{sup}}} + \lambda_{cmkd} \cdot{{\mathcal{L}}_{\text{cmkd}}}
\end{eqnarray}
\begin{eqnarray}
	{{\mathcal{L}}_{\text{sup}}}=\mathcal{L}_{m2f}(x^{s}, y^{s}; F)
\end{eqnarray}
${\mathcal{L}}_{\text{sup}}$ is training loss on source domain. We experimented with MMSegmentation$^{1}$. For the choice of hyperparameters, it remains basically the same as Mask2Former, except that the initial learning rate we set to $1 \times 10^{-5}$. $\lambda_{sup}$ is set to 0.1, while $\lambda_{cmkd}$ uses the dynamic mechanism in the Implementation details in the manuscript, $\beta$ is set to 0.25. Image size cropped to 512$\times$512 and the momentum $m$ factor is set to 0.99. We adhered to the DACS \cite{ref100} data augmentation approach and utilized RCS \cite{ref92}. For RST, similar to Section J, we utilize a dynamic mechanism to set the threshold. Specifically, the threshold gradually increases from $2 \times 10^{-6}$ to $1 \times 10^{-4}$ as the number of training rounds progresses. We compare with methods such as Deeplabv2 \cite{ref91}, DACS \cite{ref100}, and FDA \cite{ref101}. In contrast to previous transfer learning tasks where the Backbone is typically retained and a completely new randomly initialized task head is created, we directly initialize the task head using pre-trained weights in our approach. This decision is motivated by the fact that the decoder component of Mask2Former contains a substantial number of weights, accounting for nearly half of the model's total weights, and can provide an efficient initialization for the model. Additionally, previous discussions regarding PEFT have focused on its application to the backbone, while the task head is preserved in its entirety. However, this approach results in significant parameter storage requirements in Mask2Former due to the large number of task head parameters. Therefore, we employ RST for both the backbone and decoder components in this part.

Table IX illustrates the generalization capabilities of CMKD and RST on Mask2Former, along with the robustness of Mask2Former to domain shifts compared to deeplabv2, both with similar parameter counts. This comparison underscores the superiority of the Mask2Former structure. Moreover, we made slight modifications to the implementation of CMKD based on Mask2Former, resulting in a notable improvement of +7.4 mIOU compared to Mask2Former alone. This enhancement underscores the potential of CMKD in image segmentation. Furthermore, when CMKD is combined with RST, only approximately 10\% of the total number of parameters need to be preserved to achieve results comparable to full parameter fine-tuning. We anticipate that this compression ratio can be further reduced with larger pre-trained models and datasets.

\setlength\extrarowheight{-3pt}
\let\thefootnote\relax\footnote{$^{\text{1}}$\url{https://github.com/open-mmlab/mmsegmentation/blob/main/configs/mask2former/README.md}}
\let\thefootnote\relax\footnote{$^{\text{2}}$\url{https://github.com/open-mmlab/mmflow/blob/master/configs/raft/README.md}}

\subsection{\textbf{RST on Optical Flow}}

\begin{table}[t]
	\setlength{\abovecaptionskip}{0cm}  
	\setlength{\belowcaptionskip}{-0.0cm} 
	\centering
	\caption{Comparison with PEFT methods on KITTI2015.}
	\renewcommand{\arraystretch}{1.0}
	\setlength{\tabcolsep}{4.0 mm}{
		\begin{tabular}{c|cc}
			\toprule
			Method & EPE($\downarrow$) & DSP (M) \\
			\midrule
			Fine-Tuning & \textbf{0.67} & 4.1878 \\
			Linear-Probe & \textbf{0.67} & 3.1210\\
			BitFit & 1.12 & \textbf{0.0043}\\
			LoRA &  1.32 & 0.0163 \\
			\midrule
			RST & 0.96 & 0.0181\\
			RST$_{bias}$ &\textbf{0.69} & \textbf{0.0043}\\
			\bottomrule
		\end{tabular}%
	}
	\label{table6}%
\end{table}%

In this subsection, we focus on the task of optical flow, which involves estimating per-pixel motion between video frames. To evaluate the effectiveness of RST, we employ the RAFT \cite{ref96} architecture. There exist several settings for the optical flow task. For instance, models can be randomly initialized and trained on FlyingChairs \cite{ref98} and FlyingThings \cite{ref99} datasets, and then directly tested on the Sintel dataset. Alternatively, after pre-training on the serval datasets, models can be fine-tuned on the down-stream dataset. Given that RST is a parameter-efficient training algorithm that relies on pre-training knowledge, we solely consider the second task setting. We adopt the setup employed in the MMFlow$^{2}$ implementation of RAFT\cite{ref96}, which involves pre-training on a mixed dataset followed by fine-tuning on KITTI2015 \cite{ref97}. The mixed dataset consisted of FlyingChairs\cite{ref98}, FlyingThing3d\cite{ref99}, Sintel \cite{ref102}, KITTI2015\cite{ref97}, and HD1K \cite{ref103}. Additionally, we utilize the pre-training weights provided by MMFlow$^{2}$ for fine-tuning on downstream tasks. We opted to compare it with various PEFT methods, notably BitFit \cite{ref70}, LoRA \cite{ref14}. As with Section L, PEFT is used for both encoder and decoder. The rank of LoRA is set to 1. During the reproduction process, we encountered a significant deviation in the results when using the profiles and pre-training weights provided by MMFlow. These results were notably distant from those reported in the original paper. Upon examination, it became apparent that the model was experiencing underfitting. To address this issue, we extended the training duration to 100,000 steps, leveraging the original profile as a reference. 

For RST, a slight adjustment is necessary. Given the limited capacity of the model used for RAFT, it becomes challenging to determine the optimal thresholds for parameter filtering using predefined thresholds. As a workaround, we adopt a strategy wherein we select, the parameter with the top $r\%$ change, gradually decreasing the value of r from 100 to 0.2 over time. It's important to note that this parameter selection method is utilized only when the model possesses a very small parameter count. In scenarios where the parameter count is large, this selection method may strain the CPU significantly, and employing the predefined threshold method would be a more efficient choice.

Table X reveals unexpected experimental findings. Despite RAFT undergoing extensive pre-training, techniques like BitFit and LoRA prove challenging to adapt effectively to downstream tasks. Although RST outperforms BitFit and LoRA, it still falls behind methods that train a larger number of parameters, such as Fine-Tuning. This phenomenon may stem from RAFT's relatively small model size, making it challenging for parameter-efficient fine-tuning methods to effectively adjust the model weights. Based on this observation, we analyzed the positions of the non-zero weights obtained by RST and found that many of them are Bias terms. Given this, one might wonder why BitFit cannot achieve similar results. Intuitively, we decided to utilize RST to exclusively train the Bias items while retaining all Bias items, denoted as RST$_{bias}$. Conceptually, RST$_{bias}$ aligns with BitFit. However, since optical flow tasks often involve gradient norm clipping techniques during training, BitFit's gradient constraints are derived solely from the Bias term, whereas RST$_{bias}$'s gradient constraints encompass all parameters. Surprisingly, this simple modification significantly enhances model performance, approaching results close to those achieved by Fine-Tuning approximation, and requires only fine-tuning of the Bias term. This underscores the considerable potential of related techniques such as parameter-efficient fine-tuning in optical flow tasks.

\subsection{\textbf{Analysis of CMKD and RST}}
In this section, we will analyze and discuss CMKD and RST, focusing primarily on elucidating the reasons behind their effectiveness and exploring their applicability in deployment scenarios.

\textbf{Q1: How does CMKD reconcile the perspectives of the student and teacher models?}

To address this question, we initially simplify the gradient formula for CMKD. Given that the primary components of CMKD are $\mathcal{L}_{task}$ and $\mathcal{L}_{distill}$, and for the sake of analytical convenience, we exclude $\mathcal{L}_{reg}$ from our discussion. Thus, we define $\mathcal{L}$ as the combination of $\mathcal{L}_{task}$ and $\mathcal{L}_{distill}$. The gradient formula can be expressed as follows:
\begin{eqnarray}
	\begin{aligned}
		\frac{\partial \mathcal{L}}{\partial \theta}= &\alpha\cdot\frac{\partial \mathcal{L}_{\mathrm{task}}}{\partial \theta} + (1-\alpha)\cdot\frac{\partial \mathcal{L}_{\mathrm{distill}}}{\partial \theta}\\
		= & -\alpha\cdot\sum_{i=1}^{c}{[2\cdot p_{h}^{t}(y=i|{{x}^{t}})\cdot \frac{\partial p_{h}^{t}(y=i|{{x}^{t}})}{\partial \theta }]}\\
		 & -(1-\alpha)\cdot\sum_{i=1}^{c} p_{h}^{t}\left(y=i \mid x^{t}\right) \cdot \frac{\partial p_{h}^{t}\left(y=i \mid x^{t}\right)}{\partial \theta} \\
		& -(1-\alpha)\cdot\sum_{i=1}^{c} p_{\text{g}}^{t}\left(y=i \mid x^{t}\right) \cdot \frac{\partial p_{h}^{t}\left(y=i \mid x^{t}\right)}{\partial \theta}\\
		=&-\sum_{i=1}^{c} [(1+\alpha) \cdot p_{h}^{t}\left(y=i \mid x^{t}\right) \\  
		& +(1-\alpha) \cdot p_{g}^{t}\left(y=i \mid x^{t}\right)] \cdot \frac{\partial p_{h}^{t}\left(y=i \mid x^{t}\right)}{\partial \theta}
	\end{aligned}
\end{eqnarray}
For ease of discussion, we denote Eq. (10) as:
\begin{eqnarray}
	\begin{aligned}
		\frac{\partial \mathcal{L}}{\partial \theta}= &\alpha\cdot\frac{\partial \mathcal{L}_{\mathrm{task}}}{\partial \theta} + (1-\alpha)\cdot\frac{\partial \mathcal{L}_{\mathrm{distill}}}{\partial \theta}\\
		=&-\sum_{i=1}^{c}[(1+\alpha) \cdot p_{h(i)}^{t}  +(1-\alpha) \cdot p_{g(i)}^{t}] \cdot \frac{\partial p_{h(i)}^{t}}{\partial \theta}
	\end{aligned}
\end{eqnarray}
For detailed explanations regarding the calculation of $\mathcal{L}_{\mathrm{task}}$, $\mathcal{L}_{\mathrm{distill}}$ and $\alpha$ please refer to Section III.B of the manuscript. From equation (20), we observe that the scaling factor representing the student's perspective is $(1+\alpha) \cdot p_{h(i)}^{t}$, while the scaling factor representing the teacher's perspective is $(1-\alpha) \cdot p_{g(i)}^{t}$. Given that $\alpha$ takes values in the range [0, 1], training from the student's perspective will invariably dominate. This mechanism ensures that the model does not become excessively influenced by generic knowledge in instances where the teacher's perspective proves ineffective in providing adequate guidance.

Further, we can briefly explore the design philosophy of CMKD self-training from the perspective of whether the student and the teacher are aligned.

\textit{Alignment of teacher and student perspectives.} In this scenario, when both the teacher perspective and the student perspective predictions are correct, the model undergoes further training in a positive direction. Conversely, when both predicted labels are incorrect, the dynamic design of $\lambda_{1}$ can effectively mitigate the issue. Essentially, the model utilizes smaller training weights during the early stage, gradually increasing the trade-off as the training progresses. This straightforward dynamic training mechanism proves effective in preventing the model from being misled by incorrect training signals too early in the process. For specific implementation details, please refer to Section IV.A Implementation Details, of the manuscript.

\textit{Misalignment of teacher and student perspectives.} When the student perspective predicts incorrectly, there are two potential scenarios for the teacher perspective: it may predict correctly or incorrectly. However, due to the discrepancy in the distributions of the two predictions, the value of $\alpha$ may become very small, possibly nearing 0 at its minimum. This scaling factor $(1+\alpha) \cdot p_{h(i)}^{t}  +(1-\alpha) \cdot p_{g(i)}^{t}$ effectively smooths the learning intensity for each category, thereby reducing the model's learning rate for such samples. This adjustment prevents the model from being further misled in cases where the student perspective is incorrect, aiding in maintaining model stability and performance. Indeed, it is plausible to encounter scenarios where the student's perspective is correct while the teacher's perspective is incorrect. However, based on our previous analysis, we understand that the student perspective consistently dominates, with the teacher perspective serving more as a regularizer. Consequently, the model will ultimately learn in the correct direction, guided primarily by the accurate predictions from the student's perspective.

\textbf{Q2: Scenarios for the application of RST.}

From the experiments conducted in the paper, it is evident that RST effectively reduces the number of weights required to be preserved for downstream task fine-tuning. Moreover, the larger the scale of upstream task training, the more RST can reduce the number of parameters, as extensive upstream training already covers features necessary for the downstream task. Additionally, RST exhibits broad applicability beyond its original purpose of addressing the ``small model and large number of tasks'' challenge in UDA tasks. It demonstrates excellent generalization across various tasks and different pre-training weights.

For instance, taking CLIP as an example, we can utilize the original CLIP model as a solution for general problems. When faced with domain-specific problems, we can combine the sparse weights obtained from RST training with the pre-trained CLIP model, effectively transforming it into a specialized model for solving domain-specific knowledge. As the weights obtained from RST training are lightweight, we can store multiple specialized weights on deployment devices, enabling the system to solve problems in diverse domains while maintaining generality.

While multitasking techniques can enable a single model to tackle multiple tasks, conflicts between tasks may arise as the number of tasks increases. Moreover, adding a new task often necessitates retraining the model along with existing tasks to prevent forgetting previous task knowledge. Furthermore, accommodating multiple tasks can significantly increase model capacity. RST effectively mitigates these challenges by training individual tasks independently, thus eliminating the possibility of task conflicts and preserving the weights of previous tasks.

While structured reparameterization PEFT techniques like LoRA could provide similar advantages, experiments have shown that RST outperforms them, particularly for small-scale models and insufficient pre-training. However, it is important to acknowledge that RST has its limitations. RST does not possess features that mitigate explicit memory during model training. Nonetheless, in tasks where explicit memory is not a primary concern due to limited model capacity, RST proves highly effective. Looking ahead, further exploration of the RST technique could potentially contribute to fine-tuning large language models.

\bibliographystyle{IEEEtran}
\bibliography{reference_appendix}
	
\newpage
	
\vfill